\documentclass[10pt,twocolumn,letterpaper]{article}
\usepackage{iccv}              

\usepackage{amsmath,amsfonts,bm}

\def\eqref#1{equation~\ref{#1}}

\def\1{\bm{1}}

\DeclareMathAlphabet{\mathsfit}{\encodingdefault}{\sfdefault}{m}{sl}
\SetMathAlphabet{\mathsfit}{bold}{\encodingdefault}{\sfdefault}{bx}{n}

\usepackage{url}

\usepackage{booktabs} 
\usepackage{adjustbox} 
\usepackage{multicol}
\usepackage{multirow}
\usepackage{rotating}
\usepackage{pifont}
\newcommand{\cmark}{\ding{51}}%
\newcommand{\xmark}{\ding{55}}%
\usepackage{wrapfig}
\usepackage{xcolor}
\usepackage{float} 
\usepackage{amsmath} 
\usepackage{amssymb} 

\usepackage{colortbl}

\definecolor{navyblue}{HTML}{3B76AF}
\definecolor{nicepurple}{HTML}{C6B2D3}
\definecolor{nicegreen}{HTML}{559E3F}
\definecolor{GTgreen}{HTML}{459948}

\newcommand{\model}{ResidualViT}
\usepackage{subcaption}

\definecolor{iccvblue}{rgb}{0.21,0.49,0.74}
\usepackage[pagebackref,breaklinks,colorlinks,allcolors=iccvblue]{hyperref}

\title{ResidualViT for Efficient Temporally Dense Video Encoding}

{\author{
Mattia Soldan$^{1}$
\quad Fabian Caba Heilbron$^{3}$
\quad Bernard Ghanem$^{1}$
\quad Josef Sivic$^{2,3}$
\quad Bryan Russell$^{3}$\\
$^{1}$KAUST \quad\quad $^{2}$CIIRC CTU \quad\quad $^{3}$Adobe Research
}}

\begin{document}
\twocolumn[{%
\renewcommand\twocolumn[1][]{#1}%
\maketitle
\begin{center}
    \vspace{-.6cm}
    \centering
    \captionsetup{type=figure}
    \includegraphics[width=\textwidth, trim = 0cm 0cm 0cm 6mm]{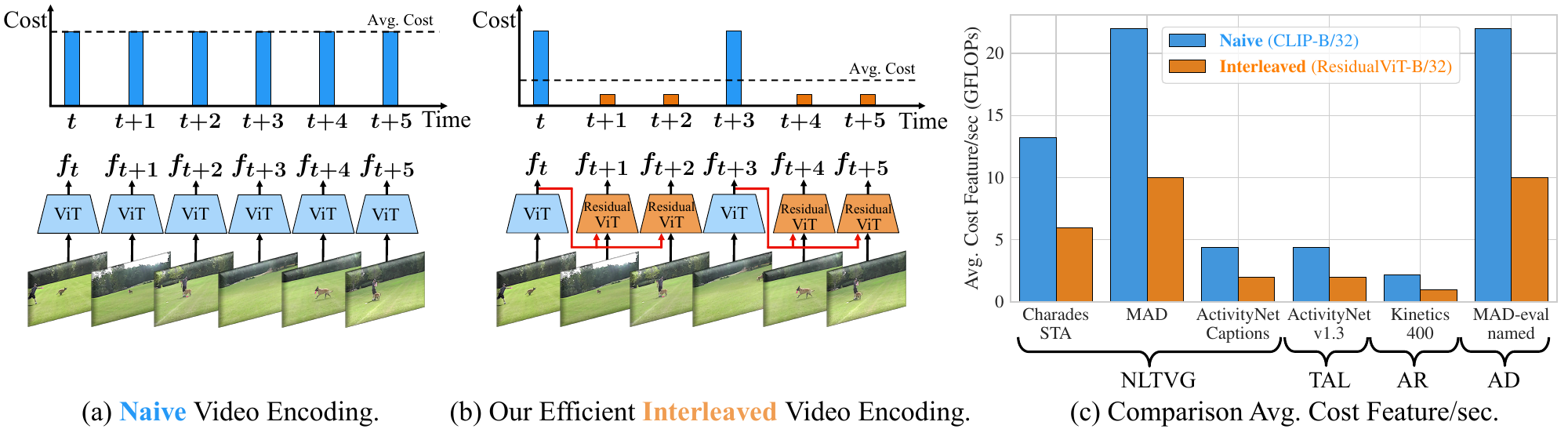}
    \vspace{-.6cm}
    \captionof{figure}{\textbf{Efficient temporally dense video encoding.}
    (a) Naively encoding videos incurs a high computational cost when computing frame-level features for temporally dense tasks.
    (b) Our efficient interleaved approach significantly reduces this cost, enabling efficient temporally dense feature extraction.
    (c) On benchmark datasets, \model{} reduces the computational cost by an average of $56\%$ compared to the CLIP encoder while \textbf{maintaining nearly identical accuracy} across multiple downstream tasks: Natural Language Temporal Video Grounding (NLTVG), Temporal Activity Localization (TAL), Audio Description generation (AD), and Action Recognition (AR).
    }
    \label{fig:data-overview}
\end{center}%
}]
\begin{abstract}
\label{sed:abstract}
Several video understanding tasks, such as natural language temporal video grounding, temporal activity localization, and audio description generation, require ``temporally dense'' reasoning over frames sampled at high temporal resolution. However, computing frame-level features for these tasks is computationally expensive given the temporal resolution requirements. In this paper, we make three contributions to reduce the cost of computing features for temporally dense tasks. 
First, we introduce a vision transformer (ViT) architecture, dubbed \model{}, that leverages the large temporal redundancy in videos to efficiently compute temporally dense frame-level features. Our architecture incorporates (i) learnable residual connections that ensure temporal consistency across consecutive frames and (ii) a token reduction module that enhances processing speed by selectively discarding temporally redundant information while reusing weights of a pretrained foundation model.
Second, we propose a lightweight distillation strategy to approximate the frame-level features of the original foundation model. 
Finally, we evaluate our approach across four tasks and five datasets, in both zero-shot and fully supervised settings, demonstrating significant reductions in computational cost (up to 60\%) and improvements in inference speed (up to 2.5$\times$ faster), all while closely approximating the accuracy of the original foundation model. 
\end{abstract}
{\let\thefootnote\relax\footnote{$^1$King Abdullah University of Science and Technology.}}
{\let\thefootnote\relax\footnote{$^2$Czech Institute of Informatics, Robotics and Cybernetics at the Czech Technical University in Prague.}}
\vspace{-.8cm}
\section{Introduction}
\label{sec:introduction}
Video content is now widespread across various platforms, creating a demand for scalable methods to enable video understanding applications. While dual-encoder foundation models~\citep{radford2021learning} have demonstrated impressive recognition capabilities and versatility through zero-shot learning, deploying these models over large video collections presents significant computational challenges. As videos are data-heavy, a common strategy to ease the computational burden is to sparsely sample frames throughout a video (\eg, $0.1{-}0.5$ FPS~\cite{luo2022clip4clip}) and compute their features. This strategy is effective for tasks that require reasoning over a few key frames in a video, such as retrieving short few-seconds video clips from a database~\cite{wang2022internvideo, ma2022x, gorti2022x, luo2022clip4clip, bain2021frozen}.

However, several video understanding tasks require ``temporally dense'' reasoning over frames sampled at a higher temporal resolution (Figure~\ref{fig:data-overview}\textcolor{navyblue}{a}). Examples of such temporally dense tasks include natural language temporal video grounding~\cite{anne2017localizing, gao2017tall}, temporal activity localization~\cite{liu2025opentad, caba2015activitynet}, and audio description generation~\cite{han2023autoad, lin2023mm}. Increasing the sampled frame rate from $0.1{-}0.5$ FPS to $1{-}5$ FPS for a temporally dense task requires $2{-}50\times$ more computational resources. Therefore, reducing the computational demands of current pretrained foundation models for these tasks is imperative for enabling their deployment at scale.

Prior approaches for reducing the compute cost of a pre-trained model primarily aim to distill a model's representation directly into a lower-capacity model~\citep{dehghani2023scaling,hao2022learning,heo2019comprehensive,tiny_vit}. While these efforts result in a more efficient model, distilling all the information from the larger model into the smaller one is challenging and often leads to a degradation in recognition accuracy. Moreover, these approaches naively treat video frames independently and do not explicitly take advantage of the temporal redundancy inherent in videos, which could further optimize processing.

To overcome these limitations, this work aims to compute video frame features efficiently given the pretrained vision transformer (ViT) tower of a dual-encoder foundation model (\eg, CLIP). As illustrated in Figure~\ref{fig:data-overview}\textcolor{navyblue}{b}, our solution capitalizes on the observation that nearby frames are often visually similar. Drawing inspiration from standard video compression techniques, which store a sparse set of {\em I-frames} (self-contained, fully-formed frames) and a denser set of {\em P-frames} (differences or changes from the previous frame), where the latter have high compression ratios (up to two orders of magnitude~\citep{wu2018compressed}), we adopt a similar strategy.

Our first contribution is an approach that computes the full ViT model representation on a sparse set of frames while providing an efficient approximation for representing the dense set of nearby frames. This interleaved strategy effectively mirrors the I-frame and P-frame method used in video compression, leading to significant reductions in computational demand. We refer to the two sets of output representations as {\em I-features} (self-contained computed via a regular full ViT model) and {\em P-features} (efficiently computed using I-features and exploiting the temporal continuity of video). 
To compute the efficient P-features, we propose a vision transformer architecture (dubbed {\em \model{}}) that comprises two changes to the architecture of the pretrained ViT encoder. 
First, we compute a learnable {\em residual token} given a nearby I-feature. This residual token allows the \model{} encoder to exploit the temporal continuity of nearby video frames by incorporating their computed features. Second, we include a token reduction module~\citep{ding2023prune, haurum2023tokens, hou2022token, bolya2022token} in the \model{} encoder that significantly reduces the number of tokens used to compute P-features, substantially reducing their encoding costs. Combining these modules allows the \model{} encoder to efficiently and accurately approximate the target features.

As our second contribution, we propose a student-teacher distillation training objective that minimizes the loss between the vision-language embedding similarities produced by our efficient \model{} encoder and the features obtained from CLIP's pretrained ViT backbone. This setup enables our \model{} encoder to replicate features from CLIP. The training is lightweight, as only the residual tokenizer module is learned while the ViT weights remain frozen. This strategy allows us to fully harness the capabilities of CLIP without the need for large-scale training.  

As our third contribution, we evaluate \model{}'s efficient video encoding on four tasks requiring temporally dense features computation: Natural Language Temporal Video Grounding (NLTVG), Audio Description (AD) generation, Temporal Activity Localization (TAL), and Action Recognition (AR). We evaluate performance on five diverse datasets: MAD~\cite{soldan2022mad}, Charades-STA~\cite{gao2017tall}, ActivityNet-Captions~\cite{Krishna_2017_ICCV}, ActivityNet-v1.3~\cite{caba2015activitynet}, and Kinetics-400~\cite{kay2017kinetics}. Across all tasks and datasets, our model achieves significant reductions in frame encoding costs (up to $60\%$, illustrated in Figure~\ref{fig:data-overview}\textcolor{navyblue}{c}) and increased inference speed (up to $2.5\times$ faster, see Section~\ref{sec:supplementary-time-efficiency} of supplementary), all while closely approximating the accuracy of the original foundation model. 
Furthermore, our experiments span both zero-shot and fully-supervised settings, as well as short- and long-form videos, highlighting the versatility of our proposed architecture.

\section{Related Work}
\label{sec:related-work}
\noindent\textbf{Image Foundation Models for Video Applications.}
The analysis of video data introduces many technical challenges arising from its inherent temporal and spatial complexities, large data volume, and high temporal redundancy. As a way to mitigate these challenges and ease the development of new tools, the research community has resorted to applying image-based models~\citep{he2016deep, radford2021learning,simonyan2014very} to video tasks~\citep{castro2022fitclip, diwan2023zero, luo2022clip4clip,  nam2021zero, soldan2022mad, soldan2021vlg} with much success despite the image-based architectures' inability to reason about the temporal dimension. 
Moreover, dedicated temporal modeling~\citep{liu2023revisiting, ma2022x, tu2023implicit, xue2022clip} can offer potential accuracy gains at the expense of increased computational demands, highlighting a nuanced balance of efficiency and efficacy.
Our work capitalizes on the CLIP image foundation model~\citep{radford2021learning} to build an efficient video feature extraction framework that can be adopted for multiple downstream video tasks. We choose CLIP because of its excellent performance on multiple tasks~\citep{radford2021learning, shen2021much, Lin2022frozen} and native multi-modality (image and text), which can be adapted for video processing. Previous approaches leveraging CLIP for video tasks have utilized various strategies. These include applying temporal aggregation over frame representations~\citep{buch2022revisiting, luo2022clip4clip, ni2022expanding}, fine-tuning the model to capture motion patterns in videos~\citep{castro2022fitclip, weng2023open}, and employing carefully designed spatial and temporal adapters to harness the valuable pre-trained weights without modification~\citep{Lin2022frozen, pan2022st, yang2023aim, park2023dual}. Additionally, some methods have introduced prompt learning as a mechanism for domain adaptation~\citep{ju2022prompting}. In a similar spirit, our work seeks to leverage pre-trained network weights without modification; however, we focus on reducing the computational cost of encoding individual frames by minimizing redundant temporal computations while preserving essential semantic details.

\begin{figure*}[!t]
    \centering
        \includegraphics[trim={0cm 0cm 1.5cm 0cm},width=\textwidth,clip]{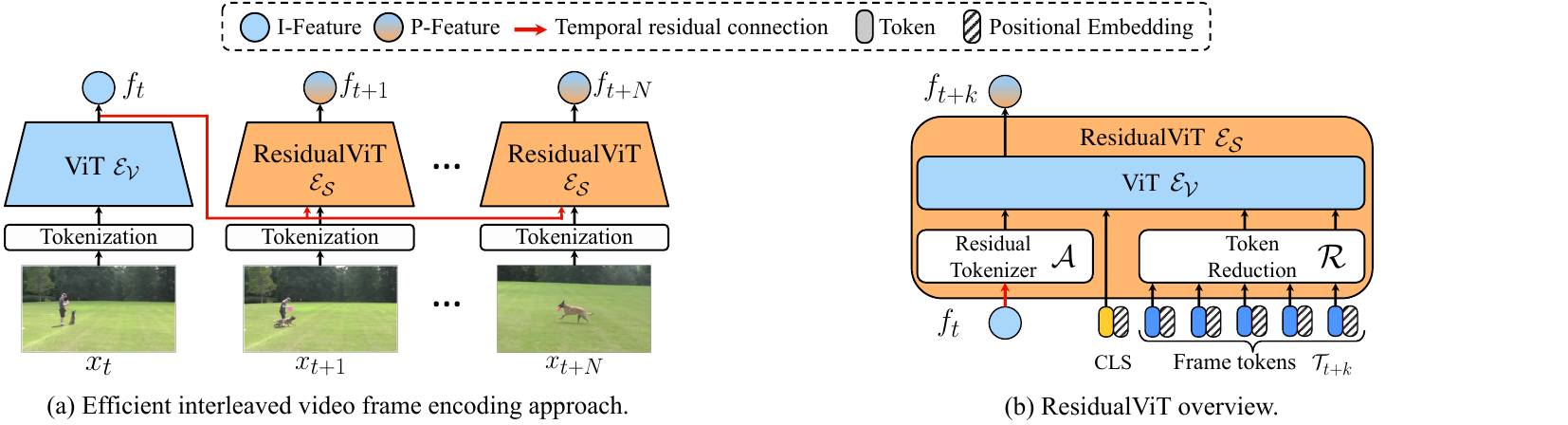}
    \vspace{-.5cm}
    \caption{\textbf{Model overview.} (a) Video frames are processed via two visual encoders $\mathcal{E}_{\mathcal{V}}$ and $\mathcal{E}_{\mathcal{S}}$ in an \textit{interleaved} manner. For each frame encoded via the ViT $\mathcal{E}_{\mathcal{V}}$, $N$ subsequent frames are encoded using our lightweight \model{} $\mathcal{E}_{\mathcal{S}}$, significantly reducing the computational cost. (b) \model{} incorporates a token reduction module $\mathcal{R}$ to reduce computation and the residual tokenizer $\mathcal{A}$ to ensure temporal consistency by propagating information from preceding frames.
    }
    \label{fig:model-overview}
    \vspace{-.2cm}
\end{figure*}

\noindent\textbf{Efficient Video Representations and Distillation.}
Prior work has also looked at distilling into a lower-capacity model~\citep{dehghani2023scaling, hao2022learning, heo2019comprehensive, tiny_vit} or developing efficient video representations for tasks such as semantic video segmentation~\citep{liu2020efficient} or video recognition~\citep{lin2019tsm,wu2022memvit,wu2018compressed}. The former approaches result in a degradation of recognition accuracy due to the difficulty of distilling to a small model from a larger model. The latter approaches have investigated how to efficiently compute convolution in time~\citep{lin2019tsm}, leverage the video compression representation in a convolutional network~\citep{wu2018compressed}, or avoid computing the cross-attention in time for long videos~\citep{wu2022memvit}. 
Additionally, other methods tackle the efficient inference challenge through network pruning~\citep{fang2023structural, molchanov2016pruning, he2017channel} reducing the number of parameters in convolutional networks and, consequently, the computational cost of pre-trained models.
In contrast, we target recent transformer-based architectures such as ViT~\citep{dosovitskiy2020image}, which offer strong scaling properties and are paired with language encoders, enabling zero-shot capabilities, unlike lightweight convolutional models such as X3D~\cite{Feichtenhofer_2020_CVPR} or MobileNet variants~\cite{howard2017mobilenets, sandler2018mobilenetv2, koonce2021mobilenetv3}, which lack these features.
Moreover, we focus on single-frame representations (such as CLIP~\citep{radford2021learning}) that are often the video representation of choice for their versatility in large-scale practical setups involving natural language~\citep{castro2022fitclip,luo2022clip4clip,soldan2022mad}, and consider the task of natural language video grounding, discussed next.
\section{Efficient Interleaved Video Encoding}
\label{sec:method}

We propose a video encoding approach for \textit{temporally dense tasks} that require encoding frames sampled at a high temporal resolution. To alleviate the computational burden, we adapt dual-encoder transformer-based pretrained models, focusing on improving the efficiency of the visual encoder tower (ViT)~\citep{dosovitskiy2020image}. Our approach exploits temporal redundancy in videos, where consecutive frames share overlapping visual and semantic content, making independent encoding computationally inefficient.

In our approach, we adopt the visual encoder from the dual-encoder vision-language model CLIP~\citep{radford2021learning}, which has demonstrated strong zero-shot recognition accuracy across various downstream tasks. CLIP provides a robust foundation for our visual encoder while also offering a paired language encoder, enabling both uni- and multi-modal tasks.

Figure~\ref{fig:model-overview}\textcolor{navyblue}{a} outlines our efficient visual encoding approach. 
Our encoding strategy interleaves computing features using the original visual encoder $\mathcal{E}_\mathcal{V}$ (I-features) with computing $N$ features using a fast approximate visual encoder $\mathcal{E}_\mathcal{S}$ (P-features).  We define $N$ as the interleave factor between I- and P-features.
Consider a video comprised of $n_v$ frames extracted at a constant frame rate, denoted as $\mathcal{X} = \{x_t\}_{t=1}^{n_v}$ with $x_t \in \mathbb{R}^{H\times W \times C}$, where $H$, $W$ and $C$ are the height, width, and number of channels of each frame. In alignment with standard vision transformer data processing, each frame $x_t$ is divided into patches $\{x_{t,j}\}_{j=1}^{K}$, with patch size (P), which are projected into an embedding space to form a set of tokens  $\mathcal{T}_t = \{\tau_{t,j}\}_{j=1}^K$ with $\tau_{t,j} \in \mathbb{R}^{d}$, where $d$ is the token dimension, $t$ the frame index and $j$ the token index.  The number of tokens is determined following the relationship $K = H \times W / P^2$.

We embed every $N+1$-th frame $x_t$ by applying the visual encoder $\mathcal{E}_\mathcal{V}: \mathbb{R}^{K \times d} \rightarrow \mathbb{R}^{b}$ on all frame tokens $\mathcal{T}_t$ to obtain an I-feature representation, $f_t = \mathcal{E}_\mathcal{V}(\mathcal{T}_t) \in \mathbb{R}^b$, where $b$ is the feature dimension. This process represents the standard encoding procedure for a given input frame.
The subsequent $N$ frames $\{x_{t+k}\}_{k=1}^{N}$ are encoded using our \model{} encoder $\mathcal{E}_\mathcal{S}: \mathbb{R}^{b} \times \mathbb{R}^{K \times d} \rightarrow \mathbb{R}^{b}$ to obtain P-features. Formally, we compute the P-features for those $N$ frames as $f_{t+k}=\mathcal{E}_\mathcal{S}(f_t, \mathcal{T}_{t+k}) \in \mathbb{R}^b$, where I-feature $f_t$ from frame $x_t$ is routed through the temporal residual connection (shown in red in Figure~\ref{fig:model-overview}) to the \model{} encoder $\mathcal{E}_\mathcal{S}$ as temporal context. Note that in our work, we use the output representation of the \texttt{[CLS]} token from the transformer architecture as our feature representation.   

The following provides a detailed explanation of the design of our \model{} architecture (Section~\ref{sec:residualvit-architecture}) and the associated training strategy (Section~\ref{sec:method-training}).

\subsection{ResidualViT Architecture}
\label{sec:residualvit-architecture}

Our technical solution involves equipping the ViT encoder with two key components, as illustrated in  Figure~\ref{fig:model-overview}\textcolor{navyblue}{b}: (i) \textit{a token reduction module} $(\mathcal{R})$ and (ii) \textit{a residual tokenizer module} $(\mathcal{A})$. The token reduction module selectively prunes input tokens to the ViT, retaining only the most informative ones, to significantly reduce the encoding computational cost.  
Concurrently, the residual tokenizer module propagates information from the last I-feature to the current P-feature compensating for the information discarded by the token reduction process. 

Reducing the token count improves efficiency, but selecting which tokens to discard to minimize loss of information remains a challenge. We explore three token reduction strategies:  
(i) token dropping (PatchDropout)~\citep{ding2023prune, haurum2023tokens, hou2022token, liu2023patchdropout}, which removes tokens based on a dropping probability $p$;  
(ii) token merging~\citep{bolya2022token}, which progressively reduces the number of tokens by $r$ at each transformer layer;  
and (iii) frame resolution reduction, which decreases the number of patches per frame. For details on token dropping strategies (\eg, random, uniform, center, and motion-based), see Section~\ref{sec:supplementary-drop-strategy} 
of supplementary. A comprehensive ablation study (Section~\ref{sec:supplementary-ablations} of supplementary) shows that token dropping provides the best trade-off between efficiency and accuracy.

In our \model{} architecture, the token reduction module is used during both training and inference to reduce computational overhead.  
This setup implies that part of the visual information is discarded. Yet, thanks to the temporal redundancy of consecutive frames, we seek to exploit the semantic information present in the feature computed at time step $t$ to recover the missing spatial information induced by the token reduction operation at time step $t+k$. 
In detail, the \model{} architecture takes as input I-feature $f_t$ from frame $x_t$ via the temporal residual connection and transforms this feature into a residual token as $\mathcal{A}(f_t) \in \mathbb{R}^{d}$ via a learnable mapping $\mathcal{A}: \mathbb{R}^{b} \rightarrow \mathbb{R}^{d}$. 
This operation plays a role analogous to that of a patch tokenizer: it transforms inputs into a token representation that is compatible with the input space of the visual encoder $\mathcal{E}_\mathcal{V}$.
The residual token is then concatenated with the \texttt{[CLS]} token and a small subset of frame tokens output by the token reduction module $\mathcal{R}(\mathcal{T}_{t+k})$. The resulting concatenated tokens are then fed into the visual encoder $\mathcal{E}_\mathcal{V}$ to obtain P-feature $f_{t+k}$. In our work, we implement the residual tokenizer $\mathcal{A}$ as a linear transformation. The addition of the residual token to the input of the transformer encoder adds a negligible computational overhead of about 0.1 GFLOPS (\ie, $0.1\%$ of the frame encoding cost using the CLIP ViT-L/14 backbone).
Despite the mapping $\mathcal{A}$ being a small linear layer, our solution is capable of providing informative cues even when most frame tokens are unavailable.  

Following our design, when token dropping is used, the average embedding cost of our approach can be approximated as:
\begin{equation}
    \label{eq:cost-approximation}
    C = \frac{C_{\mathcal{E}_\mathcal{V}} + NC_{\mathcal{E}_\mathcal{S}}}{1+N} \approx C_{\mathcal{E}_\mathcal{V}} \frac{1+ (1-p)N}{1+N},
\end{equation}
where $C_{\mathcal{E}_\mathcal{V}}$ and $C_{\mathcal{E}_\mathcal{S}}$ are the costs of encoding a frame using the visual encoder $\mathcal{E}_\mathcal{V}$ and $\mathcal{E}_\mathcal{S}$, respectively. Here, the interleave factor $N$ corresponds to the number of frames encoded by the \model{} with the reduced cost, and $p$ is the token reduction probability. 
Under token dropping, the cost of $\mathcal{E}_\mathcal{S}$ can be approximated as $C_{\mathcal{E}_\mathcal{S}}  \approx (1-p) C_{\mathcal{E}_\mathcal{V}}$.
It should be noted that when $N > 0$ and $p > 0$, the average embedding cost $C$ is strictly lower than $C_{\mathcal{E}_\mathcal{V}}$. 
For empirical evidence demonstrating the reduction in wall-clock time for frame encoding when utilizing \model{} compared to a standard ViT, refer to Section~\ref{sec:supplementary-time-efficiency} in supplementary.

\subsection{Training \model}
\label{sec:method-training}

We seek to train the residual tokenizer module $\mathcal{A}$, our only trainable component, such that the output frame feature computed by our \model{} $\mathcal{E}_{\mathcal{S}}$ closely approximates the feature computed via the original ViT encoder $\mathcal{E}_{\mathcal{V}}$ for the same frame. 
The challenge lies in the fact that the original ViT encoder has access to every token $\mathcal{T}_{t+k}$ from the input frame while the transformer encoder of our \model{} only receives a sparse set of frame tokens due to the token reduction module $\mathcal{R}(\mathcal{T}_{t+k})$ together with the residual token $\mathcal{A}(f_t)$ (Figure~\ref{fig:model-overview}\textcolor{navyblue}{b}). 
We achieve this objective via feature distillation~\citep{heo2019comprehensive, hinton2015distilling, ilharco_gabriel_2021_5143773}, where the original foundation model serves as a ``teacher'' network while our \model{} acts as the ``student'' network. 
In our study, we leverage the powerful CLIP~\citep{radford2021learning} foundation model to initialize our transformer encoders (\eg, ViT-B/32, ViT-B/16, or ViT-L/14). We fully exploit the CLIP model by including its language encoder $\mathcal{E}_\mathcal{L}$ in the feature distillation pipeline and perform the training using paired video and language samples. 

\begin{figure}[!t]
    \centering
        \includegraphics[trim={0cm 0cm 0cm 0cm},width=\linewidth,clip]{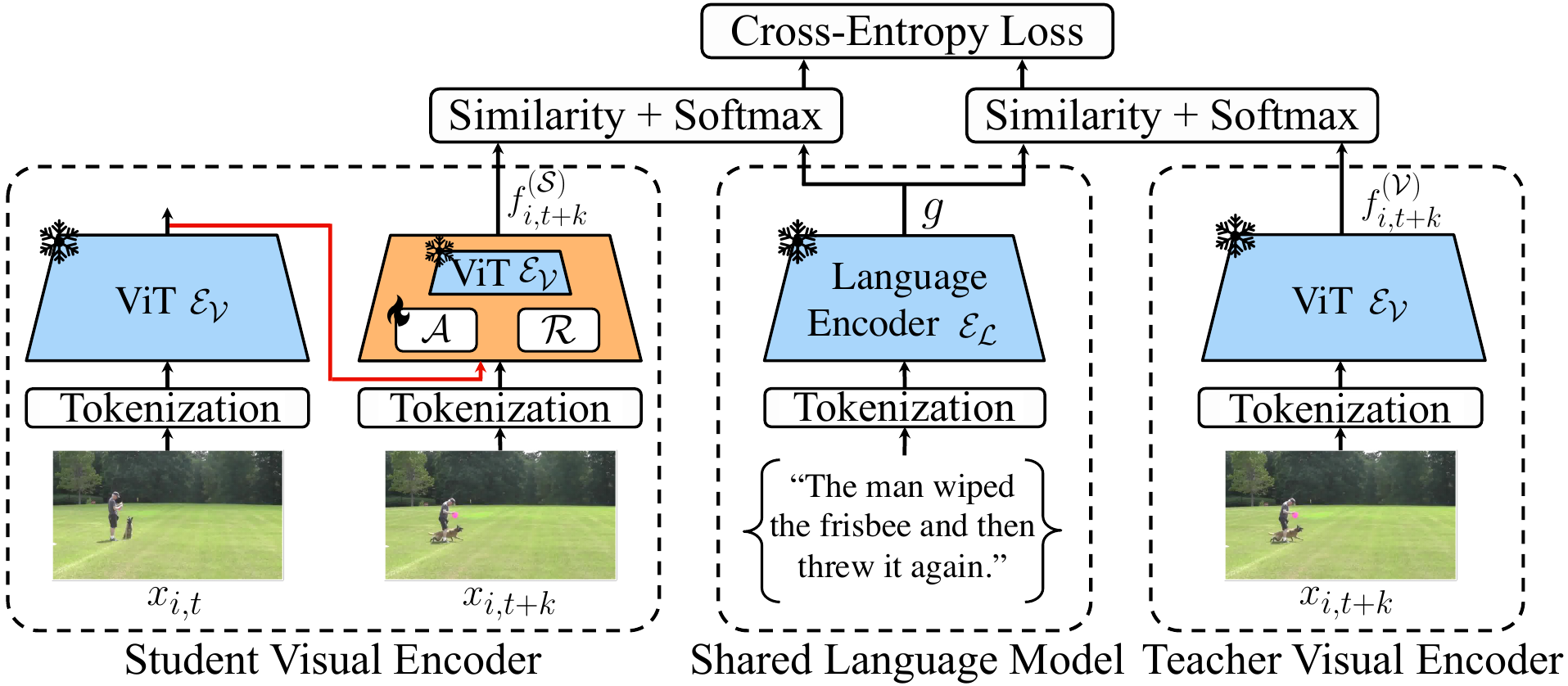}
        \vspace{-.5cm}
        \caption{\textbf{\model{} training ($\mathcal{J}_{L\rightarrow V}$ loss).} We supervise the training of the residual token projection $\mathcal{A}$ via feature distillation. The loss encourages the output features of \model{} ($f^{\mathcal{S}}_{i,t+k}$) to approximate those of the pre-trained ViT encoder ($f^{\mathcal{V}}_{i,t+k}$). 
    }
    \label{fig:model-training}
    \vspace{-.2cm}
\end{figure}

We illustrate the training process in Figure~\ref{fig:model-training}. Let $\mathcal{B}=\left\{\left(\mathcal{X}_i, \ell_i\right)\right\}_{i=1}^B$ be a batch of videos $\mathcal{X}_i$, and their corresponding textual descriptions $\ell_i$. From each video $\mathcal{X}_i$, we decode $N_{\text{Train}}+1$ frames at a constant frame rate starting at time step $t$. These frames are then encoded via the ViT $\mathcal{E}_\mathcal{V}$ (teacher) and \model{} $\mathcal{E}_\mathcal{S}$ (student) and the corresponding features $f^{(\mathcal{V})}_{i,t+k}$ and $f^{(\mathcal{S})}_{i,t+k}$ are output for each time step $t+k$ for $k\in\{1,\dots,N_{\text{Train}}\}$.
Furthermore, let $g\in\mathbb{R}^{d\times B}$ be a matrix of features with dimension $d$ computed from all the textual descriptions $\{\ell_i\}$ in the batch using the language encoder $\mathcal{E}_\mathcal{L}$. 
We aim to train the \model{} encoder to match soft targets, which are the similarities between the teacher's visual features and the language features. This ensures that the distilled model maintains alignment in the joint visual and language space. To achieve this goal, we optimize a cross-entropy loss over the softmax inner product between the vision features $f_{i,t+k}$ and language features $g$,
\begin{equation}
    \label{eq:training}
    \mathcal{J}_{L\rightarrow V}=-\sum_{i=1}^B \sum_{k=1}^{N_{\text{Train}}} \sum_{j=1}^B \sigma_j{\left(g^\intercal f_{i,t+k}^{(\mathcal{V})}\right)}\log\left(\sigma_j{\left(g^\intercal f_{i,t+k}^{(\mathcal{S})}\right)}\right),
\end{equation}
where $\sigma_j(x)=\exp(x_j)/\sum_c \exp(x_c)$ is the $j$-th component of the softmax function of vector $x$. Here, the sum over $c$ in the denominator of the softmax ensures that for a given image feature $f_{i,t+k}$ similarities to all language descriptions $g$ in the batch sum to one, converting them to a probability distribution. The inner sum in Equation~\ref{eq:training} sums over the language descriptions $j$ in the batch; the middle sum adds losses for all the frames $k$ in each video excerpt; and finally, the outer sum sums over all videos $i$ in the batch. Please note that due to the softmax normalization over the language features, the computation is asymmetric. Hence, we also define in an analogous manner a video to language loss $\mathcal{J}_{V\rightarrow L}$ where the sigmoid normalization is over the vision features in the batch.

The final loss is then the sum of the two losses. The overall learning problem is then formulated as the following minimization $\min_\mathcal{A} \left( \mathcal{J}_{L\rightarrow V} +  \mathcal{J}_{V\rightarrow L}\right) $, where $\mathcal{A}$ are the parameters of the residual tokenizer module.          
Please note that this loss not only encourages the visual representation of the two models to be close to each other but also supervises the feature distillation to preserve the joint vision-language space of the original CLIP model as the language features are shared between the teacher ViT encoder and the student \model{} encoder. We optimize the loss over samples from a paired video-language dataset. Please note that as we are learning (distilling) only a small number of parameters of the residual tokenizer $\mathcal{A}$, which is a single linear layer, our learning formulation does not require huge training datasets as in typical distillation set-ups when an entire large model is distilled into another (smaller) model.

\section{Experiments}
\label{sec:experiments}  
\noindent\textbf{Tasks, Datasets, and Settings.}
We evaluate the quality and runtime of our \model{} and compare it to the CLIP model, which represents the upper bound in terms of quality, across four tasks, namely: Natural Language Temporal Video Grounding (NLTVG)~\cite{gao2017tall,anne2017localizing} (both for short form and long form videos), Audio Description (AD) generation~\cite{ma2022x}, Temporal Activity Localization (TAL)~\cite{caba2015activitynet}, and Action Recognition (AR). 
We evaluate our efficient video encoding approach over five diverse datasets: MAD~\cite{soldan2022mad}, Charades-STA~\cite{gao2017tall}, ActivityNet-Captions~\cite{Krishna_2017_ICCV}, ActivityNet-v1.3~\cite{caba2015activitynet}, and Kinetics-400~\cite{kay2017kinetics}.
Moreover, we showcase the quality and versatility of our approach by addressing tasks both in zero-shot and fully supervised settings, highlighting the strong generalization capabilities of our proposed architecture.

\noindent\textbf{Evaluation Metrics.} 
Following~\cite{anne2017localizing,gao2017tall}, grounding accuracy is measured via Recall@$K$ for IoU=$\theta$ with $K{=}1$ and $\theta{\in}\{0.5, 0.7\}$. The temporal localization accuracy is reported using mean Average Precision (mAP). Following the literature~\cite{Xu_2020_CVPR, caba2015activitynet}, we compute the mAP metric with temporal IoU thresholds $\{0.5, 0.55, 0.6, 0.65, 0.7, 0.75, 0.8, 0.85, 0.9, 0.95\}$. 
Following~\cite{ma2022x}, the captioning quality in the AD task is evaluated using ROUGE-L~\cite{lin2004rouge} (\textbf{R-L}), CIDEr~\cite{vedantam2015cider} (\textbf{C}), METEOR~\cite{banerjee2005meteor} (\textbf{M}), SPICE~\cite{anderson2016spice} (\textbf{S}), and BertScore~\cite{zhang2019bertscore} (\textbf{BertS}). For action recognition, we measure Accuracy@$\phi$ with $\phi \in \{1, 5\}$. The computational cost of encoding is measured in GFLOPs, reflecting the average cost per second based on frame rate and cost to encode a single frame.

\noindent\textbf{Implementation Details.} 
We build on the publicly available OpenCLIP~\cite{ilharco_gabriel_2021_5143773} implementation and use the default training parameters and loss with the exceptions noted next. We train our method on video-text pairs from the WebVid-2.5M dataset~\cite{bain2021frozen} for $5$ epochs. Our batch size is $2048$ for ViT-B/32 and ViT-B/16 models and $1536$ for ViT-L/14. We encode one frame using the visual encoder $\mathcal{E}_\mathcal{V}$ and the three subsequent frames ($N_{\text{Train}}{=}3$) with \model{} encoder $\mathcal{E}_\mathcal{S}$. For all backbone sizes, we use a constant learning rate of $0.0005$ while weight decay and warmup are disabled. All model training is performed on 4 V100 GPUs, while inference only requires 1 V100 GPU. At inference time, videos are processed at $3$ frames per second for Charades-STA, $1$ frame per second for ActivityNet-Captions, and $5$ frame per second for MAD. For the AD task, we set the visual temporal context to $8$ clips, each containing $8$ frames, and we do not use any pretraining data nor AD context.  
For TAL, we train ActionFormer~\cite{zhang2022actionformer} from scratch using OpenTAD~\cite{liu2025opentad} implementation with default hyperparameters.
We measure GFLOPs via the fvcore library~\cite{fvcore}.

\subsection{Ablation Study}
\label{sec:experiments-ablations}
In this section, we perform multiple ablations to assess the impact of our design choices. We report results for the NLTVG task on the Charades-STA dataset using the ViT-L/14 backbone. When token reduction is used, we employ the motion-based strategy (see details in Section~\ref{sec:supplementary-drop-strategy}
of supplementary) with probability $p{=}85\%$. For all experiments that interleave frames, we set $N{=}2$. 
Seven additional ablations are detailed in Section~\ref{sec:supplementary-ablations} of supplementary. These ablations investigate the (i) token drop strategy for token reduction (\ie, random, uniform, center, motion-based), (ii) token drop probability, (iii) adoption of token merging for token reduction, (iv) impact of frame resolution reduction as an alternative to token reduction and (v) replacement of the distillation objective, (vi) frame rate during inference, and (vii) interleave factor at training time. Additionally, we measure the runtime reduction in Section~\ref{sec:supplementary-time-efficiency} of supplementary.
Below, we include the ablations of our model architecture key components, and the interleave factor. 

\noindent\textbf{Architecture Ablation.} In Table~\ref{tab:architecture-ablation}, we analyze the contribution of the main architecture components of our model to downstream accuracy for the Natural Language Temporal Video Grounding task. We select this task for our ablations as it provides the ideal testbed for \model{}. First, the necessity for fine-grained temporal resolution demands dense extraction of visual features from large volumes of video data. Second, it requires language and visual
understanding to enable querying the model through natural language queries. Both requirements are fulfilled by \model{}.

\begin{table}[!t]
\centering
\vspace{-.5cm}
\resizebox{\linewidth}{!}{%
    \scalebox{1.0}{
        \begin{tabular}{lccc|cc|c}
            \toprule\toprule
            &&&Residual& \multicolumn{2}{c|}{Charades-STA} & Avg. Cost \\
            &Token & Interleave & Tokenizer & \multicolumn{2}{c|}{R@1 $\uparrow$} & Feature/sec $\downarrow$ \\
            &Reduction& Factor& (Distilled) & IoU=0.5 & IoU=0.7 & (GFLOPs) \\ 
            \midrule\midrule
            \textbf{a.} &&&& \color{gray} $42.9$ & \color{gray} $24.1$ & \color{gray} $233.4$\\
            \textbf{b.} &\cmark &&& $28.5$ & $14.5$ & $35.7_{(-85\%)}$\\
            \textbf{c.} &\cmark &\cmark && $38.9$ & $22.8$ & $102.0_{(-56\%)}$\\
            \textbf{d.} &\cmark &\cmark &\cmark & $41.5$ & $23.8$ & $102.6_{(-56\%)}$\\
            \bottomrule\bottomrule
        \end{tabular}
    }
}
\vspace{-.1cm}
\caption{{\bf Architecture ablation.} We ablate the main components of our architecture: the token reduction module, the interleave factor, and the distilled residual tokenizer. We set the token reduction probability $p$ to $85\%$, $N=2$, and use the ViT-L/14 backbone. }
\label{tab:architecture-ablation}
\vspace{-.3cm}
\end{table}

With an average frame encoding cost of $233.4$ GFLOPs, the CLIP baseline ({\bf a.} in Table~\ref{tab:architecture-ablation}) establishes our upper bound target grounding accuracy, which we wish to maintain but with much higher computational efficiency. 
When we apply token reduction across all frames ({\bf b.}), we observe an $85\%$ decrease in computational cost. However, this setting induces marked absolute declines in grounding accuracy of $14.4\%$ and $9.6\%$ in our metrics, which translates to a relative drop of $34{-}40\%$. 
The introduction of our interleave strategy ({\bf c.}), which alternates encoding one frame without token reduction and $N$ frames with token reduction (where $N{=}2$), shows an increase in grounding accuracy of $10.4\%$ and $8.3\%$ while only using $44\%$ of the original computational budget, which is a first step in closing the grounding accuracy gap with respect to the target CLIP accuracy. Compared to the full CLIP model, the grounding accuracy drop narrows to a modest $4$ and $1.3$ percentage points ($5{-}9\%$ relative drop), yet this configuration only incurs $44\%$ of the original computational cost. Further, adding the residual tokenizer learned via distillation ({\bf d.}) comes at a negligible compute cost but further boosts grounding accuracy closer to the target CLIP model, showing only a minor ${\sim}1\%$ absolute drop. 

\begin{figure}[!t]
    \centering
    \vspace{-.5cm}
    \includegraphics[trim={0cm 0cm 0cm 0cm},width=\linewidth,clip]{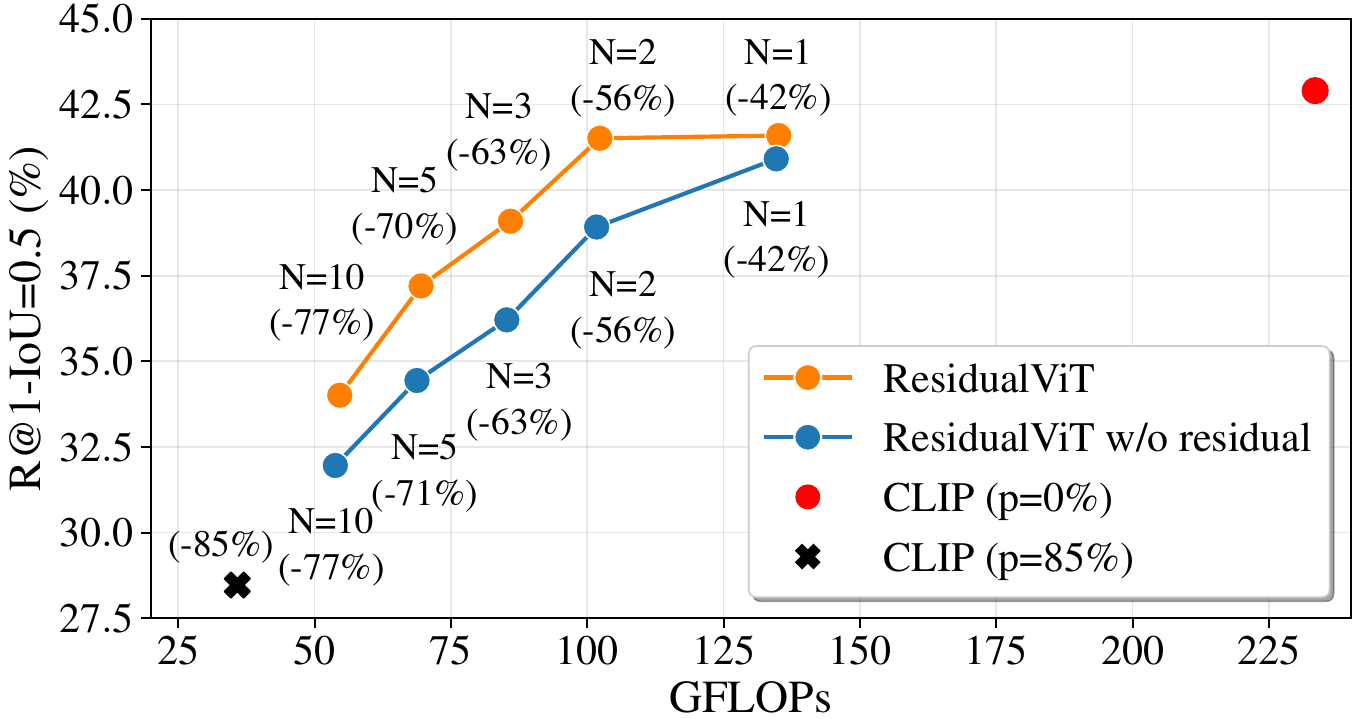}
    \vspace{-.5cm}
    \caption{\textbf{Interleaving frames ($N$)}. Our \model{} (\textcolor{orange}{\textbf{orange}}) closely retains CLIP's (\textcolor{red}{\textbf{red}}) performance for $N{=}1$ and $2$ while reducing cost by $56\%$.}
    \label{fig:N-ablation}
    \vspace{-.3cm}
\end{figure}

\noindent\textbf{Interleave Factor $N$ and Benefits of Distillation.} In Figure~\ref{fig:N-ablation}, we explore the relationship between grounding accuracy and computational cost as we vary the number of interleaved frames ($N$). 
Here, the baseline CLIP model is shown in \textcolor{red}{\textbf{red}}, while our \model{}, applied with and without the distilled residual tokenizer module, is shown in \textcolor{orange}{\textbf{orange}} and \textcolor{navyblue}{\textbf{blue}}, respectively. We vary $N \in \{1,2,3,5,10\}$. 
When setting $N{=}1$, grounding accuracy is marginally impacted, yet a large computational cost reduction is already achieved ($42\%$). Notably, we see a further accuracy drop when the residual tokenizer is removed (\textcolor{navyblue}{\textbf{blue}}), demonstrating the importance of the distillation training. At $N{=}2$, the cost savings increases to $56\%$ with virtually no accuracy change for \model{}. However, the importance of the learnable residual connection (via the residual token learnt by the distillation training) becomes more evident as the difference between the two configurations widens with substantial drops when the residual token is not employed. Increasing $N$ beyond this point ($N=2$)  sees diminishing returns in cost savings, now at $63\%$, and a noticeable decrease in accuracy. This decline is attributed to the growing temporal gap between I-features and P-features, leading to a weakened visual correlation and, thus, reduced efficacy. We regard $N{=}2$ as the best trade-off.

\subsection{Benchmarking on Downstream Tasks}
\label{sec:experiments-quantitative}
In this section, we compare the quality vs.\ runtime trade-offs achieved by \model{} with respect to  CLIP. We do so on three benchmarks for the multi-modal Natural Language Temporal Video grounding task, comprising two short-form video datasets (Charades-STA and ActivityNet-Captions) and one long-form video dataset (MAD). Additionally, we also explore \model{}'s generalization capabilities by applying it to the complementary task of Automatic Audio Description generation~\cite{han2023autoad}, Temporal Activity Localization~\cite{caba2015activitynet} and Action Recognition (see Section~\ref{sec:kinetics} of supplementary). In all experiments, unless specified, we use \model{} ($N{=}2$, $p{=}85\%$), which yields a cost saving of $53\%-56\%$.

\begin{table}[!t]
\vspace{-.5cm}
\centering 
\setlength{\tabcolsep}{2pt}
\resizebox{\linewidth}{!}{
\scalebox{1.0}{
\begin{tabular}{l|ccc|ccc} 

\toprule
\toprule

&\multicolumn{2}{c}{Charades-STA}
& Avg. Cost
& \multicolumn{2}{c}{A.Net-Captions} 
& Avg. Cost \\

& \multicolumn{2}{c}{R@1 $\uparrow$} & Feature/sec $\downarrow$ 
& \multicolumn{2}{c}{R@1 $\uparrow$} & Feature/sec $\downarrow$\\

& IoU=0.5 & IoU=0.7 & (GFLOPs)
& IoU=0.5 & IoU=0.7 & (GFLOPs) \\ 

\midrule
\midrule

UniVTG~\cite{lin2023univtg} 
& $25.2$ & $10.0$ & $70.0$  
& $-$ & $-$ & $-$ \\ 

MR-FVLM~\cite{luo2024zero} 
& $\mathbf{42.9}$ & $20.1$ & $1370.0$  
& $27.9$ & $11.6$ & $370.0$ \\  

\midrule
\midrule

 CLIP (B/32)
& $35.9$ & $18.7$ & $\underline{13.2}$ 
& $27.8$ & $\mathbf{13.9}$ & $\underline{4.4}$ \\ 

\model{} (B/32)
& $34.2$ & $17.7$ & $\mathbf{6.1}_{(-53\%)}$ 
& $27.3$ & $13.7$ & $\mathbf{2.0}_{(-53\%)}$ \\ 

 CLIP  (B/16)
& $37.7$ & $21.2$ & $50.7$  
& $28.1$ & $\underline{13.8}$ & $16.9$ \\ 

\model{} (B/16)
& $37.8$ & $21.0$ & $22.4_{(-56\%)}$ 
& $27.5$ & $\underline{13.8}$ & $7.5_{(-56\%)}$ \\ 

 CLIP (L/14)
& $\mathbf{42.9}$ & $\mathbf{24.1}$ & $233.4$ 
& $\mathbf{29.1}$ & $\underline{13.8}$ & $77.8$ \\

\model{} (L/14)
& $\underline{41.5}$ & $\underline{23.8}$ & $102.6_{(-56\%)}$ 
& $\underline{28.3}$ & $13.5$ & $34.2_{(-56\%)}$ \\

\bottomrule
\bottomrule
\end{tabular}
}
}
\vspace{-.1cm}
\caption{\label{tab:nltvg-short}
{\bf NLTVG in short videos.} Our zero-shot grounding algorithm (Section~\ref{sec:grounding-algorithm} of supplementary) equipped with CLIP features, achieves new state-of-the-art accuracy (CLIP L/14). When compared against CLIP features, \model{} (L/14) reduces the cost of frame encoding by $56\%$ with minimal accuracy degradation. Other backbones (B/16 and B/32) follow a similar trend.}
\vspace{-.5cm}
\end{table}

\noindent\textbf{NLTVG in Short Videos.}  
Table~\ref{tab:nltvg-short} presents the main results for the Natural Language Temporal Video Grounding (NLTVG) task in short videos. Here, we focus on the zero-shot setup, comparing our approach against the only two available baselines: UniVTG~\cite{lin2023univtg} and MR-FVLM~\cite{luo2024zero}. Additionally, we introduce a new zero-shot grounding baseline, which utilizes either CLIP~\cite{radford2019language} or \model{} features. Further details on this baseline are provided in Section~\ref{sec:grounding-algorithm} of the supplementary material. 

Beyond zero-shot baselines, additional results can be found in the supplementary material. Section~\ref{sec:supplementary-short-nltvg} of supplementary expands Table~\ref{tab:nltvg-short} with comparisons against fully supervised, weakly supervised, and pseudo-supervised methods. Furthermore, in Section~\ref{sec:cgdetr} (suppl.), we take CLIP and \model{} features and train a fully supervised grounding model using CG-DETR~\cite{moon2023correlation} as the grounding model.  

For each method, Table~\ref{tab:nltvg-short} reports grounding accuracy on Charades-STA~\cite{gao2017tall} and ActivityNet-Captions~\cite{Krishna_2017_ICCV}, along with the average feature embedding cost per second. Our results show that \model{} reduces frame encoding costs by $53-56\%$ across all ViT backbones while maintaining high accuracy. On Charades-STA, \model{} exhibits a negligible accuracy drop with respect to CLIP with ViT-B/16, while for ViT-B/32 and L/14, the accuracy decline is limited to 1\%-1.5\%. On ActivityNet-Captions, \model{} remains on par with CLIP-based approaches while achieving significant computational savings, with an accuracy decrease of less than 1\% across all configurations. Compared to zero-shot methods, UniVTG underperforms across all metrics, while MR-FVLM achieves comparable accuracy at IoU=0.5 on Charades-STA but incurs a substantially higher computational cost (1337 GFLOPs per feature vs. 102.6 GFLOPs for \model{}), due to its reliance on the C3D backbone and InternVideo-MM-L-14~\cite{wang2022internvideo}. These findings confirm that \model{} offers a good balance between accuracy and computational efficiency, effectively achieving its design goal.

\begin{table}[!t]
\vspace{-.5cm}
\centering  
\resizebox{.9\linewidth}{!}{
    \scalebox{1.0}{
        \begin{tabular}{l|cccc}
        \toprule\toprule
        & \multicolumn{3}{c}{MAD} 
        & Avg. Cost\\
        
        & \multicolumn{3}{c}{R@1 $\uparrow$} 
        & Feature/sec $\downarrow$ \\
        
        & IoU=0.1 & IoU=0.3 & IoU=0.5 & (GFLOPs) \\ 
        
        \midrule \midrule
        
        CLIP (B/32)~\cite{soldan2022mad}
        & $6.6$ & $3.1$ & $1.4$ & $\underline{21.8}$  \\
        
        \midrule \midrule
        
        CLIP (B/32)
        & $8.7$  & $5.5$ & $3.2$ & $\underline{21.8}$  \\ 
        
        \model{} (B/32) 
        & $8.6$  & $5.4$ & $3.1$ & $\mathbf{10.2}_{(-53\%)}$  \\ 
        
        CLIP (B/16)
        & $\underline{10.8}$ & $6.8$ & $3.9$ & $84.3$  \\
        
        \model{} (B/16) 
        & $10.1$ & $6.4$ & $3.7$ & $37.3_{(-56\%)}$  \\
        
        CLIP (L/14)
        & $\mathbf{13.3}$ & $\mathbf{8.6}$ & $\mathbf{5.0}$ & $389.2$ \\
        
        \model{} (L/14) 
        & $10.7$ & $\underline{7.3}$ & $\underline{4.3}$ & $171.0_{(-56\%)}$ \\
        \bottomrule\bottomrule
        \end{tabular}
    }
}
\vspace{-.1cm}
\captionof{table}{\label{tab:sota-mad}{\bf NLTVG in long videos.} 
Our zero-shot grounding algorithm equipped with \model{} features outperforms the previous art both in accuracy and computational cost on the challenging long-form MAD dataset. 
}
\vspace{-.5cm}
\end{table}
\noindent\textbf{NLTVG in Long Videos.} Table~\ref{tab:sota-mad} presents additional results on the challenging long-form video MAD dataset~\cite{soldan2022mad}, contrasting our \model{} against the only zero-shot grounding baseline available, which is described in~\cite{soldan2022mad}. This existing zero-shot grounding algorithm employs a proposal-based approach, utilizing a multi-scale sliding window technique to generate potential video segment proposals.  For each proposal, a single feature representation is computed by average pooling frame features whose temporal locations fall within the proposal's span. Finally, cosine similarity is computed between each sentence feature representation and each proposal feature representation. In contrast, our grounding algorithm requires the number of similarity computations to be equal to the number of encoded frames, which significantly reduces the computational complexity. Specifically, the proposal-based method demands approximately $20\times$ more similarity computations compared to our zero-shot grounding approach (described in Section~\ref{sec:grounding-algorithm} of supplementary). In these experiments, \model{} was configured with $N{=}2$, a token dropping probability $p{=}85\%$, and the center token dropping strategy.

Our results demonstrate that the grounding algorithm combined with \model{} visual features significantly outperforms the existing state-of-the-art (row 1 vs row 3 of Table~\ref{tab:sota-mad}). When using the same backbone (ViT B/32), our approach achieves relative improvements ranging from $43\%$ at IoU=0.1 to $128\%$ at IoU=0.5, while also being $53\%$ more efficient in feature extraction and requiring one order of magnitude fewer similarity computations. Additionally, accuracy consistently increases with the use of more computationally demanding backbones. For example, using the ViT B/16 backbone, our method achieves a $160\%$ increase in accuracy at IoU=0.5, despite a $73\%$ higher feature extraction cost compared to the baseline~\cite{soldan2022mad}. These findings highlight an excellent tradeoff between computational cost and improved accuracy. Section~\ref{sec:supplementary-mad} 
of supplementary expands Table~\ref{tab:sota-mad} with comparisons against fully supervised methods and additional metrics. 

\noindent\textbf{Automatic Audio Description.}  
We demonstrate \model{}'s generalization by using its visual representations in an additional long-video understanding task: Automatic Audio Description~\cite{han2023autoad}. This challenging task, similar to dense video captioning, requires generating textual descriptions of video moments that capture events and characters.
\begin{table}[!t]
\centering  
\vspace{-.5cm}
\resizebox{.9\linewidth}{!}{%
    \scalebox{1.0}{
        \begin{tabular}{l|cccccc} 
        \toprule\toprule
        &&&&&& Avg. Cost\\
        
        & BertS $\uparrow$ 
        & R-L $\uparrow$ 
        & C $\uparrow$ 
        & M $\uparrow$ 
        & S $\uparrow$ 
        & Feature/sec $\downarrow$ \\
        
        &&&&&& (GFLOPs) \\ 
        
        \midrule \midrule
        
        CLIP (B/32)            
        & $21.2$  
        & $10.8$ 
        & $8.4$ 
        & $4.4$
        & $2.2$ 
        & $21.8$ \\

        ResidualViT (B/32)     
        & $21.4$  
        & $10.7$ 
        & $8.2$ 
        & $4.5$
        & $2.1$ 
        & $10.2_{(-53\%)}$ \\
        
        CLIP (B/16)            
        & $22.0$  
        & $11.0$ 
        & $9.7$ 
        & $4.4$
        & $2.4$  
        & $84.3$\\

        ResidualViT (B/16)     
        & $22.1$ 
        & $10.9$ 
        & $9.2$ 
        & $4.5$
        & $2.3$ 
        & $37.3_{(-56\%)}$ \\
        
        CLIP (L/14)            
        & $\mathbf{23.2}$  
        & $\mathbf{11.4}$ 
        & $\mathbf{11.3}$ 
        & $\underline{4.6}$
        & $\mathbf{3.0}$ 
        & $389.2$\\
        
        ResidualViT (L/14)     
        & $\underline{22.7}$  
        & $\underline{11.3}$ 
        & $\underline{10.6}$ 
        & $\mathbf{4.7}$
        & $\underline{2.7}$ 
        & $171.0_{(-56\%)}$ \\ 
        \bottomrule\bottomrule
        \end{tabular}
    }
}
\vspace{-.1cm}
\captionof{table}{\label{tab:ad}{\bf Automatic Audio Description.}
\model{} provides nearly identical captioning quality to CLIP across all backbone sizes and metrics at a significantly reduced encoding cost.}
\vspace{-.3cm}
\end{table}
\begin{table}[!t]
\setlength{\tabcolsep}{3pt}
\centering  
\resizebox{\linewidth}{!}{%
    \scalebox{1.0}{
        \begin{tabular}{l|ccc|ccc} 
        \toprule\toprule
        &&& Avg. Cost 
        &&& Avg. Cost\\
        
        & Backbone 
        & mAP $\uparrow$ 
        & Feature/sec $\downarrow$ 
        & Backbone 
        & mAP $\uparrow$ & Feature/sec $\downarrow$\\
        
        && (\%) & (GFLOPs) 
        && (\%) & (GFLOPs) \\ 
        
        \midrule \midrule
        
        \multirow{2}{*}{Action} 
        & CLIP (B/32)     & $34.05$ & $4.4$ 
        & \model{} (B/32) & $33.42$ & $2.0_{(-53\%)}$  \\
        
        \multirow{2}{*}{Former}
        & CLIP (B/16)     & $34.40$ & $16.9$ 
        & \model{} (B/16) & $33.76$ & $7.5_{(-56\%)}$  \\
        
        & CLIP (L/14)     & $\mathbf{34.83}$ & $77.8$ 
        & \model{} (L/14) & $\underline{34.46}$ & $34.2_{(-56\%)}$  \\
        
        \bottomrule\bottomrule
        \end{tabular}
    }
}
\vspace{-.1cm}
\caption{{\bf Temporal Action Localization.} \model{} provides competitive accuracy with lower computational cost.}
\label{tab:tal}
\vspace{-.5cm}
\end{table}
The method proposed in~\cite{han2023autoad} adopts a video-LLM model by integrating CLIP~\cite{radford2021learning} for visual feature extraction and GPT-2~\cite{radford2019language} for caption generation, using a learned transformer encoder to align vision and language features. We reimplement this approach, training the baseline from scratch while replacing the input visual features with either CLIP or \model{} representations. 
Following AutoAD evaluation protocol, Table~\ref{tab:ad} compares CLIP and ResidualViT representations
on the \textit{MAD-eval Named} subset of MAD (see Section 4 of~\cite{han2023autoad}). 
Across all backbone architectures, ResidualViT consistently keeps pace with CLIP’s quality while significantly reducing computational cost ($53\%-56\%$). In particular, \model{} shows only a relative average accuracy drop of $0.9\%$, $1.5\%$, and $3.4\%$ for backbones B/32, B/16, and L/14, respectively. These findings highlight \model{}'s applicability to video captioning, a complementary task to NLTVG, further demonstrating its flexibility and generalization capabilities. More results can be found in Section~\ref{sec:supplementary-ad} of supplementary.

\noindent\textbf{Temporal Activity Localization (TAL).}  We evaluate \model{} on the uni-modal, temporally dense TAL task using the ActivityNet-v1.3 dataset~\cite{caba2015activitynet}. As a baseline, we select ActionFormer~\cite{zhang2022actionformer}, a recent high-performing model. The baseline is trained from scratch for each feature set.  
Table~\ref{tab:tal} compares the mAP achieved with \model{} features, which closely follow CLIP features while significantly reducing encoding cost. Notably, \model{} operates at only $44\%$–$47\%$ of CLIP’s computational cost while maintaining highly competitive accuracy with an absolute mAP drop of $0.4$-$0.6$, further validating its efficiency. Additional comparisons can be found in Section~\ref{sec:supplementary-tal} of supplementary. 

\subsection{Qualitative Results for NLTVG}
\begin{figure}[!t]
    \vspace{-.5cm}
    \centering
    \includegraphics[trim={0cm 0cm 0cm 0cm},width=\linewidth,clip]{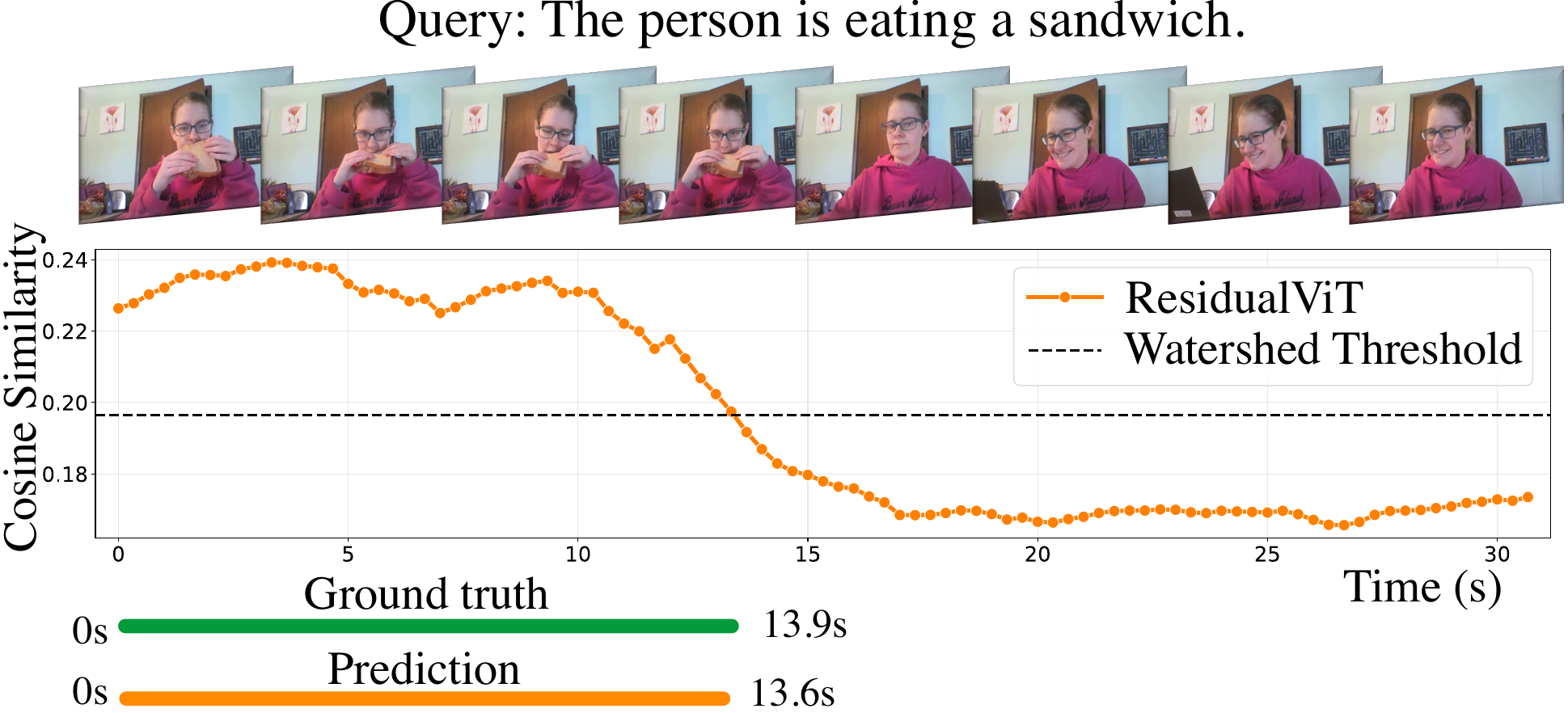}
    \vspace{-.5cm}
    \caption{\textbf{Qualitative results.} We present a qualitative example in which our zero-shot algorithm using our ResidualViT encoder can correctly ground the sentence in the video. We showcase the comparison between the ground truth annotation (\textcolor{GTgreen}{\textbf{green}}) and our top${-}$1 prediction (\textcolor{orange}{\textbf{orange}}). 
    \label{fig:qualitative}}
    \vspace{-.3cm}
\end{figure}

Figure~\ref{fig:qualitative} presents a qualitative example of temporal grounding from the Charades-STA dataset, where our zero-shot grounding algorithm (described in  Section~\ref{sec:grounding-algorithm} of supplementary) accurately predicts the temporal span corresponding to the textual query, ``the person is eating a sandwich''. This prediction is driven by the cosine similarity profile between the visual and language features, along with the watershed threshold, as illustrated in the figure. Additional visualizations are available in Section~\ref{sec:supplementary-qualitative} of supplementary.

\section{Conclusion}
\label{sec:conclusions}
We have developed a new approach for the efficient computation of frame-based video features, exploiting temporal redundancy in videos via learnable temporal residual connections. The proposed approach is lightweight, as it trains only a small number of parameters in the residual tokenizer while keeping the transformer frozen. We have demonstrated the benefits of the proposed approach on four tasks and five datasets, achieving a significant reduction (up to 60\%) in compute cost with only a marginal accuracy drop. 
This efficiency gain is particularly impactful for practical large-scale applications that require processing millions of videos, effectively reducing video encoding costs by {\bf half}.
Additionally, our work opens up the possibility of extending the distillation objective to incorporate richer interactions between visual and language representations, as well as exploring additional large-scale pre-trained models that natively model temporal relationships or generate videos.

\noindent\textbf{Acknowledgements.} The research reported in this publication was supported by funding from King Abdullah University of Science and Technology (KAUST) - Center of Excellence for Generative AI, under award number 5940.

\twocolumn[{
\vspace{1cm}
\centering
\LARGE\bfseries Supplementary Material
\vspace{1cm}
}]

\setcounter{section}{0}

\noindent We provide the following additional information:
\begin{enumerate}

\item \textbf{Token Dropping Strategies:} We present the different token dropping strategies that can be adopted in the token reduction module in Section~\ref{sec:supplementary-drop-strategy}.

\item \textbf{Motion-based Token-dropping Strategy:} Insights into the motion-based token-dropping strategy are provided in Section~\ref{sec:supplementary-motion-details}, explaining the additional RAM requirements and the pre-processing of raw motion vectors.
    
\item \textbf{Zero-shot Grounding Algorithm:} A thorough explanation of the implementation details of the zero-shot temporal grounding algorithm is presented in Section~\ref{sec:grounding-algorithm}.

\item \textbf{Additional Comparison for Short-Form NLTVG:} Additional analysis for the task of natural language temporal video grounding on the Charades-STA and ActivityNet-Caption datasets are available in Section~\ref{sec:supplementary-short-nltvg}.

\item \textbf{Feature Comparison under Full Supervision Setup:} As an additional test of the quality of our \model{} features, we investigate the accuracy of CG-DETR~\cite{moon2023correlation} when replacing the original CLIP features with our \model{} ones in Section~\ref{sec:cgdetr}.

\item \textbf{Additional Comparison for Long-Form NLTVG:} Additional analysis for the task of natural language temporal video grounding on the MAD dataset is available in Section~\ref{sec:supplementary-mad}.

\item \textbf{Additional Comparison for AD:} Additional results for the task of automatic audio descriptions on the MAD dataset are available in Section~\ref{sec:supplementary-ad}.

\item \textbf{Additional Comparison for TAL:} Additional results for the task of temporal action localization on the ActivityNet-v1.3 dataset are available in Section~\ref{sec:supplementary-tal}.

\item \textbf{Supplementary Ablations:} In Section~\ref{sec:supplementary-ablations}, we conduct additional ablations for \model{} on the Charades-STA dataset, exploring different token reductions strategies as presented in Section~\ref{sec:supplementary-drop-strategy} and discussing the role of token-dropping probability. Additionally, we investigate two distinct strategies for computational savings: token merging and reduction of the spatial resolution of the input frames. We ablated the design of the distillation approach and showcased how different distillation objectives can achieve competitive accuracy. Moreover, we ablate the interleave factor during distillation training. Finally, we report two additional ablations on the MAD dataset, investigating the main components and the interleave factor N.

\item \textbf{Video Encoding Latency:} Section~\ref{sec:supplementary-time-efficiency} empirically validates the wall-clock timings of \model{}, demonstrating significant time savings compared to a standard ViT model, despite requiring two forward passes. 

\item \textbf{Additional task - Action Recognition:} In this supplementary experiment, we test the accuracy of CLIP features against \model{} features on the task of action recognition on the Kinetics 400 dataset. Results are reported in Section~\ref{sec:kinetics}.

\item \textbf{Limitations and Discussion:} In Section~\ref{sec:limitations} we discuss the inherent limitations of our solution. 

\item \textbf{Qualitative Results:} We conclude with a showcase of several qualitative results in Section~\ref{sec:supplementary-qualitative}, highlighting the practical effectiveness of our approach.
\end{enumerate}

\begin{figure}[!b]
    \centering
    \begin{subfigure}[b]{0.23\textwidth}
        \centering
        \includegraphics[width=\textwidth]{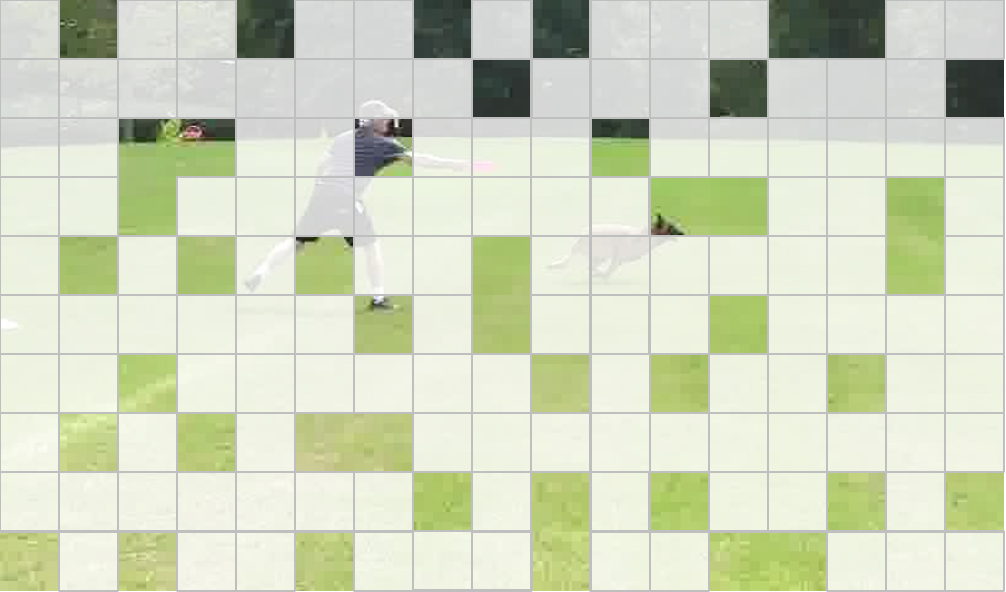}
        \caption{Random.}
        \label{fig:random-token-drop}
    \end{subfigure}
    \hfill
    \begin{subfigure}[b]{0.23\textwidth}
        \centering
        \includegraphics[width=\textwidth]{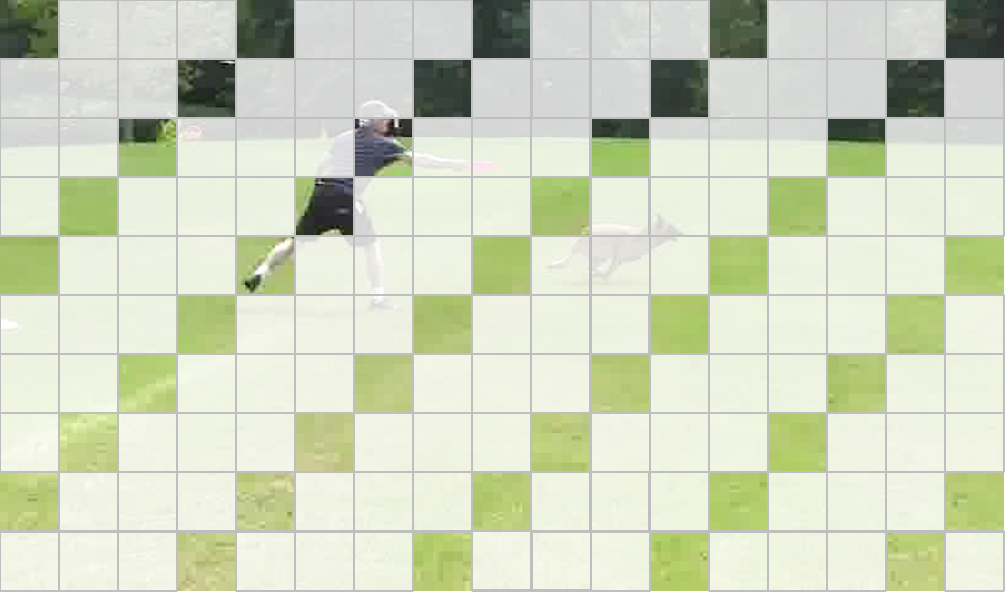}
        \caption{Uniform.}
        \label{fig:uniform-token-drop}
    \end{subfigure}
    \hfill
    \begin{subfigure}[b]{0.23\textwidth}
        \centering
        \includegraphics[width=\textwidth]{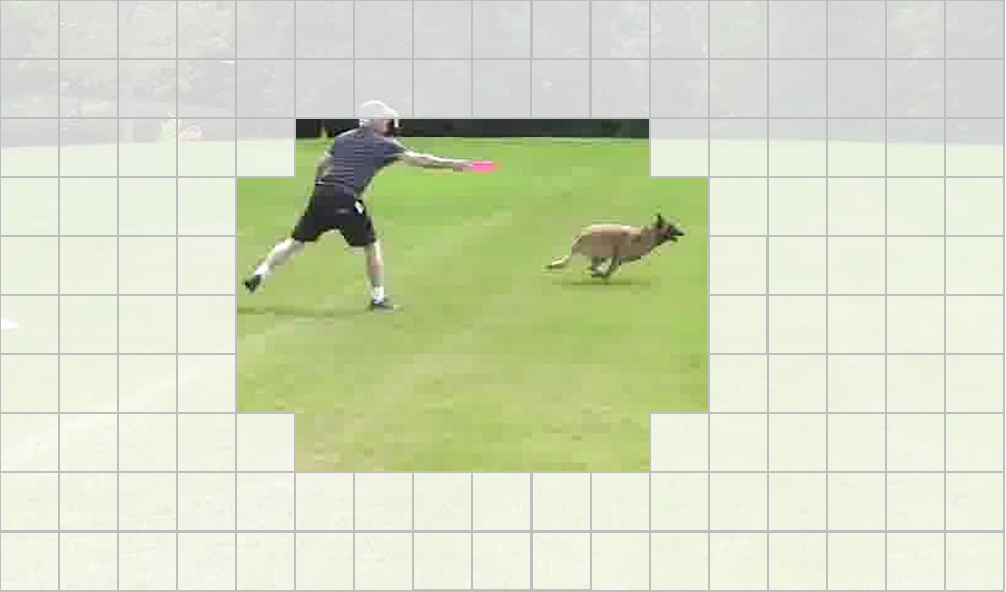}
        \caption{Center.}
        \label{fig:center-token-drop}
    \end{subfigure}
    \hfill
    \begin{subfigure}[b]{0.23\textwidth}
        \centering
        \includegraphics[width=\textwidth]{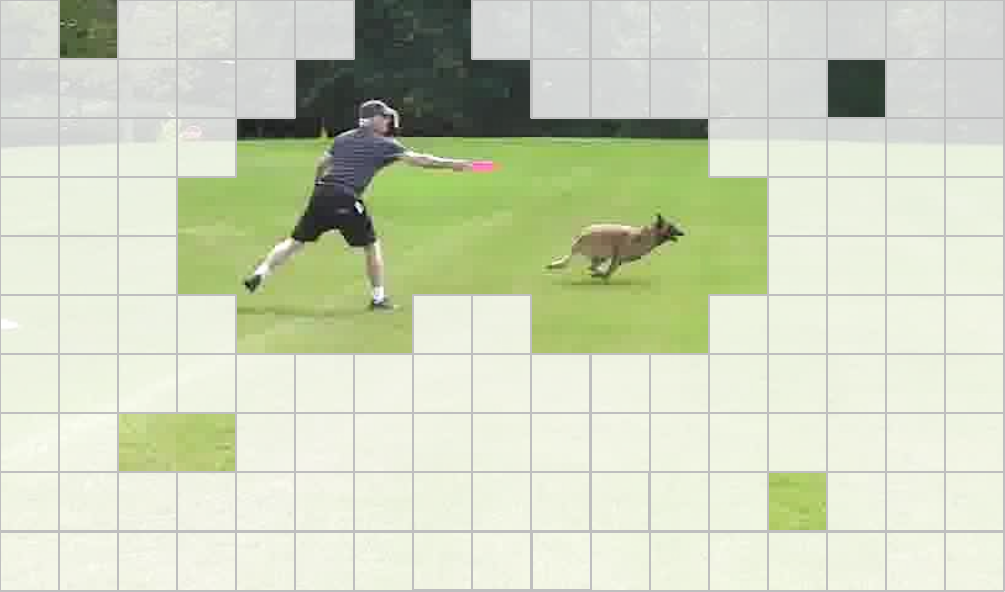}
        \caption{Motion.}
        \label{fig:motion-token-drop}
    \end{subfigure}
    \caption{\textbf{Token reduction strategies.} We implement three data-independent token reduction strategies (a-c) and one data-dependent one (d). }
    \label{fig:token-drop-strategies}
\end{figure}

\section{Token Dropping Strategies}
\label{sec:supplementary-drop-strategy}
In Section~\ref{sec:residualvit-architecture} we introduced the \model{} architecture, which consists of the token reduction module ($\mathcal{R}$), the residual tokenizer ($\mathcal{A}$), and the transformer encoder ($\mathcal{E}_\mathcal{V}$). Here, we explore four practical implementations of the token reduction module when adopting the token dropping strategy~\cite{ding2023prune, haurum2023tokens, hou2022token, liu2023patchdropout}.

For a given frame $x_{t}$, which is transformed into a set of tokens $\mathcal{T}$, each strategy retains $(1-p) \times |\mathcal{T}|$ tokens, where $p$ is the token reduction probability. Figure~\ref{fig:token-drop-strategies} visually depicts the four token reduction approaches we investigate.
(i) The \textbf{random} strategy randomly samples tokens from the set. 
Conversely, (ii) the \textbf{uniform} strategy selects tokens from patches that are evenly distributed across the 2D grid of image patches, ensuring that the selected patches are spaced at regular intervals throughout the frame.
(iii) The \textbf{center} strategy is designed to retain tokens of patches from the center of the frame. This strategy takes into consideration that, when shooting a video, we tend to center the frame around the subject or action being recorded. Finally, we design a data-dependent (iv) \textbf{motion} strategy. This strategy further exploits the characteristics of video data, which describes how characters, objects, and scenes evolve in time. We argue that motion is a valuable source of information readily accessible from encoded video files, providing information on which parts of the frame at time step $t+k$ differ from the frame at time step $t$. Consequently, we discard tokens representing patches with minimal motion, assuming their change relative to previous frames is negligible, and their information can be recovered through the residual token. 
Notably, by prioritizing tokens associated with regions of higher motion, \model{} is well-suited to handle fast-moving content.
See Section~\ref{sec:supplementary-motion-details} for additional details about motion preprocessing and memory overhead.

\section{Implementation Details for Motion-Based Token Reduction Strategy} 
\label{sec:supplementary-motion-details}
Motion is a valuable and readily available source of information for determining which spatial regions of a frame have changed with respect to the previous one. To harness this information, our method employs a compressed video reader~\cite{compressedvideoreader} that extracts motion vectors directly from compressed video streams.
Nevertheless, it is important to acknowledge that motion vectors extracted from raw video data typically exhibit a moderate level of noise, attributable to the inherent sparsity and optimization mechanisms of standard video compression techniques.
To counteract this effect and derive a more reliable motion estimation, we compute the average motion across a short temporal window surrounding a target frame $x_t$.
Specifically, we construct a set of motion vectors $M_v = \{m_i\}_{i=t-W_M/2}^{t+W_M/2}$, where each vector $m_i \in \mathbb{R}^{H' \times W' \times C'}$ corresponds to the motion information of frame $i$. Here, $H' = H/4$ and $W' = W/4$ are the reduced height and width dimensions, respectively, and $C' = 4$ signifies the channels in the motion vector, capturing the $(\Delta x, \Delta y)$ displacement of pixels with respect to adjacent frames (previous and following ones). The parameter $W_M$ denotes the size of the temporal window over which the motion is aggregated. As we are interested in the magnitude of the motion and not its direction, we compute the average $L_1$ norm along dimension $C'$ in the window $W_M$. Note that, at the start of the video ($t < W_M$) and at the end ($t > T - W_M$, where $T$ is the timestamp of the last frame), the window is reduced so that only the available motion vectors are aggregated, avoiding the need for padding.

\begin{figure}[!t]
    \includegraphics[trim={0cm 0cm 0cm 0cm},width=.48\textwidth,clip]{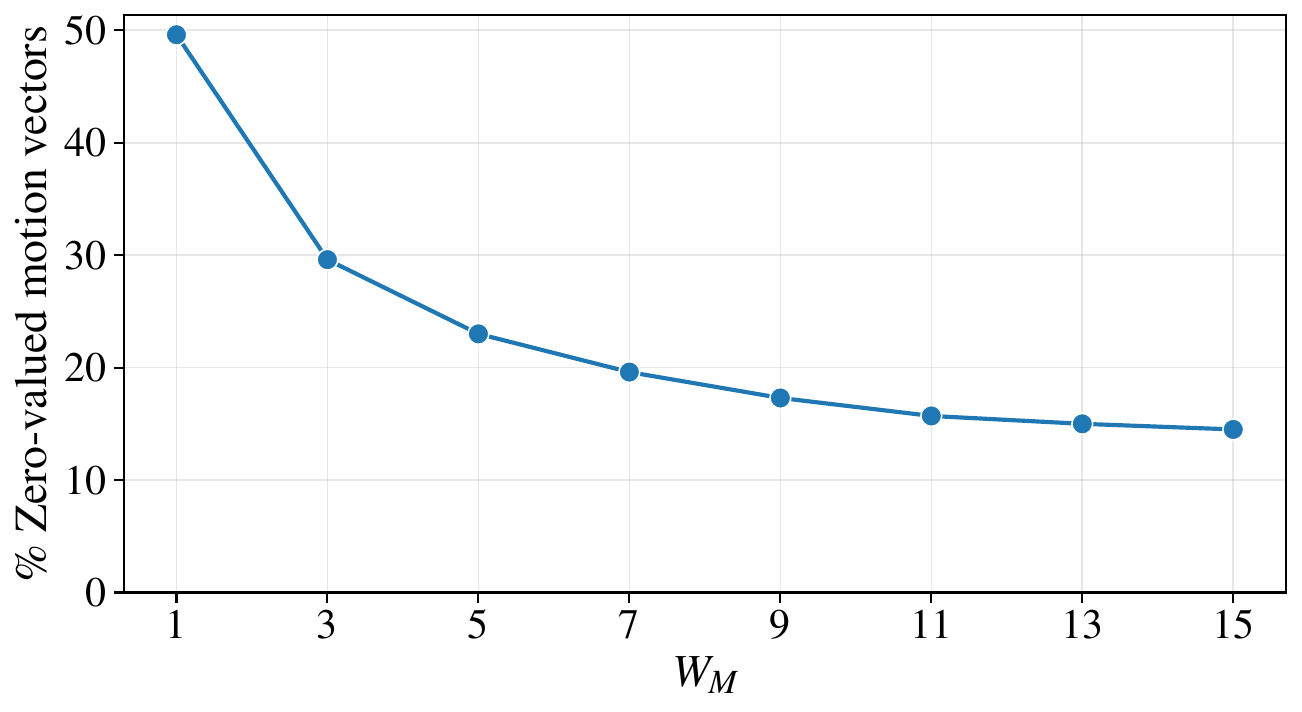}
    \caption{\textbf{Optimal $W_M$ hyperparameter value.} The plot shows the average percentage of zero-valued motion vectors on the Charades-STA dataset as the aggregation window size $W_M$ varies. The trend flattens beyond $W_M=11$, indicating diminishing returns. Therefore, we choose $W_M=11$ as our default parameter.}
    \label{fig:optimal-motion-window}
\end{figure}
We then upsample the computed motion magnitudes to the frame resolution ($H$, $W$) and select the $1-p$ frame tokens with the highest motion magnitudes at their patch's spatial location.

In our implementation, we set the motion window size $W_M = 11$. This setting implies that we incur an additional RAM memory consumption that is proportional to the cost of storing a frame in memory. We can estimate the memory cost as follows. The memory consumption of a frame can be expressed as $M_F{=}H{\times}W{\times}3$, while for motion vectors $M_{Mv}{=}(H/4){\times}(W/4){\times}4{\times}M_v$, resulting in a total memory cost $M_{F} + M_{Mv} / M_{F} ={\sim}1.9{\times}$. Note that this memory overhead does not affect GPU memory availability as the motion vectors are not required to be moved to such a device for processing. 
To determine the value of $W_M$, we measure the average percentage of zero-value motion vectors in the Charades-STA dataset. As shown in Figure~\ref{fig:optimal-motion-window}, we find that for $W_M=1$, roughly $50\%$ of motion vectors are zero while considering $W_M=11$ reduces this value to less than $15\%$. We do not observe a significant reduction beyond $W_M=11$. For simplicity, we keep this parameter constant across datasets.

Finally, note that we utilize motion information to identify frame patches that have likely undergone significant transformations relative to preceding frames. This strategy enables us to provide the transformer encoder of \model{} ($\mathcal{E}_\mathcal{S}$) with patches expected to exhibit less redundancy with previous frames. Importantly, this approach does not supply the encoder with motion information, meaning the network remains unaware of explicit motion patterns.

\section{Zero-shot Grounding Algorithm} 
\label{sec:grounding-algorithm}
The task of natural language video grounding involves temporally localizing a natural language description within a single video.  Given the fine-grained temporal localization requirements of the task, dense frame sampling and encoding are indispensable, making it an ideal testbed for our efficient \model{} approach. This section details feature encoding and describes the motivations for addressing the task in a zero-shot setting.

We argue that the zero-shot setting holds valuable properties. 
Firstly, algorithms evaluated in a zero-shot manner are not prone to be affected by the inherent biases of the downstream datasets, which have shown to be a danger for this task~\cite{otani2020uncovering, soldan2022mad, zhang2021towards}. Additionally, models exhibiting strong zero-shot capabilities typically demonstrate enhanced generalization to unseen datasets, thereby increasing their versatility and utility.
Secondly, from a practical standpoint, relying on multiple specialized models for each new dataset can severely limit the scalability and versatility of systems. In contrast, a unified model that excels in zero-shot settings streamlines system architecture and boosts scalability and adaptability. Such models simplify the maintenance and deployment of deep learning applications and readily adjust to new challenges without the need for extensive retraining.
Third, the zero-shot approach promotes environmental sustainability.  This approach significantly curtails the computational demands by drastically reducing the necessity for ongoing retraining on possibly extensive datasets, thus lowering energy consumption and the associated carbon footprint. Employing large pre-trained models in a zero-shot manner optimizes their efficacy while minimizing further environmental impacts.
We strive to pursue zero-shot evaluation in this work for all these reasons.

\noindent\textbf{Visual Encoding.} Our algorithm begins with encoding a set of video frames $\mathcal{X} = \{x_t\}_{t=1}^{n_v}$ through a designated visual encoder (either a standard ViT or our \model{}). This process generates a series of frame features $\{f_t\}_{t=1}^{n_v}$. When employing \model{}, in line with the approach illustrated in Figure 2a, we utilize a sliding window mechanism that concurrently processes $N+1$ frames. 
The first frame in each window is encoded by the foundation model encoder $\mathcal{E}_\mathcal{V}$, with the resulting features stored for subsequent use. The following $N$ frames are processed by encoder $\mathcal{E}_\mathcal{S}$ (Figure 2b), which takes as input the frame tokens and the cached feature of the first frame of the window. 
The residual feature is first transformed into the residual token via the residual tokenizer. Subsequently, in the reduction module, frame tokens are reduced according to a particular strategy and token reduction probability $p$. Finally, these sparse visual tokens are concatenated with the residual token and forwarded to the visual encoder $\mathcal{E}_\mathcal{V}$. 

\noindent\textbf{Language Encoding.} The language encoder is kept frozen throughout our experiments and initialized with CLIP weights corresponding to the specific version of the visual encoder (ViT-B/32, ViT-B/16, or ViT-L/14). To solve the task, each sentence $s$ is first tokenized and then processed through the language encoder to derive a single sentence feature $g_l$.

\noindent\textbf{Grounding Algorithm.} For the grounding task, cosine similarity between each frame embedding and the sentence embedding is calculated, creating a temporal sequence of similarity scores $\{S_{t}\}_{t=1}^{n_v}$. We post-process the similarity profiles with a moving average smoothing operation with window size $W_{MA}$.

Finally, inspired by methods in prior work such as~\cite{diwan2023zero, lei2021detecting}, we implement a watershed algorithm~\cite{roerdink2000watershed} for moment prediction. In this step, group consecutive timesteps where the similarity scores exceed a given threshold, effectively delineating temporally contiguous segments. The start and end timesteps of these segments constitute our moment predictions. 
Multiple predictions are sorted based on the highest frame-sentence similarity in their span. 

For short-video datasets, such as Charades-STA and ActivityNet-Captions, we compute the threshold as a scaled average of the scores, given by $ \frac{\alpha}{n_v} \sum_{t=1}^{n_v} S_{t} $, where $ \alpha $ is a scaling factor. Conversely, for the long-form MAD dataset, we normalize the scores to the range $[0, 1]$ and apply a fixed threshold $ \beta $, an approach that mitigates the influence of low-relevance similarities in longer sequences that can otherwise skew the average similarity score.
Section \ref{sec:supplementary-qualitative} presents several qualitative results showcasing the aforementioned similarity profile.

\section{Additional Short-Form NLTVG Comparisons} 
\label{sec:supplementary-short-nltvg}
\begin{table*}[!t]
\centering 
\resizebox{\linewidth}{!}{
\scalebox{1.0}{
\begin{tabular}{l|cc|ccc|ccc} 

\toprule
\toprule

&&Use&\multicolumn{2}{c}{Charades-STA}& Avg. Cost
& \multicolumn{2}{c}{ActivityNet-Captions} 
& Avg. Cost \\

& Supervision & Downstream
& \multicolumn{2}{c}{R@1 $\uparrow$} & Feature/sec $\downarrow$
& \multicolumn{2}{c}{R@1 $\uparrow$} & Feature/sec $\downarrow$ \\

&& Task Data 
& IoU=0.5 & IoU=0.7 & (GFLOPs)
& IoU=0.5 & IoU=0.7 & (GFLOPs) \\ 

\midrule
\midrule

 \color{gray} 2D-TAN~\citep{zhang2020learning} 
& \color{gray} Full & \cmark
& \color{gray} $39.8$ & \color{gray} $23.3$ & \color{gray} $\mathbf{74.2}$
& \color{gray} $44.0$ & \color{gray} $27.4$ & \color{gray} $\mathbf{19.3}$ \\ 

 \color{gray} CPNet~\citep{li2021proposal} 
& \color{gray} Full & \cmark
& \color{gray} $\underline{60.3}$ & \color{gray} $\underline{38.7}$ & \color{gray} $638.3$
& \color{gray} $40.6$ & \color{gray} $21.6$ & \color{gray} $\underline{38.5}$ \\ 

 \color{gray} CRaNet~\citep{sun2023video} 
& \color{gray} Full & \cmark
& \color{gray} $\mathbf{60.9}$ & \color{gray} $\mathbf{41.3}$ & \color{gray} $\underline{296.8}$
& \color{gray} $\mathbf{47.3}$ & \color{gray} $\mathbf{30.3}$ & \color{gray} $\mathbf{19.3}$ \\ 

\midrule
\midrule
 \color{gray} WSTG~\citep{chen2020look}
& \color{gray} Weak & \cmark
& \color{gray} $27.3$ & \color{gray} $12.9$ & \color{gray} $\mathbf{38.5}$ 
& \color{gray} $23.6$ & \color{gray}    $-$ & \color{gray} $\underline{38.5}$ \\ 

 \color{gray} CRM~\citep{huang2021cross} 
& \color{gray} Weak & \cmark
& \color{gray} $34.8$ & \color{gray} $16.4$ & \color{gray} $638.3$
& \color{gray} $32.2$ & \color{gray} $-$ & \color{gray} $\mathbf{23.2}$  \\ 

\color{gray} CPL~\citep{zheng2022weakly2} 
& \color{gray} Weak & \cmark
& \color{gray} $\mathbf{49.2}$ 
& \color{gray} $\mathbf{22.4}$ 
& \color{gray} $\underline{445.2}$
& \color{gray} $\mathbf{31.4}$ 
& \color{gray} $-$ 
& \color{gray} $115.5$ \\

\midrule
\midrule

 U-VMR~\citep{gao2021learning} 
& Pseudo & \cmark
& $20.1$ & $8.3$  & $\underline{289.5}$ 
& $26.4$ & $11.6$ & $962.5$ \\ 

PSVL~\citep{nam2021zero} 
& Pseudo & \cmark
& $31.3$ & $14.2$ & $638.1$ 
& $30.1$ & $14.7$ & $\underline{38.5}$ \\

PZVMR~\citep{wang2022prompt} 
& Pseudo & \cmark
& $33.2$ & $18.5$ & $638.1$
& $\underline{31.3}$ & $\mathbf{17.8}$ & $\underline{38.5}$  \\ 

CORONET~\citep{holla2023commonsense} 
& Pseudo & \cmark
& $34.6$ & $17.9$ & $638.1$
& $28.2$ & $12.8$ & $\underline{38.5}$ \\ 

LFVL~\citep{kim2023language}
& Pseudo & \cmark
& $\underline{37.2}$ & $\underline{19.3}$ & $638.1$  
& $\mathbf{32.6}$ & $\underline{15.4}$ & $\underline{38.5}$ \\ 

SPL~\citep{zheng2023generating}
& Pseudo & \cmark
& $\mathbf{40.7}$ & $\mathbf{19.6}$ & $\mathbf{166.5}$
& $27.2$ & $15.0$ & $83.3$ \\ 

\midrule
\midrule

UniVTG~\citep{lin2023univtg} 
& Zero-Shot & \xmark
& $25.2$ & $10.0$ & $70.0$  
& $-$ & $-$ & $-$ \\ 

MR-FVLM~\citep{luo2024zero} 
& Zero-Shot & \xmark
& $\mathbf{42.9}$ & $20.1$ & $1370.0$  
& $27.9$ & $11.6$ & $370.0$ \\  

 CLIP (B/32)
& Zero-Shot & \xmark
& $35.9$ & $18.7$ & $\underline{13.2}$  
& $27.8$ & $\mathbf{13.9}$ & $\underline{4.4}$ \\ 

\model{} (B/32)
& Zero-Shot & \xmark
& $34.2$ & $17.7$ & $\mathbf{6.1}_{(-53\%)}$  
& $27.3$ & $13.7$ & $\mathbf{2.0}_{(-53\%)}$ \\

 CLIP  (B/16)
& Zero-Shot & \xmark
& $37.7$ & $21.2$ & $50.7$ 
& $28.1$ & $\underline{13.8}$ & $16.9$ \\ 

\model{} (B/16)
& Zero-Shot & \xmark
& $37.8$ & $21.0$ & $22.4_{(-56\%)}$ 
& $27.5$ & $\underline{13.8}$ & $7.5_{(-56\%)}$ \\

 CLIP (L/14)
& Zero-Shot & \xmark
& $\mathbf{42.9}$ & $\mathbf{24.1}$ & $233.4$ 
& $\mathbf{29.1}$ & $\underline{13.8}$ & $77.8$ \\ 

\model{} (L/14)
& Zero-Shot & \xmark
& $\underline{41.5}$ & $\underline{23.8}$ & $102.6_{(-56\%)}$ 
& $\underline{28.3}$ & $13.5$ & $34.2_{(-56\%)}$ \\ 

\bottomrule
\bottomrule
\end{tabular}
}
}
\vspace{-.1cm}
\caption{\label{tab:sota}{\bf Short video state-of-the-art comparison.} We compare our approach against state-of-the-art methods using different levels of supervision. Our \model{} reduces the cost of frame encoding by $56\%$ while closely retaining the performance of the CLIP model. The best method in each block of directly comparable methods is bolded, and the second-best method is underlined. }
\end{table*}

For completeness, Table~\ref{tab:sota} provides a summary comparing state-of-the-art grounding methods, categorized into fully supervised~\cite{barrios2023localizing, escorcia2019finding, liu2020jointly, Mun_2020_CVPR, soldan2021vlg, xu2023boundary, li2021proposal, Zeng_2020_CVPR, zhang2020learning}, weakly supervised~\cite{chen2020look, huang2021cross, zheng2022weakly2}, pseudo-supervised~\cite{diwan2023zero, gao2021learning, holla2023commonsense, kim2023language, nam2021zero, wang2022prompt, zheng2023generating}, and zero-shot techniques~\cite{luo2024zero, lin2023univtg} within the context of short video setups.

In fully supervised settings, models are trained using video data, corresponding sentences, and temporal boundaries. In contrast, weakly supervised approaches eliminate the need for temporal annotations. Our work is more closely aligned with settings where textual or temporal labels are unavailable. In these scenarios, prior approaches have leveraged off-the-shelf concept detectors (e.g., for objects, actions, and scenes)~\cite{gao2021learning, nam2021zero, wang2022prompt}, which are used to automatically generate pseudo-annotations (sentences and temporal boundaries) for downstream tasks. These pseudo-annotations are then used to train grounding models. Other sources of pseudo-supervision include pretrained visual-language embeddings~\cite{kim2023language}, commonsense knowledge~\cite{holla2023commonsense, speer2017conceptnet}, and captioning methods~\cite{zheng2023generating}. Furthermore, methods employing complex proposal schemes, such as feature clustering~\cite{holla2023commonsense, kim2023language, nam2021zero} or sliding windows~\cite{wang2022prompt}, are often paired with strategies for supervised feature refinement. Although these methods do not rely on manually annotated labels, they still adapt model parameters using the training dataset for the target downstream task, which is why we categorize them as \textit{pseudo-supervised}.

In contrast, our zero-shot approach (described in Section~\ref{sec:grounding-algorithm}) can be directly compared to zero-shot methods such as UniVTG~\cite{lin2023univtg} and MR-FVLM~\cite{luo2024zero}, all of which avoid training on task-specific datasets.

For each method in Table~\ref{tab:sota}, we report the grounding accuracy on the Charades-STA~\cite{gao2017tall} and ActivityNet-Captions~\cite{Krishna_2017_ICCV} datasets, alongside the average embedding cost per second. Previous methods have used visual backbones such as ResNet152, C3D, BLIP, VGG-19, and I3D~\cite{carreira2017quo, he2016deep, li2022blip, simonyan2014very, tran2015learning}, with respective costs of $11.6$, $38.5$, $55.5$, $143.7$, and $148.4$ GFLOPs per feature.

Table~\ref{tab:sota} also reports the grounding accuracy using the vanilla CLIP and our \model{} features across different backbones. For \model{}, we employ motion-based token reduction with a probability of $p=85\%$ and set the interleave parameter to $N=2$. For the grounding algorithm, we set $W_{MA}{=}15$ and $\alpha{=}1.0$ for Charades-STA,  $W_{MA}{=}15$ and $\alpha{=}0.95$ for ActivityNet-Captions.

As highlighted in the main paper, our \model{} closely matches CLIP's grounding accuracy while reducing frame encoding costs by approximately $56\%$ across all ViT backbones. Despite not being trained on the downstream task data, our method still achieves competitive accuracy when compared to prior approaches that train on both datasets. Specifically, for the Charades-STA dataset, our approach offers the best cost vs.\ accuracy trade-off among all pseudo-supervised methods. For the ActivityNet-Captions dataset, our method with the B/16 backbone matches or surpasses the accuracy of three pseudo-supervised methods, while maintaining a lower computational cost.

\section{Feature Comparison under Full Supervision Setup}
\label{sec:cgdetr}
\begin{table*}[!t]
\centering
\resizebox{.85\linewidth}{!}{
\scalebox{1.0}{
\begin{tabular}{l|c|ccccc} 
\toprule
\toprule

&&\multicolumn{3}{c}{Charades-STA}&& Avg. Cost \\

& Features 
& \multicolumn{3}{c}{R@1 $\uparrow$} & mIoU $\uparrow$ & Feature/sec  $\downarrow$\\

&& IoU=0.3 & IoU=0.5 & IoU=0.7 && (GFLOPs) \\ 

\midrule
\midrule

CG-DETR 
& CLIP (B/32) 
& $63.6$ 
& $49.7$ 
& $26.8$ 
& $43.8$ 
& $4.4$\\ 

CG-DETR 
& \model{} (B/32) 
& $62.2$ 
& $48.2$ 
& $26.4$ 
& $42.5$ 
& $2.0_{(-53\%)}$\\ 

\midrule

CG-DETR 
& CLIP (B/32) + SlowFast
& $69.6$ 
& $57.1$ 
& $34.5$ 
& $49.0$
& $40.5$ \\ 

CG-DETR 
& \model{} (B/32)  + SlowFast
& $69.2$ 
& $56.5$
& $34.0$ 
& $48.7$
& $38.1$\\ 

\midrule

CG-DETR*
& CLIP (B/32) + SlowFast
& $70.4$ 
& $58.4$ 
& $36.3$ 
& $50.1$
& $40.5$ \\ 

\bottomrule
\bottomrule
\end{tabular}
}
}
\vspace{-.1cm}
\caption{\label{tab:cgdetr}{
{\bf Frame feature comparisons in full supervision setup.}
This table compares the performance of the baseline CG-DETR~\citep{moon2023correlation} on the Charades-STA dataset under two setups: (i) using either CLIP (B/32) or ResidualViT (B/32) alone, and (ii) combining SlowFast features with either CLIP (B/32) (as in the original manuscript~\citep{moon2023correlation}) or ResidualViT (B/32). Our \model{} achieves a $53\%$ reduction in frame encoding cost while closely maintaining the accuracy of the original setup. We denote with the symbol $*$ the accuracy as presented in the original paper (last row). Note that all other rows have been trained from scratch using the original codebase. }}
\end{table*}

In this section, we focus on a representative fully supervised baseline for Natural Language Temporal Video Grounding to evaluate the accuracy gap between CLIP and \model{} features. For this experiment, we selected CG-DETR~\cite{moon2023correlation}, a recent and well-performing publicly available baseline that natively utilizes CLIP features for the Charades-STA dataset. The results of our experiments are presented in Table~\ref{tab:cgdetr}, and we maintained all hyperparameters as defined by the official implementation. Notably, features were extracted at a rate of one frame per second. For all rows except the last one, we train CG-DETR from scratch. The last row reports the accuracy as presented in the original paper. We find that we cannot fully reproduce those results using the default settings. 

We begin by comparing the accuracy when using only CLIP features versus \model{} features, as shown in the first two rows of the table. For \model{}, we set $N{=}2$ and $p{=}85\%$. \model{} achieves a reduction in encoding cost of approximately $53\%$ while maintaining accuracy close to the CLIP features. Specifically, we observe a marginal drop of $1.4\%$ (relative $2.2\%$) for R@1-IoU=0.3, an absolute drop of $1.5\%$ (relative $3.0\%$) for R@1-IoU=0.5, and an absolute drop of $0.4\%$ (relative $1.5\%$) for R@1-IoU=0.7. These results indicate that, with an average relative accuracy drop of only $2.2\%$, we can achieve more than a $50\%$ reduction in encoding cost.

Additionally, we evaluated the accuracy of CG-DETR in its original configuration, where CLIP features are channel-wise combined with SlowFast~\cite{feichtenhofer2019slowfast} features. This setup significantly increases computational cost, as SlowFast features alone are estimated at $36.1$ GFLOPs per feature. While the addition of SlowFast features can boost average accuracy on average of approximately $7.0\%$, it comes with a $9.2\times$ increase in computational cost, representing an unfavorable trade-off. Nonetheless, when SlowFast features are combined with \model{} features, the computational cost is reduced by approximately $6\%$, with only a $0.5\%$ absolute drop (relative $1\%$) in average accuracy, providing once again a favorable balance between accuracy and cost reduction.

\section{Additional Long-Form NLTVG Comparisons} 
\begin{table*}[!t]
\centering 
\resizebox{\linewidth}{!}{
\setlength{\tabcolsep}{2pt}
\scalebox{1.0}{
\begin{tabular}{l|c|cc|ccc|ccc|cc} 

\toprule
\toprule

&Use 
&&
&\multicolumn{6}{c|}{MAD}& Avg. Cost\\

Grounding Algorithm
& Downstream 
& Features
& Visual 
& \multicolumn{3}{c}{R@1 $\uparrow$} &  \multicolumn{3}{c|}{R@5 $\uparrow$} & Feature/sec $\downarrow$\\

& Task Data
& 
& Backbone
& IoU=0.1 & IoU=0.3 & IoU=0.5 &  IoU=0.1 & IoU=0.3 & IoU=0.5 & (GFLOPs) \\ 

\midrule
\midrule

\color{gray} DenoiseLoc~\citep{xu2023boundary} 
& \color{gray} \cmark & \color{gray} CLIP & \color{gray} ViT-B/32 
& \color{gray} $1.1$  & \color{gray} $0.9$ & \color{gray} $0.5$ 
& \color{gray} $4.1$  & \color{gray} $3.3$ & \color{gray} $2.2$ 
& \color{gray} $21.8$  \\ 

\color{gray} 2D-TAN~\citep{zhang2020learning}
& \color{gray} \cmark & \color{gray} CLIP  & \color{gray} ViT-B/32 
& \color{gray} $3.2$  & \color{gray} $2.5$ & \color{gray} $1.6$ 
& \color{gray} $11.9$ & \color{gray} $9.3$ & \color{gray} $5.7$
& \color{gray} $21.8$  \\  

\color{gray} Moment-DETR~\citep{moment-detr} 
& \color{gray} \cmark & \color{gray} CLIP  & \color{gray} ViT-B/32 
& \color{gray} $3.6$  & \color{gray} $2.8$ & \color{gray} $1.7$ 
& \color{gray} $13.0$ & \color{gray} $9.9$ & \color{gray}$5.6$
& \color{gray} $21.8$  \\ 

\color{gray} VLG-Net~\citep{soldan2021vlg} 
& \color{gray} \cmark & \color{gray} CLIP   & \color{gray} ViT-B/32 
& \color{gray} $3.6$  & \color{gray} $2.8$  & \color{gray} $1.7$ 
& \color{gray} $11.7$ & \color{gray} $9.3$	& \color{gray} $6.0$ 
& \color{gray} $21.8$  \\ 

\color{gray} CONE~\citep{hou2022cone} 
& \color{gray} \cmark & \color{gray} CLIP  & \color{gray} ViT-B/32 
& \color{gray} $8.9$  & \color{gray} $6.9$ & \color{gray} $4.1$ 
& \color{gray} $20.5$ & \color{gray}$16.1$ & \color{gray} $9.6$
& \color{gray} $21.8$  \\

\color{gray} SOONet~\citep{Pan_2023_ICCV} 
& \color{gray} \cmark & \color{gray} CLIP   & \color{gray} ViT-B/32 
& \color{gray} $11.3$ & \color{gray} $9.0$  & \color{gray} $5.3$ 
& \color{gray} $23.2$ & \color{gray} $19.6$ & \color{gray}$13.1$
& \color{gray} $21.8$  \\ 

\color{gray} SnAG~\citep{mu2024snag}  
& \color{gray} \cmark & \color{gray} CLIP   & \color{gray} ViT-B/32 
& \color{gray} $10.4$ & \color{gray} $8.5$  & \color{gray} $5.5$ 
& \color{gray} $24.4$ & \color{gray} $20.3$ & \color{gray} $13.4$
& \color{gray} $21.8$  \\

\color{gray} RGNet~\citep{hannan2023rgnet} 
& \color{gray} \cmark & \color{gray} CLIP  & \color{gray} ViT-B/32  
& \color{gray} $12.4$ & \color{gray} $9.5$ & \color{gray} $5.6$ 
& \color{gray} $25.1$ & \color{gray}$18.7$ & \color{gray} $10.9$
& \color{gray} $21.8$  \\

\midrule
\midrule

Proposals~\citep{soldan2022mad} & \xmark & CLIP & ViT-B/32 
& $6.6$ & $3.1$ & $1.4$ 
& $15.1$ & $9.9$ & $5.4$
& $21.8$  \\

Watershed & \xmark & CLIP & ViT-B/32  
& $8.7$ & $5.5$ & $3.2$ 
& $21.1$ & $13.0$ & $7.3$
& $21.8$  \\ 

Watershed (ours) & \xmark & \model{} & ViT-B/32 
& $8.6$ & $5.4$ & $3.1$  
& $20.5$ & $12.6$ & $6.9$
& $10.2_{(-53\%)}$  \\ 

Watershed & \xmark & CLIP & ViT-B/16 
& $10.8$ & $6.8$ & $3.9$ 
& $24.5$ & $15.2$ & $8.5$
& $84.3$ \\

Watershed (ours) & \xmark & \model{} & ViT-B/16 
& $10.1$ & $6.4$ & $3.7$ 
& $23.5$ & $14.6$ & $8.1$
& $37.3_{(-56\%)}$  \\

Watershed & \xmark & CLIP & ViT-L/14 
& $13.3$ & $8.6$ & $5.0$ 
& $28.5$ & $18.2$ & $10.3$
& $389.2$ \\

Watershed (ours) & \xmark & \model{} & ViT-L/14 
& $10.7$ & $7.3$ & $4.3$ 
& $24.4$ & $16.6$ & $9.3$
& $171.0_{(-56\%)}$ \\

\bottomrule
\bottomrule
\end{tabular}
}
}
\vspace{.1cm}
\caption{\label{tab:sota-mad-appendix}
{\bf Long-form video state-of-the-art comparison.} 
\model{} outperforms the previous art both in accuracy and computational cost on the challenging long-form MAD dataset.
In these experiments, \model{} was configured with $N{=}2$, a token dropping probability $p{=}85\%$, and the center token dropping strategy.}
\end{table*}
\label{sec:supplementary-mad}
In this section, we present additional grounding results for the long-form MAD dataset. Table~\ref{tab:sota-mad-appendix} builds on Table~\ref{tab:sota-mad} from the main paper by incorporating results from supervised state-of-the-art methods and zero-shot watershed accuracy using CLIP features. 

We begin by emphasizing that our zero-shot watershed-based grounding algorithm, detailed in Section~\ref{sec:grounding-algorithm}, significantly outperforms the proposal-based method introduced by \cite{soldan2022mad}.
By comparing rows 9 and 10 of the table, where both algorithms utilize the same visual backbone (CLIP ViT-B/32), we isolate and evaluate their individual contributions. Our zero-shot watershed-based approach demonstrates superior accuracy, with relative improvements ranging from $43\%$ to $128\%$. Remarkably, our zero-shot results are comparable with, or even surpass, several fully supervised methods listed in rows 1 through 8.

Table~\ref{tab:sota-mad-appendix} also enables a direct comparison of different backbone features while keeping the grounding algorithm fixed, thereby contrasting CLIP with our \model{}. For \model{}, we utilize configurations of $N=2$, $p=85\%$, and a center token dropping strategy, resulting in an embedding cost reduction of $53\%$ to $56\%$. For the grounding algorithm, we set  $W_{MA}{=}7$ and $\beta{=}0.7$.

When using the ViT-B/32 backbone (rows 10-11), \model{} reduces computational costs by approximately $53\%$, with an average accuracy degradation of just $0.1\%$ compared to CLIP weights, a negligible decrease. Similarly, employing the ViT-B/16 backbone (rows 12-13) \model{} achieves a $56\%$ reduction in computation with respect to CLIP, accompanied by an average accuracy drop of $0.5\%$. For the larger ViT-L/14 backbone, the average accuracy drop is $1.5\%$, with the most significant decrease occurring for the less stringent metric (R@1 IoU=0.1). 
We hypothesize that the slightly larger accuracy drop observed on the MAD dataset is due to frequent shot transitions (${\sim}1$k per movie), which disrupt the temporal correlations between I- and P-frames, unlike Charades-STA, which contains no such transitions.
Nonetheless, these results demonstrate that \model{} offers an excellent accuracy-to-cost reduction trade-off across all ViT variants within the MAD dataset.

\section{Additional Automatic Audio Description Setting}
\label{sec:supplementary-ad}
This section presents further results on the task of Automatic Audio Description. Unlike the results in the main paper (Table~\ref{tab:ad} in Section~\ref{sec:experiments}), we include $64$ contextual Audio Descriptions as additional input to the GPT-2 model. Table~\ref{tab:ad-supp} details the accuracy of both CLIP and \model{} features across various backbone sizes without audio context.

In this setup, both the AudioAD baseline equipped with CLIP or \model{} features exhibit comparable accuracy. Notably, with the ViT-B/32 backbone, \model{} shows an average accuracy drop of $2.2\%$. However, for the ViT-B/16 and ViT-L/14 backbones, \model{} model outperforms CLIP by $1.6\%$ and $1.9\%$, respectively, while being $56\%$ more efficient. 

This experiment further highlights the excellent accuracy vs cost tradeoff achieved by \model{} with respect to the CLIP model.  

\begin{table}[!t]
\centering  
\resizebox{\linewidth}{!}{%
    \scalebox{1.0}{
        \begin{tabular}{l|cccccc} 
        \toprule\toprule
        &&&&&& Avg. Cost\\
        
        & BertS $\uparrow$
        & R-L $\uparrow$
        & C $\uparrow$ 
        & M $\uparrow$ 
        & S $\uparrow$ 
        & Feature/sec $\downarrow$ \\
        
        &&&&&& (GFLOPs) \\ 
        
        \midrule \midrule
        
        CLIP (B/32)            
        & $23.8$  
        & $\underline{13.0}$ 
        & $\underline{17.7}$ 
        & $5.7$
        & $\underline{5.0}$ 
        & $21.8$ \\

        ResidualViT (B/32)     
        & $23.9$  
        & $12.8$ 
        & $17.0$ 
        & $5.5$
        & $4.9$ 
        & $10.2_{(-53\%)}$ \\
        
        CLIP (B/16)           
        & $\mathbf{25.0}$  
        & $12.9$ 
        & $17.5$ 
        & $5.4$
        & $\mathbf{5.1}$  
        & $84.3$\\

        ResidualViT (B/16)     
        & $24.3$ 
        & $\mathbf{13.2}$ 
        & $\mathbf{18.0}$ 
        & $\underline{5.8}$
        & $\underline{5.0}$ 
        & $37.3_{(-56\%)}$ \\
        
        CLIP (L/14)           
        & $\underline{24.5}$  
        & $\underline{13.0}$ 
        & $17.0$ 
        & $5.6$
        & $4.8$ 
        & $389.2$\\
        
        ResidualViT (L/14)     
        & $24.2$  
        & $\mathbf{13.2}$ 
        & $16.9$ 
        & $\mathbf{5.9}$
        & $\mathbf{5.0}$ 
        & $171.0_{(-56\%)}$ \\ 
        \bottomrule\bottomrule
        \end{tabular}
    }
}
\vspace{-.1cm}
\caption{{\bf Automatic Audio Description.}
\model{} provides nearly identical captioning quality to CLIP across all backbone sizes and metrics at a cheaper encoding cost.}
\label{tab:ad-supp}
\end{table}

\section{Additional TAL Comparisons}
\label{sec:supplementary-tal}
\begin{table}[!t]
\centering  
\resizebox{.8\linewidth}{!}{%
    \scalebox{1.0}{
        \begin{tabular}{l|cc} 
        \toprule\toprule
        && Avg. Cost \\
        
        Backbone 
        & mAP $\uparrow$ 
        & Feature/sec $\downarrow$ \\
         
        & (\%) & (GFLOPs) \\ 
        
        \midrule \midrule

        TSM-R50~\cite{lin2019tsm}         
        & $34.51$ & $123.3 $ \\
        
        TSN-R50~\cite{TSN2016ECCV}         
        & $34.64$ & $385.1 $ \\
        
        SlowFast-R101~\cite{feichtenhofer2019slowfast}   
        & $35.95$ & $247.9 $ \\
        
        VideoSwin-B~\cite{vswin}     
        & $35.60$ & $675.0 $ \\
        
        VideoSwin-L~\cite{vswin}     
        & $35.91$ & $2238.8$  \\
        
        VideoMAE-H~\cite{tong2022videomae}      
        & $\underline{36.96}$ & $4470.0$  \\
        
        InternVideo2-6B~\cite{wang2024internvideo2} 
        & $\mathbf{38.95}$ & $5137.5$  \\

        \midrule \midrule
        
        CLIP (B/32)     & $34.05$ & $4.4$   \\
        \model{} (B/32) & $33.42$ & $2.0_{(-53\%)}$  \\
        
        CLIP (B/16)     & $34.40$ & $16.9$   \\
        \model{} (B/16) & $33.76$ & $7.5_{(-56\%)}$  \\
        
        CLIP (L/14)     & $\mathbf{34.83}$ & $77.8$   \\
        \model{} (L/14) & $\underline{34.46}$ & $34.2_{(-56\%)}$  \\
        
        \bottomrule\bottomrule
        \end{tabular}
    }
}
\vspace{-.1cm}
\caption{{\bf Temporal Action Localization.} \model{} provides competitive accuracy at a fraction of the CLIP computational cost. When comparing against additional backbones, \model{} provides a good cost vs.\ accuracy tradeoff with an orders-of-magnitude cheaper model and comparable mAP.}
\label{tab:tal-supp}
\vspace{-.2cm}
\end{table}

This section presents further results on the task of Temporal Activity Localization. Table~\ref{tab:tal-supp} complements the results presented in the main paper with comparisons against additional backbones. 
The ActionFormer~\cite{zhang2022actionformer} baseline is trained from scratch for all sets of features.

We find that all commonly used backbones (\ie, TSN, TMS, SlowFast, VideoSwin, VideoMAE, and InternVideo2) impose a much higher computational cost for feature extraction. This is due to their temporal modeling design, which makes them particularly costly. Conversely, frame-based encoders such as CLIP and \model{} are not subjected to such high computational demands yet provide competitive accuracy for the task. 

In particular, contrasting \model{} (L/14) against TSM-R50 and TSN-R50, we observe that for equivalent mAP, \model{} is $3.6$-$11.2\times$ more efficient. This finding showcases the excellent accuracy vs.\ cost tradeoff achieved by \model{}.

\section{Additional Ablations}
\label{sec:supplementary-ablations}
In this section, we delve deeper into the design choices of \model{} by performing ablation studies on its token reduction mechanisms and distillation strategy. We begin by testing several designs for token-dropping strategies as presented in Section \ref{sec:supplementary-drop-strategy} and discussing the role of token-dropping probability. Next, we explore an alternative approach to the token reduction module by replacing token-dropping with a token merging strategy~\cite{bolya2022token}. We then assess the impact of reducing input frame resolution on the total number of tokens, providing insights into its effectiveness as a computational saving technique. Finally, we investigate an alternative distillation objective that eliminates the need for language annotations.

Note that, while semantically aware token reduction strategies~\cite{ding2023prune} could be incorporated, we leave this for future work due to their additional computational demands (\ie, complex token relevance computation at each level of the transformer encoder).

\noindent\textbf{Token Reduction Module Ablation - Token Drop Strategy.}
Here, we ablate the different token reduction strategies presented in Section~\ref{sec:supplementary-drop-strategy}. 
In Table~\ref{tab:naive-drop-ablation-resclip}, we contrast the grounding accuracy of the CLIP model (first row) against our \model{}  encoder. The lowest grounding accuracy is achieved by the center token reduction strategy with relative drops (vs. the CLIP model, first row) in the range of $10\%-12\%$. Uniform sampling produces slightly better accuracy with relative drops in the range of $6\%-7\%$. The second-best performing method is random, which decreases the drop to $3\%-5\%$. Finally, the motion-based strategy closely matches the grounding accuracy of the CLIP baseline with a relative drop in the range of $1\%-3\%$. Given the fixed token reduction probability, all settings result in a cost reduction of $56\%$ with respect to the naive CLIP frame encoding baseline. 

\begin{table}[!t]
\centering
\resizebox{\linewidth}{!}{%
\scalebox{1.0}{%
\begin{tabular}{c|c|cc|cc}
\toprule
\toprule
& \multirow{3}{*}{\begin{tabular}[c]{@{}c@{}}Drop \\ Strategy\end{tabular}}
& \multicolumn{2}{c|}{Charades-STA} 
& Avg. Cost & Memory Cost \\
& & \multicolumn{2}{c|}{R@1 $\uparrow$} & per Feature $\downarrow$ & per Feature $\downarrow$ \\
& & IoU=0.5 & IoU=0.7 & (GFLOPs) & (normalized) \\ 
\midrule\midrule
\multirow{5}{*}{\begin{sideways}ViT-L/14\end{sideways}} &
$ - $     & $42.9$ & $24.1$ & $233.4$ & $1 \times$ \\
\cmidrule{2-6}
& Random  & $40.8$ & $23.3$ & $102.6$ & $1 \times$ \\
& Uniform & $39.6$ & $22.5$ & $102.6$ & $1 \times$ \\
& Center  & $38.6$ & $21.1$ & $102.6$ & $1 \times$ \\
& Motion  & $41.5$ & $23.8$ & $102.6$ & $1.9 \times$ \\
\bottomrule
\bottomrule
\end{tabular}
}
}
\vspace{-.1cm}
\caption{\label{tab:naive-drop-ablation-resclip}{\bf Token reduction strategy ablation for \model{}.} We ablate four different token reduction strategies on the Charades-STA dataset. For all, we fix the token reduction probability to $85\%$. Memory cost is normalized according to the baseline memory footprint.}
\end{table}

\begin{table}[!t]
\centering
\resizebox{\linewidth}{!}{%
\scalebox{1.0}{%
\begin{tabular}{c|c|cc|cc}
\toprule
\toprule
& \multirow{3}{*}{\begin{tabular}[c]{@{}c@{}}Drop \\ Strategy\end{tabular}}
& \multicolumn{2}{c|}{Charades-STA} 
& Avg. Cost & Memory Cost \\
& & \multicolumn{2}{c|}{R@1 $\uparrow$} & per Feature $\downarrow$ & per Feature $\downarrow$ \\
& & IoU=0.5 & IoU=0.7 & (GFLOPs) & (normalized) \\ 
\midrule\midrule
\multirow{5}{*}{\begin{sideways}ViT-L/14\end{sideways}} &
$ - $     & $42.9$ & $24.1$ & $233.4$ & $1 \times$ \\
\cmidrule{2-6}
& Random  & $20.8$ &  $9.5$ & $102.6$ & $1 \times$ \\
& Uniform & $21.0$ & $10.6$ & $102.6$ & $1 \times$ \\
& Center  & $25.8$ & $13.2$ & $102.6$ & $1 \times$ \\
& Motion  & $28.5$ & $14.5$ & $102.6$ & $1.9 \times$ \\
\bottomrule
\bottomrule
\end{tabular}
}
}
\vspace{-.1cm}
\caption{\label{tab:naive-drop-ablation-clip}{\bf Token reduction strategy ablation for CLIP.} We ablate four different token reduction strategies on the Charades-STA dataset. For all, we fix the token reduction probability to $85\%$. Memory cost is normalized according to the baseline memory footprint.}
\end{table}

\begin{figure*}[!t]
    \begin{subfigure}[b]{.47\textwidth}
        \centering
        \includegraphics[width=\textwidth]{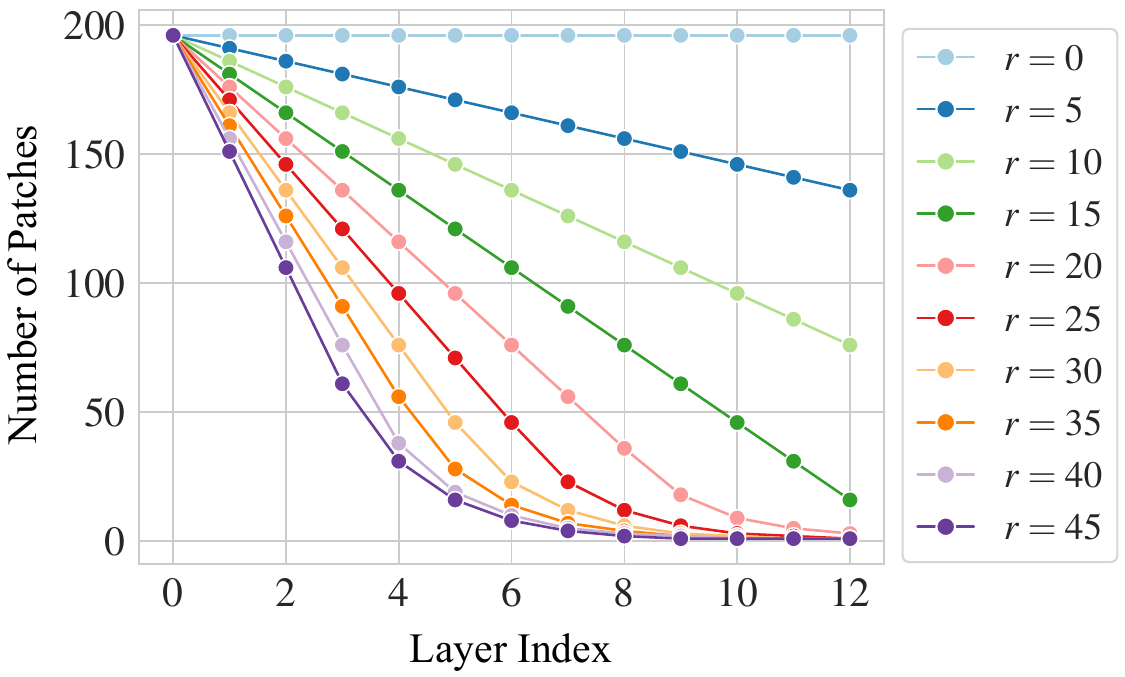}
        \caption{Tokens decay profile for different $r$ factors.}
        \label{fig:token-decay-merge}
    \end{subfigure}
    \hfill
    \begin{subfigure}[b]{0.51\textwidth}
        \centering
        \includegraphics[trim={0cm 0cm 0cm 0cm},width=\linewidth,clip]{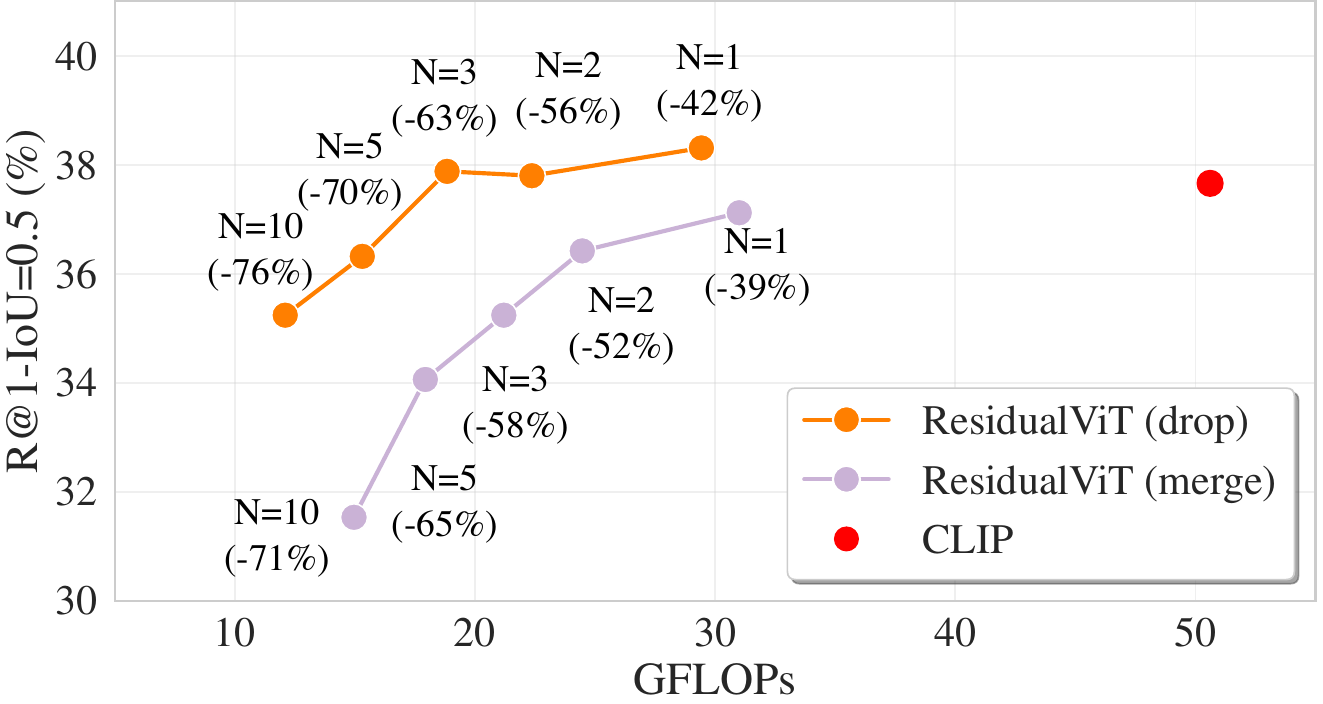}
        \caption{Drop vs Merge ablation.}
        \label{fig:token-drop-vs-merge}
    \end{subfigure}
    \caption{\textbf{Token dropping vs merging}. (a) We illustrate the relationship between the ViT layer index and the number of tokens resulting from the token merging operation for several canonical merging factors (r). (b) We compare the cost (GFLOPs) vs performance (R@1-IoU=0.5) for CLIP and \model. We present CLIP without any token reduction strategy (\textcolor{red}{\textbf{red}}), against our \model{} when the token reduction is token dropping (\textcolor{orange}{\textbf{orange}}) or token merging (\textcolor{nicepurple}{\textbf{purple}}). The ablation can conclude that token merging is less favourable due to lower performance at a comparable cost reduction.}
    \label{fig:N-ablation-merge}
\end{figure*}

Additionally, in Table~\ref{tab:naive-drop-ablation-clip}, we report the accuracy when the different token reduction strategies are applied to the CLIP model. In this case, we observe much wider differences between different token reduction strategies, where random and uniform strategies perform the worst with a relative accuracy drop in the range of $50\%-60\%$. The center token reduction strategy provides better accuracy, reducing the losses to $40\%-45\%$, while motion provides the best trade-off with a $3\%4-40\%$ drop. 

\begin{figure}[!t]
    \centering
    \includegraphics[trim={0cm 0cm 0cm 0cm},width=0.48\textwidth,clip]{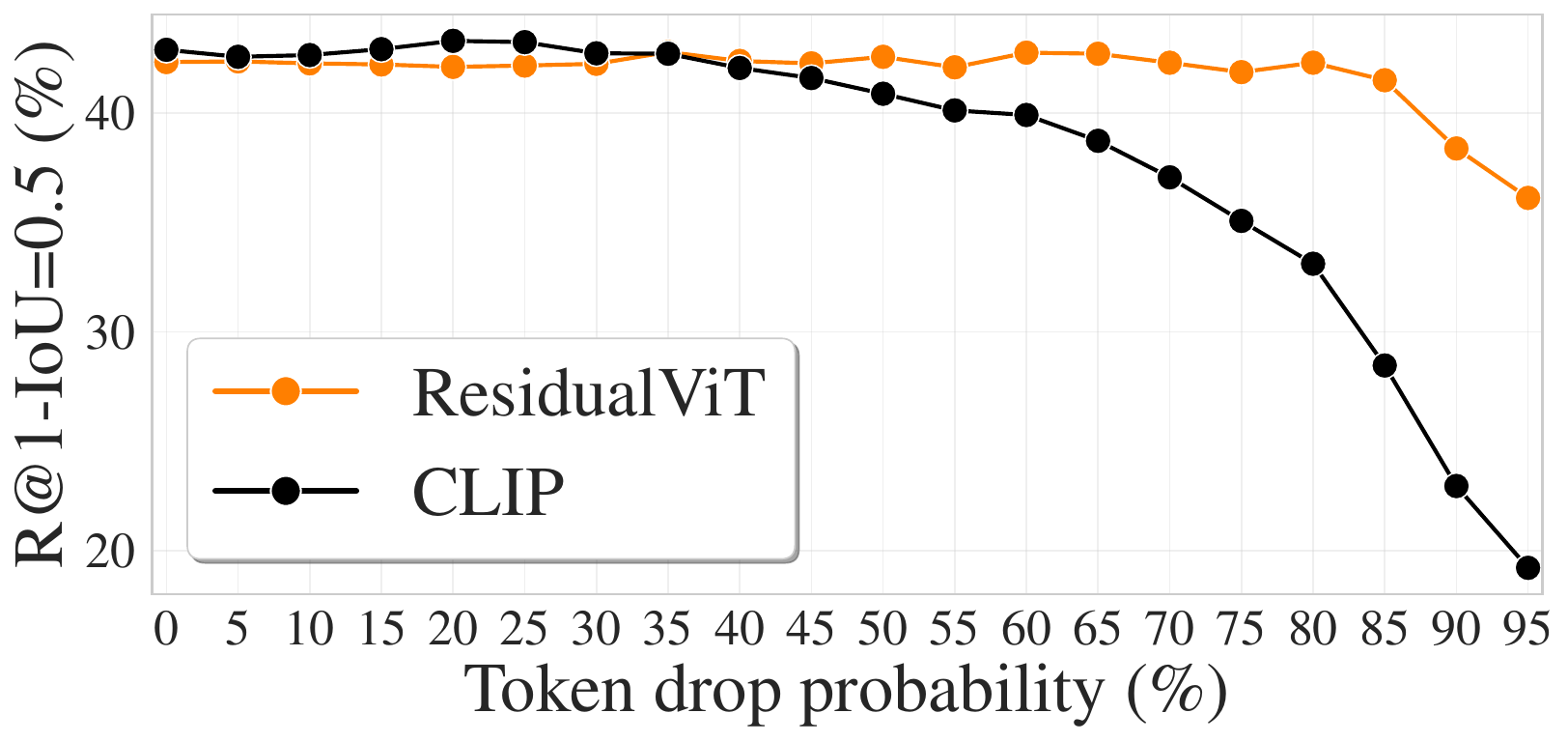}
    \caption{\textbf{Token drop probability ablation}. We showcase the performance of CLIP (\textbf{black}) and our \model{} (\textcolor{orange}{\textbf{orange}}) when progressively increasing the token drop probability. }
    \label{fig:p-drop-ablation-motion}
\end{figure}
It is important to observe that our model (Table~\ref{tab:naive-drop-ablation-resclip}) provides a certain level of resilience to the type of token reduction strategy compared to the baseline CLIP model (Table~\ref{tab:naive-drop-ablation-clip}). 
This finding suggests that for \model{}, token reduction strategies that avoid motion computation can serve as viable alternatives, especially in scenarios with limited memory or restricted computational resources. We attribute this finding to the learnable temporal residual connection, which enables the model to effectively compensate for the discarded tokens. 

\noindent\textbf{Token reduction probability.}
Here, we assess how varying the token reduction probability affects the accuracy of both the baseline CLIP model and our \model{} model. As depicted in Figure~\ref{fig:p-drop-ablation-motion}, the CLIP model (\textbf{black}) demonstrates a degree of robustness to the dropped tokens, maintaining relatively stable grounding accuracy until the token reduction probability reaches $35-40\%$. Beyond this setting, we observe a gradual decline in accuracy, which becomes more pronounced when the probability exceeds $80\%$.  In contrast, thanks to our model design, \model{} (\textcolor{orange}{\textbf{orange}}) exhibits a higher tolerance to dropped tokens, retaining relatively high grounding accuracy up to $p{=}85\%$ of dropped tokens.  

\noindent\textbf{Token Reduction Module Ablation - Token Merging.}
Our \model{} is agnostic to the implementation of the token reduction method. Therefore, we ablate replacing the token dropping strategy~\cite{liu2023patchdropout, hou2022token} with token merging~\cite{bolya2022token}, which has shown promising results in reducing the inference time of pre-trained ViT models.

This solution opts for merging a fixed number of tokens per layer, denoted by the $r$ parameter. Within each transformer block, the set of frame tokens at layer $l$, denoted as $\mathcal{T}^l$, is divided into two subsets: $\mathcal{T}^l_\textbf{odd}$, containing tokens at odd indices, and $\mathcal{T}^l_\textbf{even}$, containing tokens at even indices.
A bipartite matching is computed over the two sets by calculating the cosine similarity between the \textit{key} embeddings of tokens derived from the self-attention mechanism. The $r$ edges of the bipartite graph characterized by the highest similarity define the assignment. The connected tokens are then merged together via a weighted sum, where each token weight represents how many tokens were previously aggregated in it. Note that neither the \texttt{[CLS]} token nor the residual token is merged with the frame tokens. 
Following the bipartite assignment, the maximum number of token mergers per layer is limited to half of the total number of tokens available at layer $l$ ($min(|\mathcal{T}^l|/2,r)$). 

\begin{figure*}[!t]
    \begin{subfigure}[b]{.50\textwidth}
        \centering
        \includegraphics[width=\textwidth]{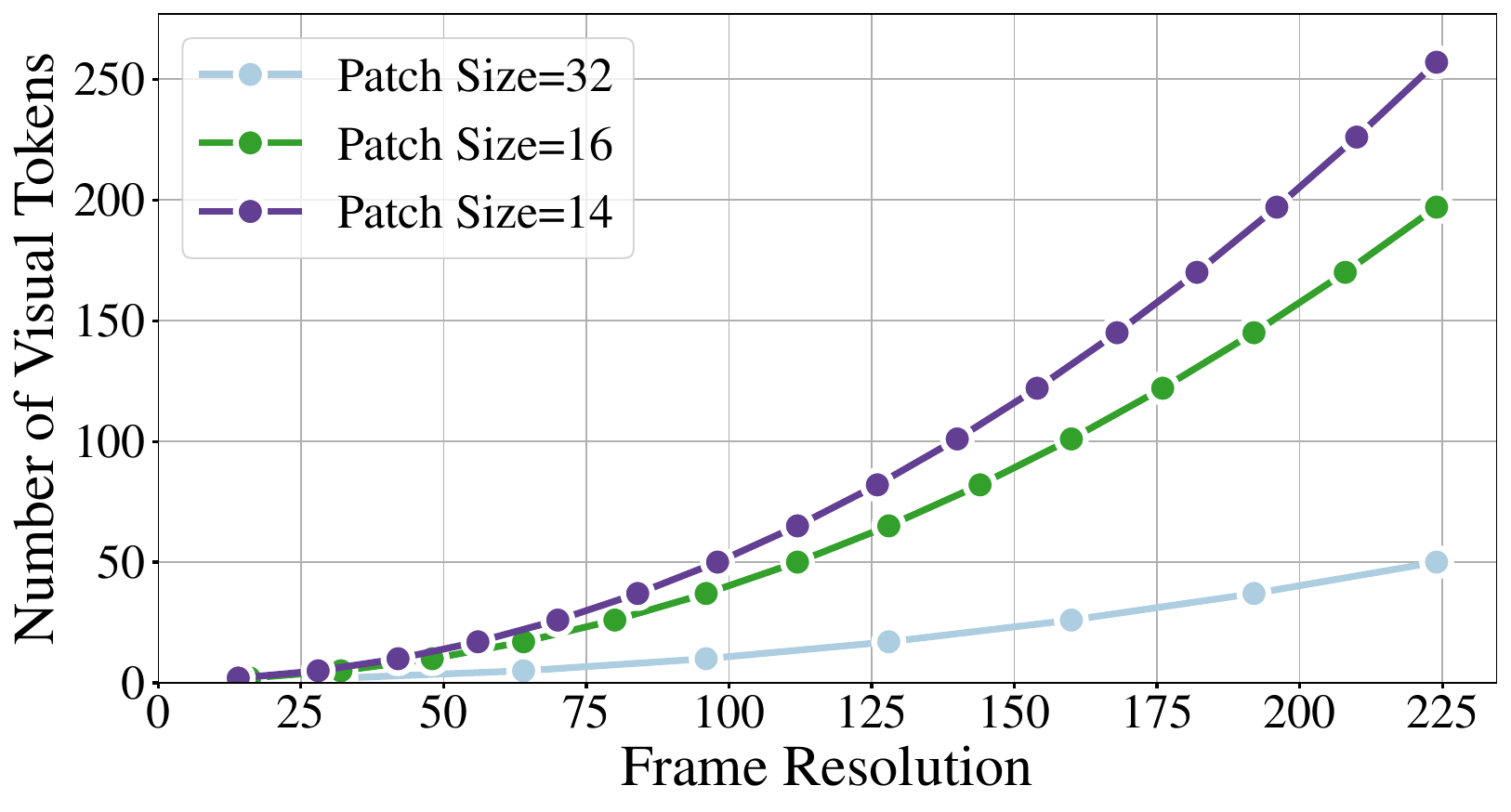}
        \caption{Number of visual tokens vs. frame resolution.}
        \label{fig:token-decay-resolution}
    \end{subfigure}
    \hfill
    \begin{subfigure}[b]{0.50\textwidth}
        \centering
        \includegraphics[trim={0cm 0cm 0cm 0cm},width=\linewidth,clip]{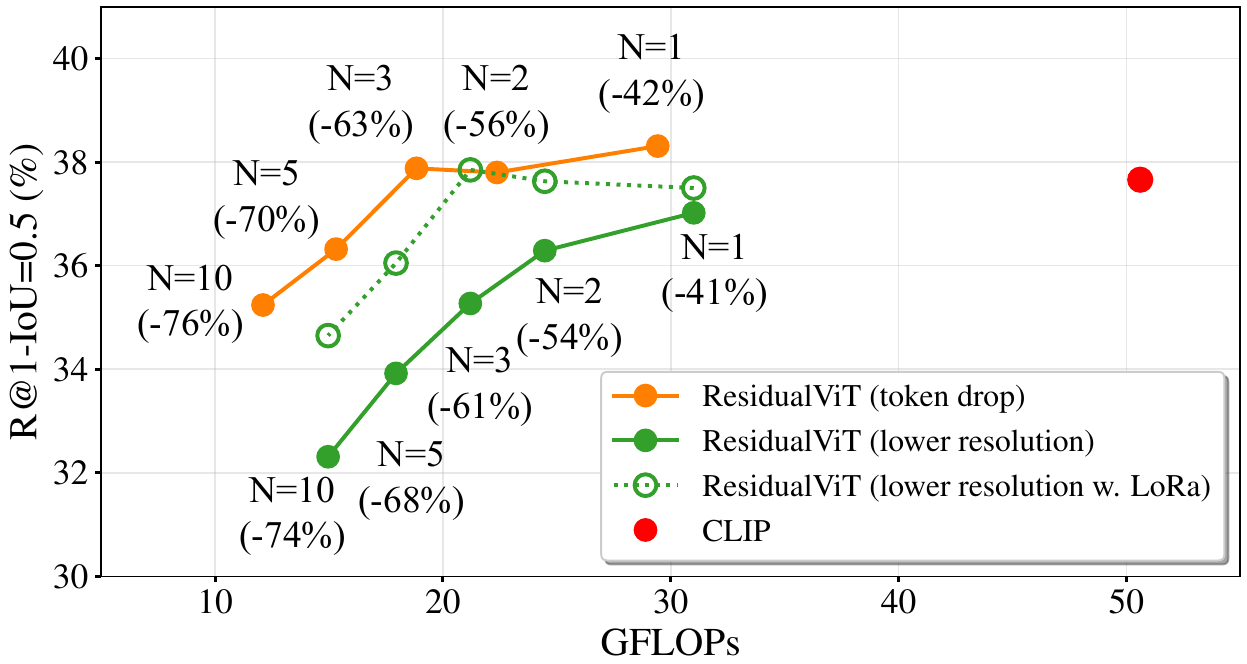}
        \caption{Token Drop vs. lower resolution input ablation.}
        \label{fig:token-drop-vs-resolution}
    \end{subfigure}
    \caption{\textbf{Token drop vs lower resolution}. (a) We illustrate the relationship between frame resolution and number of tokens as a function of three canonical patch sizes. (b) We compare the cost (GFLOPs) vs performance (R@1-IoU=0.5) for CLIP and \model. We present CLIP without any token reduction strategy (\textcolor{red}{\textbf{red}}), against our \model{} with token drop (\textcolor{orange}{\textbf{orange}}) or with lower input resolution (\textcolor{nicegreen}{\textbf{green}}). For the lower resolution setting, we additionally explore using LoRa~\citep{hu2021lora} adapters to finetune the input 2D convolution that implements the patchyfication operation. 
     }
    \label{fig:N-ablation-resolution}
\end{figure*}

This token-reduction strategy has the potential to reduce the information loss that affects the token dropping strategy, as the content of the tokens is retained even if their number is reduced. However, it presents other limitations. 
(i) Due to the progressive nature of the merging operation (after each transformer layer), to achieve a comparable cost reduction to token dropping, the $r$ parameter must be large. (ii) When the $r$ factor is moderately large, the majority of the tokens are merged together. This effect is showcased in Figure~\ref{fig:token-decay-merge}, where we see that for higher values of $r$, the number of tokens reduces to one quite early in the network (\eg, around the depth of layer 8 for $r=45$).

In Figure~\ref{fig:token-drop-vs-merge}, we conduct a comparative analysis of the token dropping and token merging strategies. For both strategies, we employ the ViT-B/16 backbone model. We set $p=85\%$ and used the motion-based strategy for token dropping. We set $r=40$ for token merging. We report R@1-IoU=0.5 grounding accuracy on Charades-STA. We compare the CLIP baseline in \textcolor{red}{\textbf{red}} against \model{} equipped with token dropping (\textcolor{orange}{\textbf{orange}}) or token merging (\textcolor{nicepurple}{\textbf{purple}}). 

Figure~\ref{fig:token-drop-vs-merge} shows token merging achieves overall lower grounding accuracy and incurs a significantly higher computation cost for its highest grounding accuracy setting ($\sim$30 GFLOPs for token merging with N=1 versus $\sim$17 GFLOPs for token dropping with N=3). Nonetheless, both strategies are capable of effectively reducing the cost with respect to the CLIP baseline (\textcolor{red}{\textbf{red}}). This result validates our design choices for the token reduction module of our \model{}.

\noindent\textbf{Spatial Resolution Ablation.}
An alternative to directly manipulating the number of frame tokens involves adjusting the spatial resolution of input frames.  This strategy has proven effective in dual-branch architectures~\cite{feichtenhofer2019slowfast}, where one branch processes a few high-resolution frames, and the other handles many low-resolution frames. In this section, we compare our approach against this strategy. 

In particular, we forward the full resolution I-frame to the ViT encoder $\mathcal{E}_\mathcal{V}$ and $N$ low-resolution P-frames to the \model{} encoder $\mathcal{E}_\mathcal{S}$. The number of tokens $|\mathcal{T}|$ for an input frame with resolution $(H, W)$ is calculated as $|\mathcal{T}| = \frac{H \times W}{P^2}$, where $P$ denotes the patch size. Consequently, reducing the frame resolution directly decreases the total number of tokens produced from the frame. We provide the relationship between frame resolution, patch size, and number of tokens in Figure~\ref{fig:token-decay-resolution}, examining trends across three canonical patch sizes: $P \in \{14, 16, 32\}$.

Subsequently, in Figure~\ref{fig:token-drop-vs-resolution}, we contrast the accuracy of two variations of \model{}. One variant employs token dropping (\textcolor{orange}{\textbf{orange}}), while the other utilizes a reduced input frame resolution (\textcolor{nicegreen}{\textbf{green}}), both using the ViT-B/16 backbone model. For the token dropping variant, we set $p=85\%$, and for reduced resolution, we adjust the spatial dimensions to $H=W=96$ pixels (as opposed to the default $H=W=224$). These modifications yield comparable reductions in computational cost, as demonstrated by the alignment of the data points along the x-axis. However, our results indicate that reducing the input resolution is less effective than employing token dropping in terms of accuracy (y-axis). Note that, in all experiments where the token reduction module is modified, we re-train the residual tokenizer to ensure consistent performance evaluation.

\begin{figure*}[!t]
    \centering
    \begin{subfigure}[b]{.49\textwidth}
        \centering
        \includegraphics[width=\textwidth]{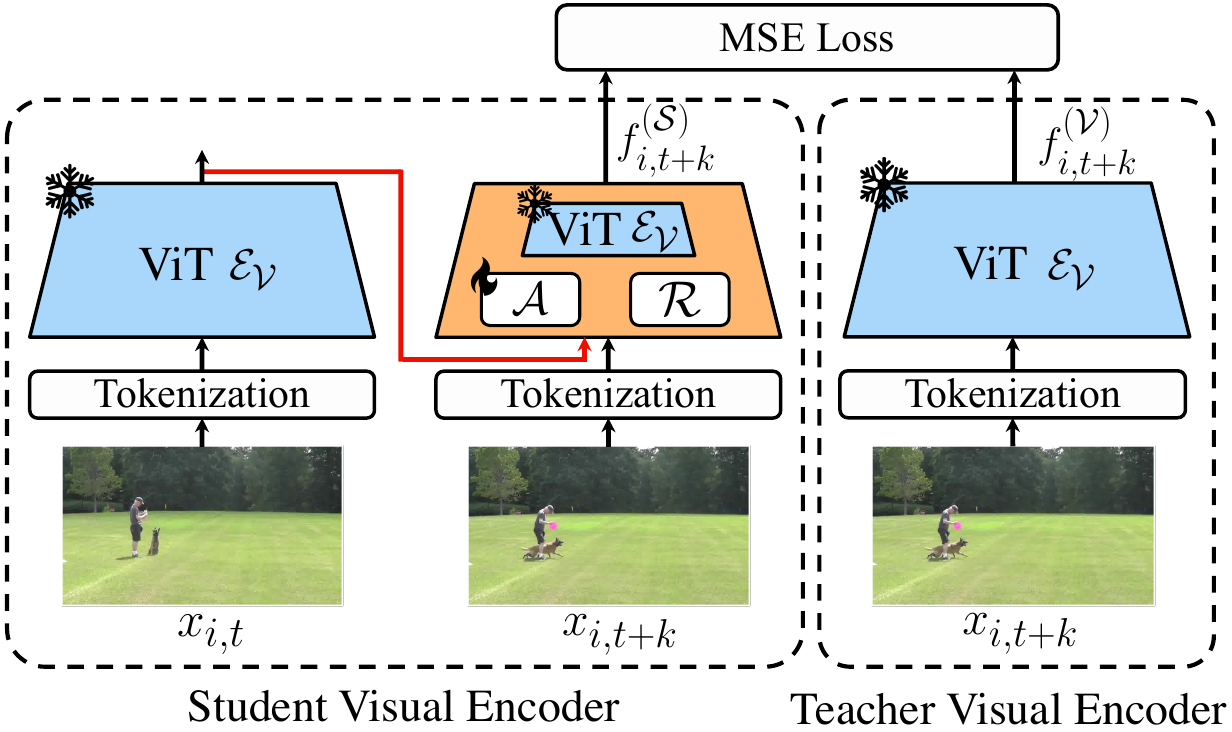}
        \caption{Distillation pipeline.}
        \label{fig:distillation-ablation-pipeline}
    \end{subfigure}
    \hfill
    \begin{subfigure}[b]{0.49\textwidth}
        \centering
        \includegraphics[trim={0cm 0cm 0cm 0cm},width=\linewidth,clip]{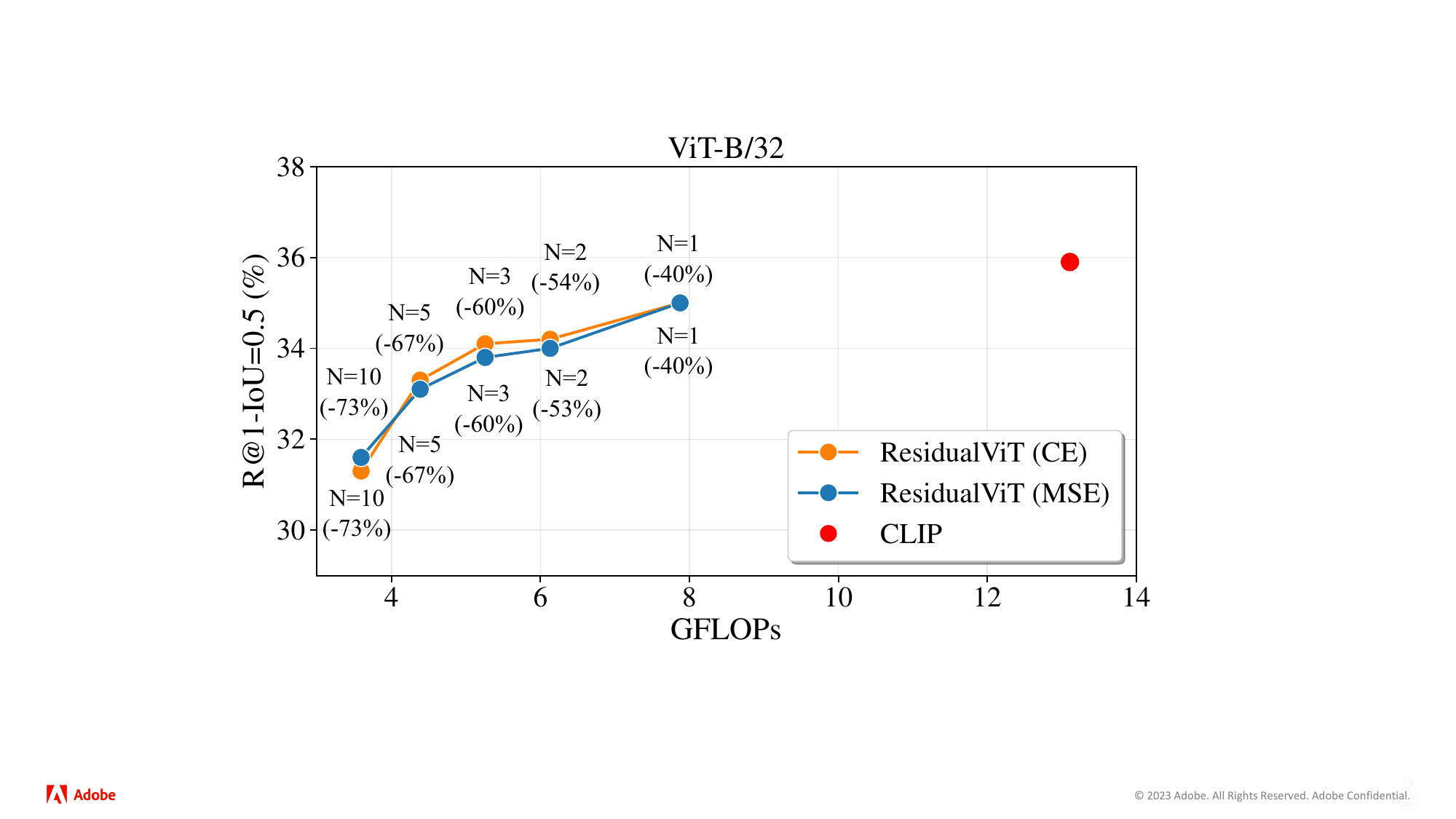}
        \caption{Downstream performance comparison ViT-B/32 backbone.}
        \label{fig:distillation-ablation-performance-b32}
    \end{subfigure}
    \begin{subfigure}[b]{0.49\textwidth}
        \centering
        \includegraphics[trim={0cm 0cm 0cm 0cm},width=\linewidth,clip]{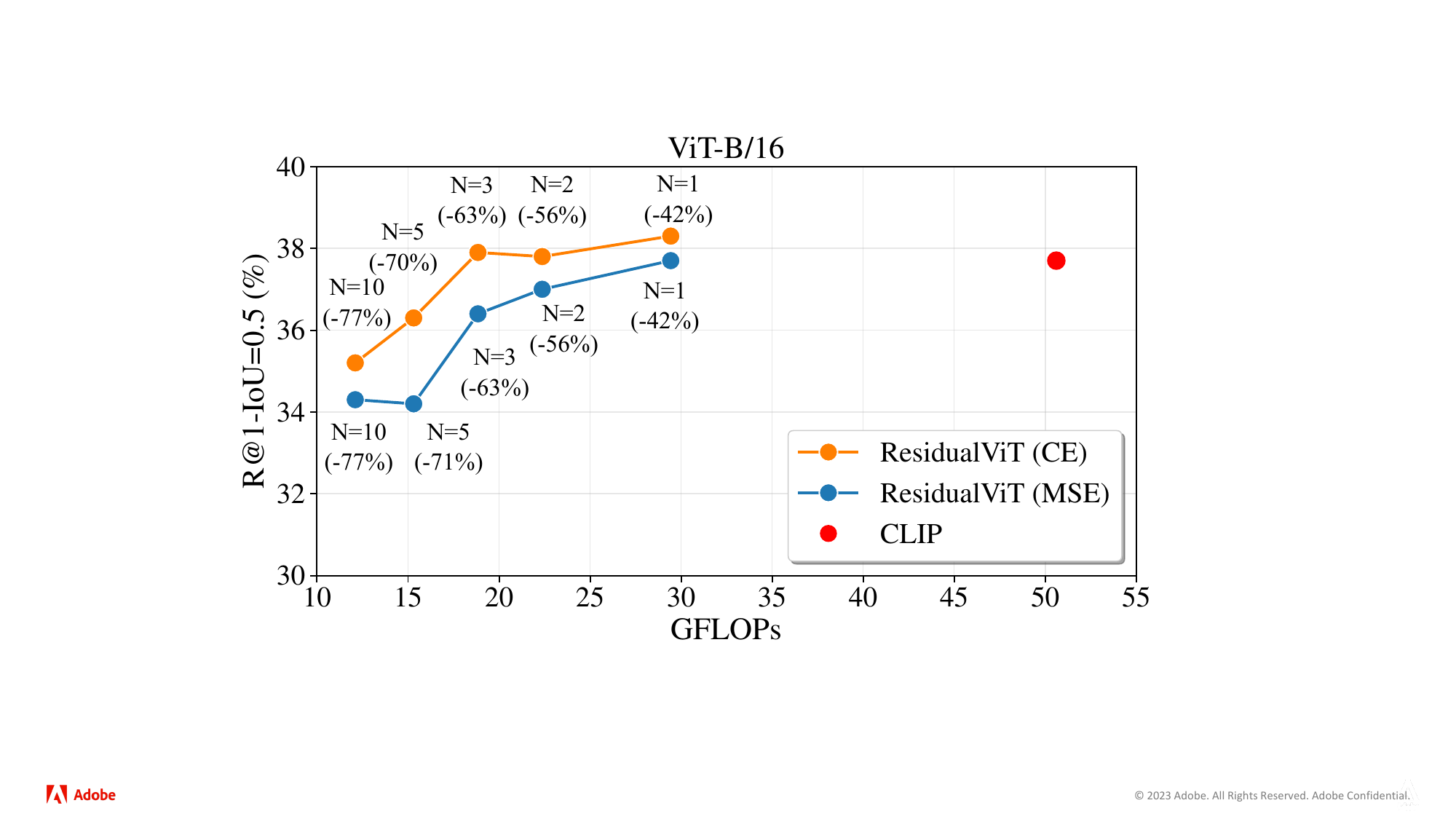}
        \caption{Downstream performance comparison ViT-B/16 backbone.}
        \label{fig:distillation-ablation-performance-b16}
    \end{subfigure}
    \hfill
    \begin{subfigure}[b]{0.49\textwidth}
        \centering
        \includegraphics[trim={0cm 0cm 0cm 0cm},width=\linewidth,clip]{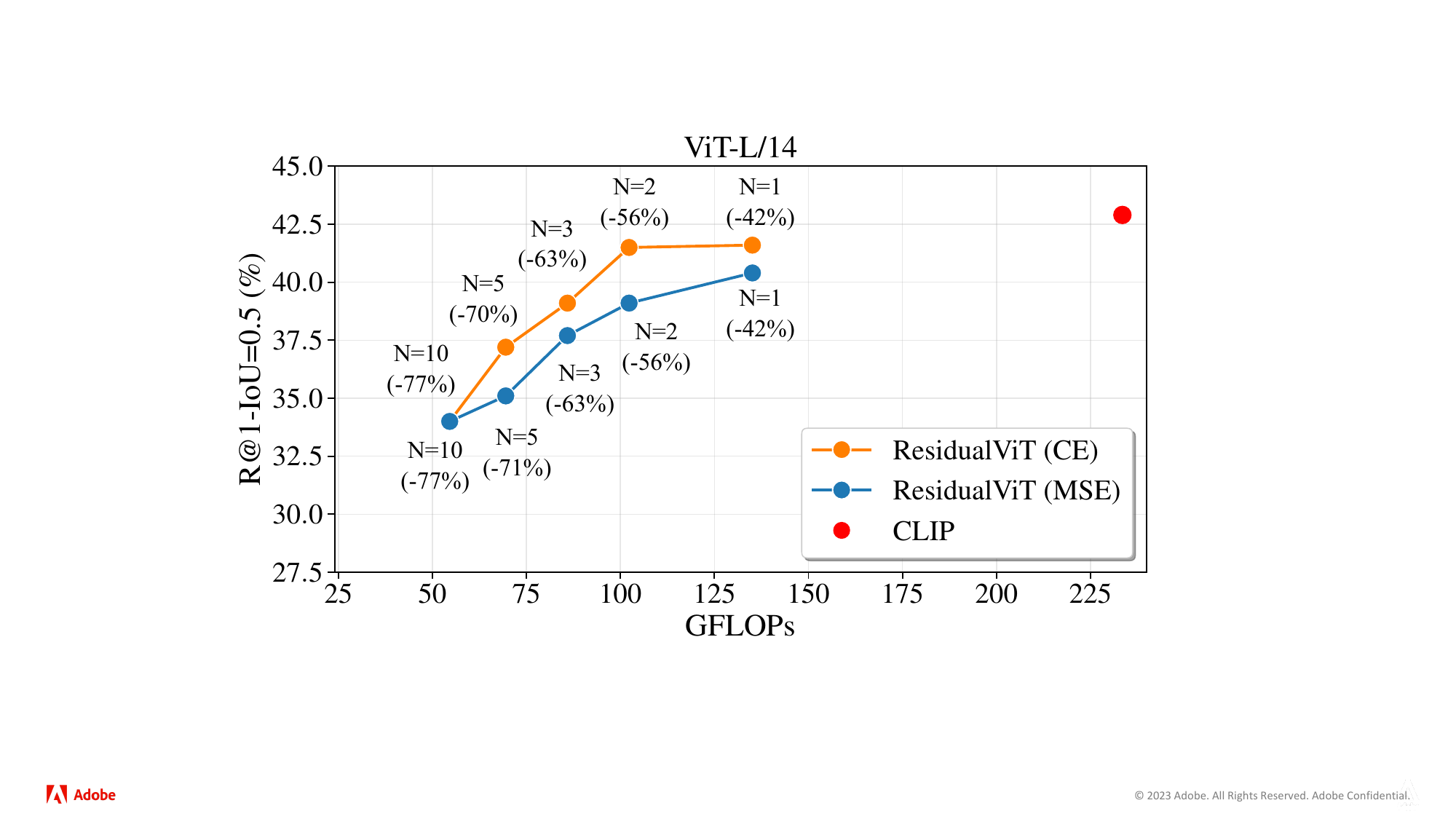}
        \caption{Downstream performance comparison ViT-L/14 backbone.}
        \label{fig:distillation-ablation-performance-l14}
    \end{subfigure}
    \caption{\textbf{Distillation loss ablation.} We ablate replacing the CE loss (Equation 2) with a Mean Square Error (MSE) loss. (a) Depicts the distillation pipeline when the MSE loss is used. (b-c) Summarizes the downstream performance comparison for the three different backbone sizes (ViT-B/32, ViT-B/16, ViT-L/14). The \textcolor{red}{\textbf{red}} represents CLIP's performance, while the \textcolor{orange}{\textbf{orange}} and \textcolor{navyblue}{\textbf{blue}} curves represent the performance of \model{} on the Charades-STA dataset when the distillation uses the original CE loss or the MSE loss respectively. We perform this ablation adopting the ViT-B/32 backbone. We conclude that the MSE loss, which does not require language annotations, produces near-identical results. }
    \label{fig:distillation-ablation}
\end{figure*}

We hypothesize that reducing the input frame resolution compromises the quality of the token representations inputted to the transformer. The process of converting image frames into tokens is implemented through a 2D convolution where both the kernel size and stride are set to the patch size. Previous research has indicated that although convolutional kernels can generalize to different resolutions, substantial changes in resolution can negatively impact accuracy~\cite{kannojia2018effects, richter2021input}. In our experiments, to match the computational cost reductions observed with the token reduction strategy, we decreased the resolution of inputs to the \model{} encoder by a factor of four. To address the resulting resolution mismatch, we explored fine-tuning the 2D convolutional layers using LoRa adapters~\cite{hu2021lora}. This adjustment helps account for the impact of lower-resolution inputs on token representation quality. 
Our findings show that incorporating LoRa adapters with lower-resolution inputs improves accuracy across all $N$ values and achieves accuracy comparable to the token drop strategy for $N=3$. However, the token drop strategy consistently outperforms this approach while maintaining the advantage of not requiring any weight modifications to the encoder $\mathcal{E}_{\mathcal{V}}$.

\noindent\textbf{Distillation Strategy.}  
To evaluate our distillation approach, we replace the Cross-Entropy (CE) loss (Equation~\ref{eq:training}) with a Mean Squared Error (MSE) loss, computed between frame features as $||f^{\mathcal{S}}_{i,t+k} - f^{\mathcal{V}}_{i,t+k}||_2$.  

This alternative setup, illustrated in Figure~\ref{fig:distillation-ablation-pipeline}, removes the need for language annotations, reducing training costs by approximately $10\%$ for ViT-B/32, $3\%$ for ViT-B/16, and $1.5\%$ for ViT-L/14. However, we emphasize that our primary focus is on the efficient deployment of a trained model (forward inference), rather than optimizing training efficiency.

Figure~\ref{fig:distillation-ablation-performance-b32} presents the results for ViT-B/32, where both loss functions achieve comparable accuracy, indicating that our distillation method remains effective regardless of the loss choice. However, in Figure~\ref{fig:distillation-ablation-performance-b16} and Figure~\ref{fig:distillation-ablation-performance-l14}, the CE loss consistently outperforms the MSE loss, supporting our choice of a language-supervised CE approach as the optimal strategy. For consistency, all training and testing hyperparameters remain unchanged across these ablations, ensuring that accuracy differences stem solely from the choice of the loss function.

\begin{figure*}[!t]
    \centering
        \includegraphics[trim={0cm 0cm 0cm 0cm},width=.8\textwidth,clip]{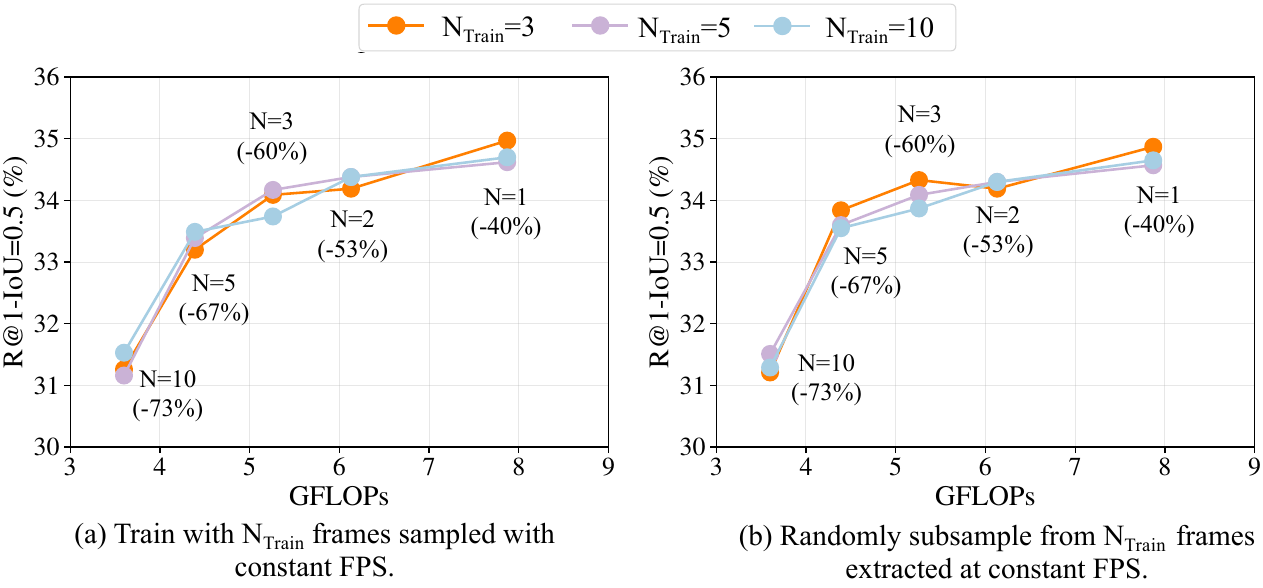}
    \vspace{-.3cm}
    \caption{\textbf{Training interleave factor ($N_{Train}$) ablation.} We compare the accuracy of \model{} when three different values of $N_{Train} \in \{ 3,5,10\}$ are used, and two different frame sampling strategies are implemented. In particular, we investigate (a) using all $N_{Train}$ frames sampled at a constant FPS$=1.0$, and (b) sampling a random number of frames from the $N_{Train}$ frames extracted at a constant FPS$=1.0$. 
    Results are reported on the Charades-STA dataset using the B/32 backbone. }
    \label{fig:N-training2}
\end{figure*}

\noindent\textbf{Training Interleave Factor ($N_{\text{Train}}$) Ablation.}
In this section, we evaluate how varying the interleave factor ($N_{\text{Train}}$) during training impacts \model{}'s accuracy and computational cost for different $N$ values during inference. Additionally, we explore whether different frame sampling strategies during training affect the model's final accuracy. We consider two distinct sampling approaches: (a) Sample $N_{\text{Train}}$ frames per training video at a constant frame rate.
(b) Extract $N_{\text{Train}}$ frames at a constant frame rate, but randomly subsample the frames before inputting them into the network.
Our findings are summarized in Figure~\ref{fig:N-training2}, where we present the accuracy vs.\ cost trade-off on the Charades-STA dataset using the B/32 backbone.

In this experiment, we train \model{} with varying $N_{\text{Train}} \in {3, 5, 10}$ and test these models with $N \in {1, 2, 3, 5, 10}$. Note that $N_{\text{Train}} = 3$ is our default setting, used for all other results in the manuscript.

Focusing on Figure~\ref{fig:N-training2}(a), we observe that different values of $N_{\text{Train}}$ produce very similar results, with $N_{\text{Train}} = 10$ showing slightly better accuracy for $N = 5$ and $N = 10$ compared to models trained with $N_{\text{Train}} = 3$. 

Figure~\ref{fig:N-training2}(b) supports the same conclusion. In this case, no clear advantage is observed for larger $N_{\text{Train}}$, as accuracy remains very similar across all configurations. Interestingly, $N_{\text{Train}} = 3$ (our default setting) shows slightly better accuracy for $N = 1$, $N = 3$ and $N = 5$.

\noindent\textbf{Frame Rate Ablation.} In this section, we evaluate the accuracy-cost trade-off between frame rate and computational cost for CLIP and \model{} on the Charades-STA dataset.

\begin{figure}[!t]
    \centering
    \includegraphics[trim={0cm 0cm 0cm 0cm},width=0.48\textwidth,clip]{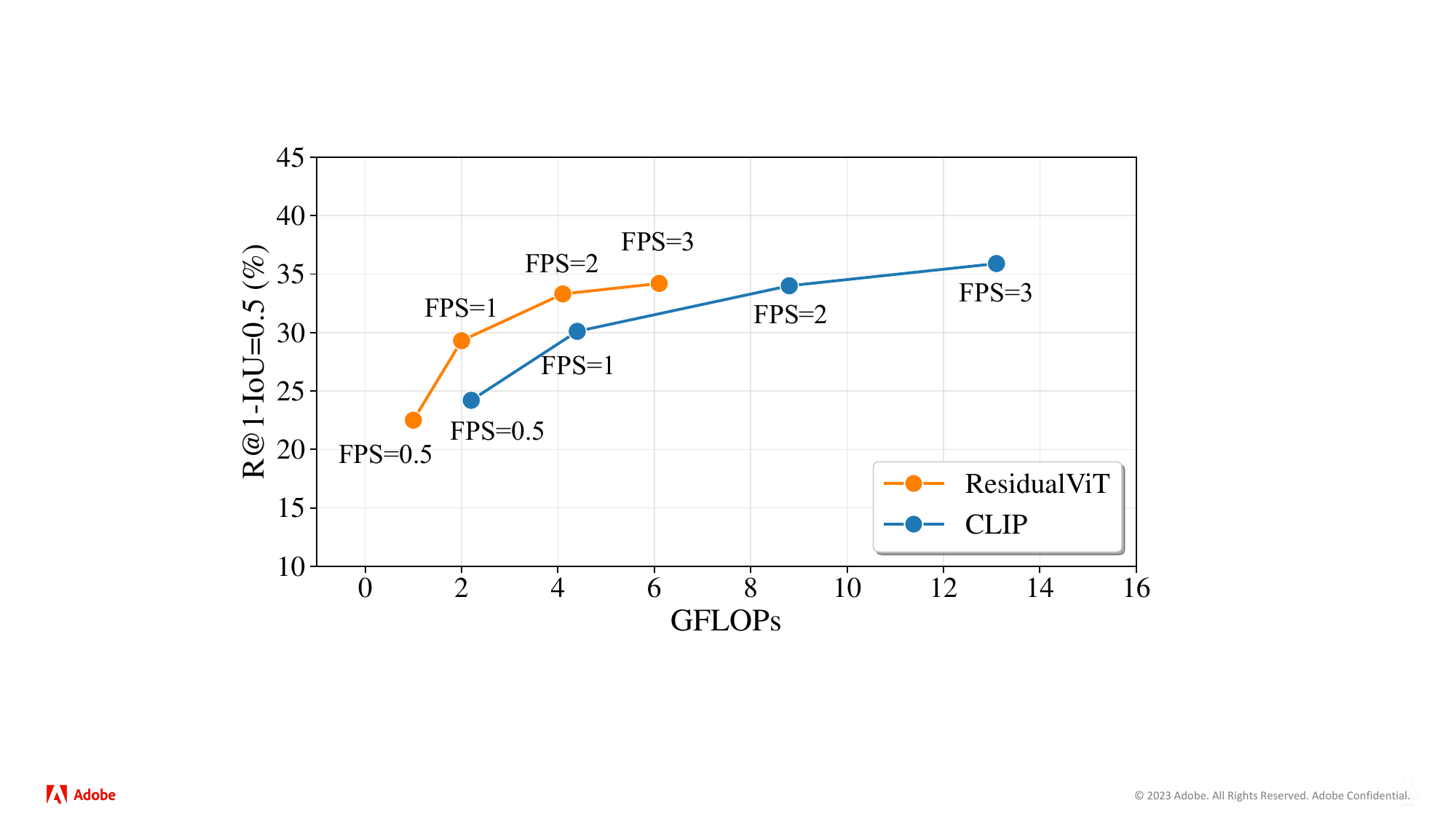}
    \caption{\textbf{Frame rate ablation.} We compare CLIP (\textcolor{navyblue}{\textbf{blue}}) and \model{} (\textcolor{orange}{\textbf{orange}}) features on the Charades-STA dataset for varying frame rates. The figure presents the accuracy (y-axis) vs.\ cost (x-axis) trade-off. }
    \label{fig:frame-rate}
\end{figure}

Figure~\ref{fig:frame-rate} illustrates the accuracy of both models on the NLTVG task as the frame rate varies from $0.5$ to $3.0$ (our default value). At the default frame rate of $3.0$, CLIP achieves an R@1-IoU=0.5 score of $35.9$, while \model{} achieves $34.2$—a slight accuracy drop, but with an approximate $53\%$ reduction in encoding cost. As the frame rate decreases, both methods exhibit a steady decline in accuracy. However, it is noteworthy that \model{} at FPS=3 incurs a lower cost than CLIP at FPS=2 while achieving comparable accuracy. Additionally, \model{} at FPS=2 outperforms CLIP at FPS=1, with similar computational cost.

Finally, we observe that the decrease in accuracy for \model{} as the FPS decreases becomes steeper than for CLIP. We believe that this is due to the large temporal gap between consecutive frames, which hinders the ability of the residual tokenizer to provide valuable information when computing P-features.

\noindent\textbf{Architecture Ablation on the MAD dataset.}
Table~\ref{tab:ablation_mad} presents the ablation of the main architecture components of our model on the MAD~\cite{soldan2022mad} dataset. This ablation setup is equivalent to the one presented in Table~1 of the main paper. 

The first row ({\bf a}) reports the accuracy of the original CLIP model, which requires $21.8$ GFLOPs/sec for feature computation. Applying the token reduction strategy uniformly across all frames ({\bf b}) results in an $85\%$ reduction in computational cost but leads to substantial accuracy losses of over $55\%$ across all IoU thresholds. Introducing our interleave strategy with $N{=}2$ ({\bf c}) significantly improves grounding accuracy while using only $47\%$ of the original compute budget, with a relative accuracy drop of approximately $25\%$. Finally, incorporating the residual tokenizer learned via distillation ({\bf d}) adds virtually no computational overhead and nearly matches the original CLIP accuracy, with only a $0.1\%$ absolute drop in accuracy.

These results are consistent with the findings in Table 1 of the main paper and further reinforce the effectiveness of each architectural component.

\begin{table}[!t]
\centering
\resizebox{\linewidth}{!}{%
    \scalebox{1.0}{
        \begin{tabular}{lccc|cc|c}
            \toprule\toprule
            &&&Residual& \multicolumn{2}{c|}{MAD} & Avg. Cost \\
            &Token & Interleave & Tokenizer & \multicolumn{2}{c|}{R@1 $\uparrow$} & Feature/sec $\downarrow$ \\
            &Reduction& Factor& (Distilled) & IoU=0.3 & IoU=0.5 & (GFLOPs) \\ 
            \midrule\midrule
            \textbf{a.} &&&& \color{gray} $5.5$ & \color{gray} $3.2$ & \color{gray} $21.8$\\
            \textbf{b.} &\cmark &&& $2.3$ & $1.4$ & $4.4_{(-85\%)}$\\
            \textbf{c.} &\cmark &\cmark && $4.2$ & $2.6$ & $10.2_{(-53\%)}$\\
            \textbf{d.} &\cmark &\cmark &\cmark & $5.4$ & $3.1$ & $10.2_{(-53\%)}$\\
            \bottomrule\bottomrule
        \end{tabular}
    }
}
\vspace{-.1cm}
\caption{{\bf Architecture ablation.} We ablate the main components of our architecture: the token reduction module, the interleave factor, and the distilled residual tokenizer. We set the token reduction probability $p$ to $85\%$, $N=2$, and use the ViT-B/32 backbone on the MAD dataset.  }
\label{tab:ablation_mad}
\end{table}
\begin{figure}[!b]
    \includegraphics[width=\linewidth]{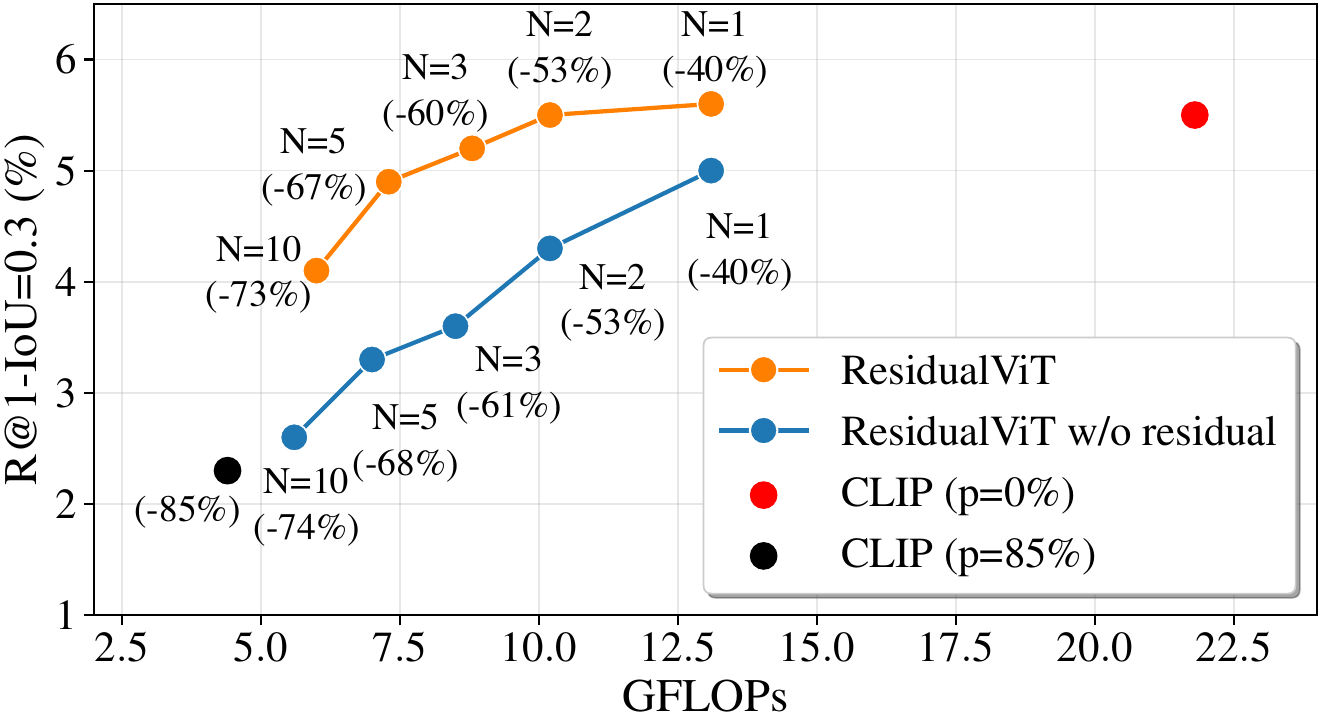}
    \vspace{-.3cm}
    \caption{\textbf{Interleaving frames ($N$)}. Our \model{} (\textcolor{orange}{\textbf{orange}}) closely retains CLIP's (\textcolor{red}{\textbf{red}}) performance for $N{=}1$ and $2$ while reducing cost by up to $53\%$.}
    \label{fig:N_factor_mad}
\end{figure}
\noindent\textbf{Interleave Factor $N$ and Benefits of Distillation on the MAD dataset.}
In Figure~\ref{fig:N_factor_mad}, we explore the relationship between grounding accuracy and computational cost as we vary the number of interleaved frames ($N$) on the MAD dataset. 
Here, the baseline CLIP model is shown in \textcolor{red}{\textbf{red}}, while our \model{}, applied with and without the distilled residual tokenizer module, is shown in \textcolor{orange}{\textbf{orange}} and \textcolor{navyblue}{\textbf{blue}}, respectively. We vary $N \in \{1,2,3,5,10\}$.  This ablation setup mirrors the one presented in the main paper for the Charades-STA dataset.

We observe that for $N{=}1$ and $N{=}2$, \model{} achieves accuracy on par with CLIP while reducing frame encoding costs by $40\%$ and $53\%$, respectively. Notably, removing the residual tokenizer (\textcolor{navyblue}{\textbf{blue}}) leads to a larger accuracy drop, highlighting the importance of distillation. Increasing $N$ beyond $2$ yields diminishing returns as cost savings start to plateau around $60\%$, while accuracy declines. This accuracy drop is attributed to the growing difficulty of predicting the CLIP feature at time $t{+}N$ from frame $t$, as CLIP similarity decreases with larger temporal gaps, especially in MAD, which contains more diverse content. Since ResidualViT approximates $f_{t+N}$ using a token subset and the residual tokenizer, lower similarity makes the prediction harder. This can be mitigated by retaining more tokens, but at higher computational cost. As shown in Figure~\ref{fig:N_factor_mad}, $N{=}2$ provides the best trade-off between efficiency and accuracy.

These results align with those in Figure 4 of the main paper and further validate the contributions of both the interleave encoding strategy and the residual tokenizer module.

\section{\model{} Runtime}
\label{sec:supplementary-time-efficiency}
\begin{figure}[!t]
    \centering
    \includegraphics[trim={0cm 0cm 0cm 0cm},width=0.48\textwidth,clip]{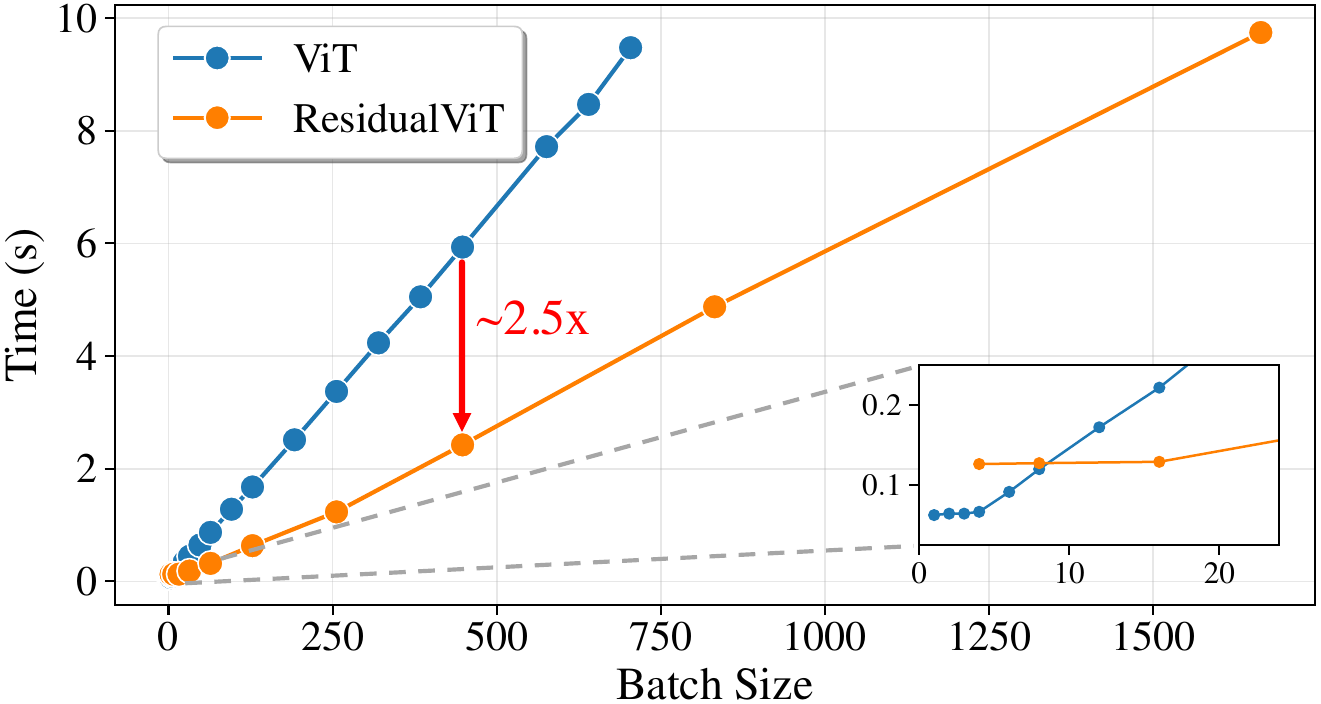}
    \vspace{-.3cm}
    \caption{\textbf{Inference time comparison}. When varying the batch size, we showcase the runtime difference of a standard ViT (\textcolor{navyblue}{\textbf{blue}}) against our \model{} (\textcolor{orange}{\textbf{orange}}). We demonstrate that our approach is ${\sim}2.5\times$ faster than a standard ViT. Moreover, for the same time budget (\ie, 10 seconds), we can accommodate ${\sim}2.5\times$ more samples in the batch without incurring Out Of Memory issues.}
    \label{fig:time-comparison}
\end{figure}
In our manuscript, we have focused on characterizing the computational cost reductions in terms of GFLOPs. However, our system introduces a dependency where P-feature computation relies on the prior computation of I-features. Specifically, the I-features are first processed through the ViT encoder $\mathcal{E}_\mathcal{V}$, followed by the computation of P-features via the \model{} encoder $\mathcal{E}_\mathcal{S}$, which also incorporates the residual token. This design necessitates two sequential forward passes through distinct encoders, prompting us to examine the encoding latency costs inherent to this approach. One possible way to mitigate the latency due to this sequential dependency is via parallel processing via batching of the I-features, followed by batching of the P-features.

In Figure~\ref{fig:time-comparison}, we present the forward pass wall-clock latency as a function of batch size, comparing the timings for a standard ViT-L/14 model and our \model{}, which employs the same ViT-L/14 backbone. For each batch size, the total time for \model{} is calculated as the sum of the time taken to compute the I-features and the time to process the P-features.

The graph indicates that our \model{} is more time-efficient than the ViT baseline, benefiting from our design optimized for efficient video encoding. In practice, our architecture requires roughly 2.5 times less wall-clock time to encode frames into features across most batch sizes. Additionally, when the encoding time is constrained, \eg $10$ seconds, the baseline model can process a batch size of approximately $700$ frames, whereas \model{} can handle a batch size of about $1700$ frames.

Note that, in the regime of small batch sizes (\ie, $\leq 8$), highlighted in the zoomed box in Figure~\ref{fig:time-comparison}, the ViT model proves more economical compared to \model{}. Nonetheless, it is crucial to remember that our focus is on efficiently encoding numerous video frames for dense tasks, making \model{} the preferred choice under these conditions.

These experiments were performed using a single NVIDIA V100 GPU. Timings for each batch size were obtained by averaging results from $100$ consecutive forward passes to ensure statistical reliability. To guarantee precise timing measurements, we employed the PyTorch function \texttt{torch.cuda.synchronize()}, which halts the execution of the code until all pending GPU operations are completed. This function is critical for avoiding discrepancies in timing due to asynchronous GPU execution.

\section{Additional Task: Action Recognition}
\label{sec:kinetics}

In this section, we evaluate the task of action recognition by examining the accuracy gap between CLIP and \model{} ($N=2$, $p=85\%$) features, using the ViT-B/32 backbone for both models. The experiments are conducted on the Kinetics-400 dataset~\cite{kay2017kinetics} in a zero-shot setting.

The accuracy comparisons are presented in Table~\ref{tab:kinetics}. In particular, we investigate the accuracy trends and total encoding costs as the number of frames increases.

We observe that \model{} delivers competitive accuracy compared to CLIP features, with a minimum gap of $0.8\%$ for Accuracy@1 at $3$ frames and a maximum gap of approximately $3.2\%$ for Accuracy@1 at $4$ frames. Similar trends are observed for Accuracy@5. However, when analyzing the accuracy versus total encoding cost, \model{} demonstrates a clear advantage: with $4$ frames and a total cost of $12.8$ GFLOPs, it outperforms CLIP with $3$ frames and a total cost of $13.2$ GFLOPs for both Accuracy@1 and Accuracy@5. Furthermore, \model{} with $7$ frames and a total encoding cost of $21.2$ GFLOPs achieves nearly identical accuracy to CLIP with $5$ frames, which has a higher total cost of $22.0$ GFLOPs.

For the zero-shot setup of this experiment, each frame is encoded using either CLIP or \model{}, and the resulting visual feature representations are averaged. Classification is performed by combining the class labels with prompt templates provided by the CLIP baseline\footnotemark and encoding the text using the language encoder. All prompt features per class are then averaged, and cosine similarity between the visual and text representations for each class is computed. The classes are ranked by their similarity scores, and the accuracy metric is computed accordingly.
\footnotetext{Prompt templates can be found here: \url{https://github.com/openai/CLIP/blob/main/data/prompts.md##kinetics700}}
\begin{table}[!t]
\centering
\setlength{\tabcolsep}{2pt}
\resizebox{\linewidth}{!}{%
\scalebox{1.0}{
\begin{tabular}{c|ccc|ccc}
    
    \toprule\toprule

    Number 
    & \multicolumn{2}{c}{CLIP}
    & Total encoding 
    & \multicolumn{2}{c}{\model{}}
    & Total encoding \\
    
    of Frames
    & Acc@1  $\uparrow$
    & Acc@5  $\uparrow$
    & cost (GFLOPS)
    & Acc@1  $\uparrow$
    & Acc@5  $\uparrow$
    & cost (GFLOPS)  \\
    
    \midrule\midrule
    
    1 & $44.5$ & $72.3$ & $4.4$ & $44.5$ & $72.3$ & $4.4$ \\ 
    2 & $45.0$ & $73.0$ & $8.8$ & $43.4$ & $71.1$ & $6.4$ \\ 
    3 & $43.9$ & $71.5$ & $13.2$ & $43.1$ & $70.7$ & $8.4$ \\ 
    4 & $48.1$ & $76.0$ & $17.6$ & $44.8$ & $73.0$ & $12.8$ \\ 
    5 & $46.5$ & $74.5$ & $22.0$ & $45.1$ & $73.5$ & $14.8$ \\ 
    6 & $48.7$ & $76.8$ & $26.4$ & $45.5$ & $73.9$ & $16.8$ \\ 
    7 & $47.6$ & $75.9$ & $30.8$ & $44.4$ & $72.9$ & $21.2$ \\ 
    8 & $49.3$ & $77.1$ & $35.2$ & $46.5$ & $74.9$ & $23.2$ \\ 
    9 & $48.2$ & $76.7$ & $39.6$ & $46.2$ & $74.8$ & $25.2$ \\ 
    10 & $49.3$ & $77.4$ & $44.0$ & $46.5$ & $75.1$ & $29.6$ \\
    
    \bottomrule \bottomrule
\end{tabular}
}
}
\vspace{-.1cm}
\caption{\textbf{Action Recognition.} We report accuracy at 1 (Acc@1) and accuracy at 5 (Acc@5) for CLIP and \model{} ($N=2$, $p=85\%$) features on the Kinetics 400~\citep{kay2017kinetics} dataset under a zero-shot setting.}
\label{tab:kinetics}
\end{table}

\section{Limitations and Discussion}
\label{sec:limitations}
We acknowledge several technical limitations of our approach. First, our method is specifically designed for the Vision Transformer (ViT) architecture, making it less applicable to other architectures, such as convolutional or recurrent neural networks. Nonetheless, we argue that transformer-based models have proven to be among the most versatile and scalable options in the deep learning landscape, supporting their continued adoption and adaptation.

Second, \model{} is optimized for dense video processing tasks, which may limit its efficacy in scenarios that benefit from sparse frame sampling, such as action recognition or video retrieval. For such applications, the semantic continuity captured by the residual token across temporally distant frames may not be sufficient, suggesting a potential area for future research.

Third, our token drop strategy relies on motion information derived from the video’s compressed representation, which may fail to capture subtle or small-scale movements, such as those occurring in crowded scenes or during fine-grained interactions. 
This limitation highlights the need for more nuanced motion-aware selection strategies in future work.

Fourth, our method is tailored to reduce the computational burden of video frame encoding in temporally dense tasks. Other aspects of fine-grained video understanding (\eg, precise spatial localization) remain open directions for future investigation.

Lastly, our solution's effectiveness heavily relies on the quality of the underlying large pre-trained foundation model, such as CLIP~\cite{radford2021learning}. Consequently, any inherent biases or limitations in the pre-trained model's weights could adversely affect our method's accuracy.

Note that we hypothesize that the WebVid training does not introduce new information beyond what is already in CLIP as the distillation trains only the residual tokenizer so that \model{} approximates CLIP features (Section\ 3.2). Moreover, less than $0.3\%$ of the parameters differ between the two models. We leave for future exploration the full fine-tuning of CLIP on WebVid which has the potential to improve accuracy on downstream tasks. Nonetheless, our approach can seamlessly benefit from  this fine-tuned representation instead of CLIP.

 \begin{figure*}[!t]
    \centering
    \begin{subfigure}[b]{.95\textwidth}
        \centering
        \includegraphics[width=\textwidth, trim={.5cm 2cm .5cm 2cm}, clip]{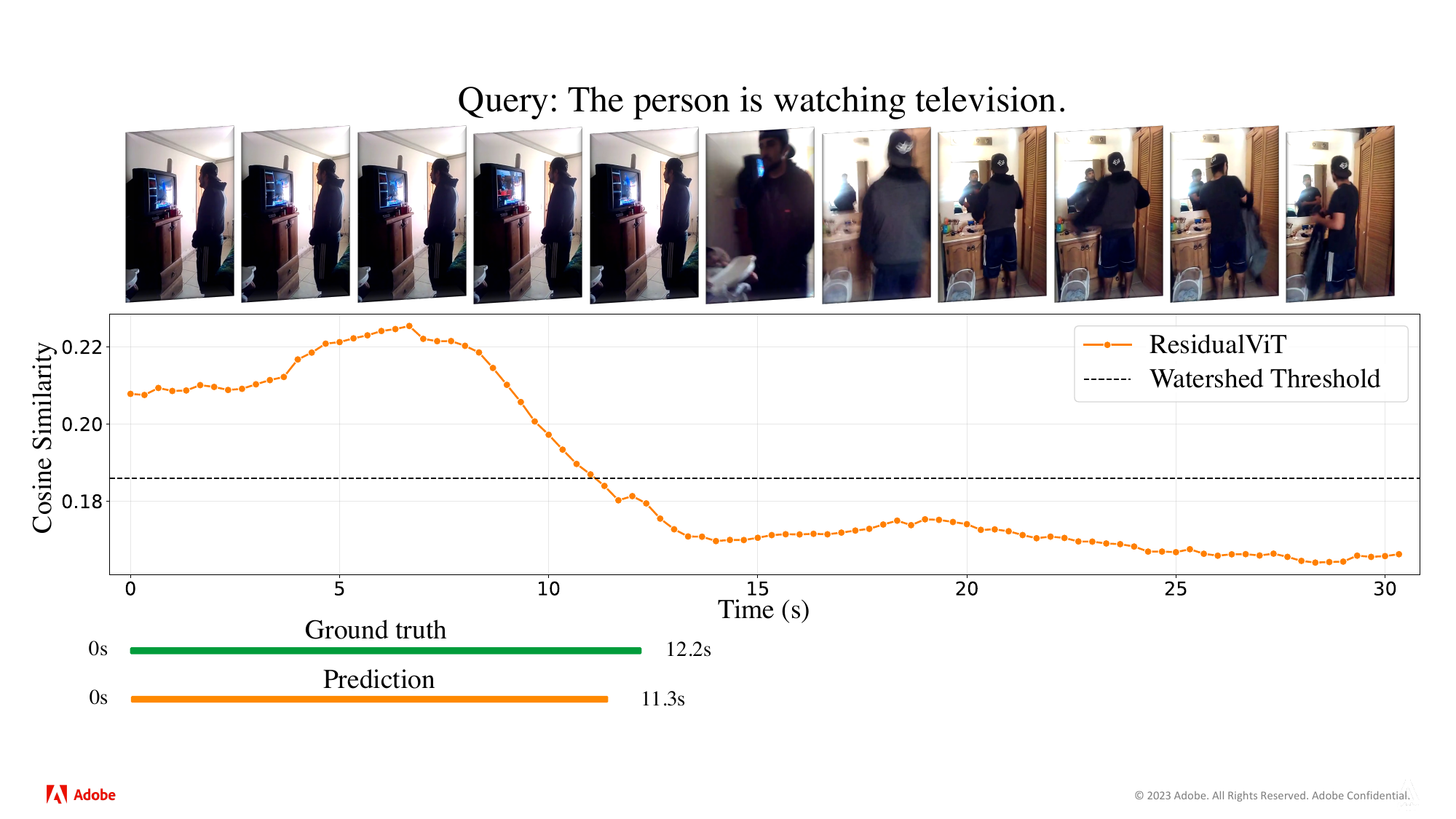}
        \caption{\textbf{Grounding example.} We observe an IoU $=0.93$ between the ground truth moment and the predicted one.}
        \label{fig:qualitative1a}
    \end{subfigure}
    \begin{subfigure}[b]{\textwidth}
        \centering
        \includegraphics[width=\textwidth, trim={.5cm 2cm .1cm 1cm}, clip]{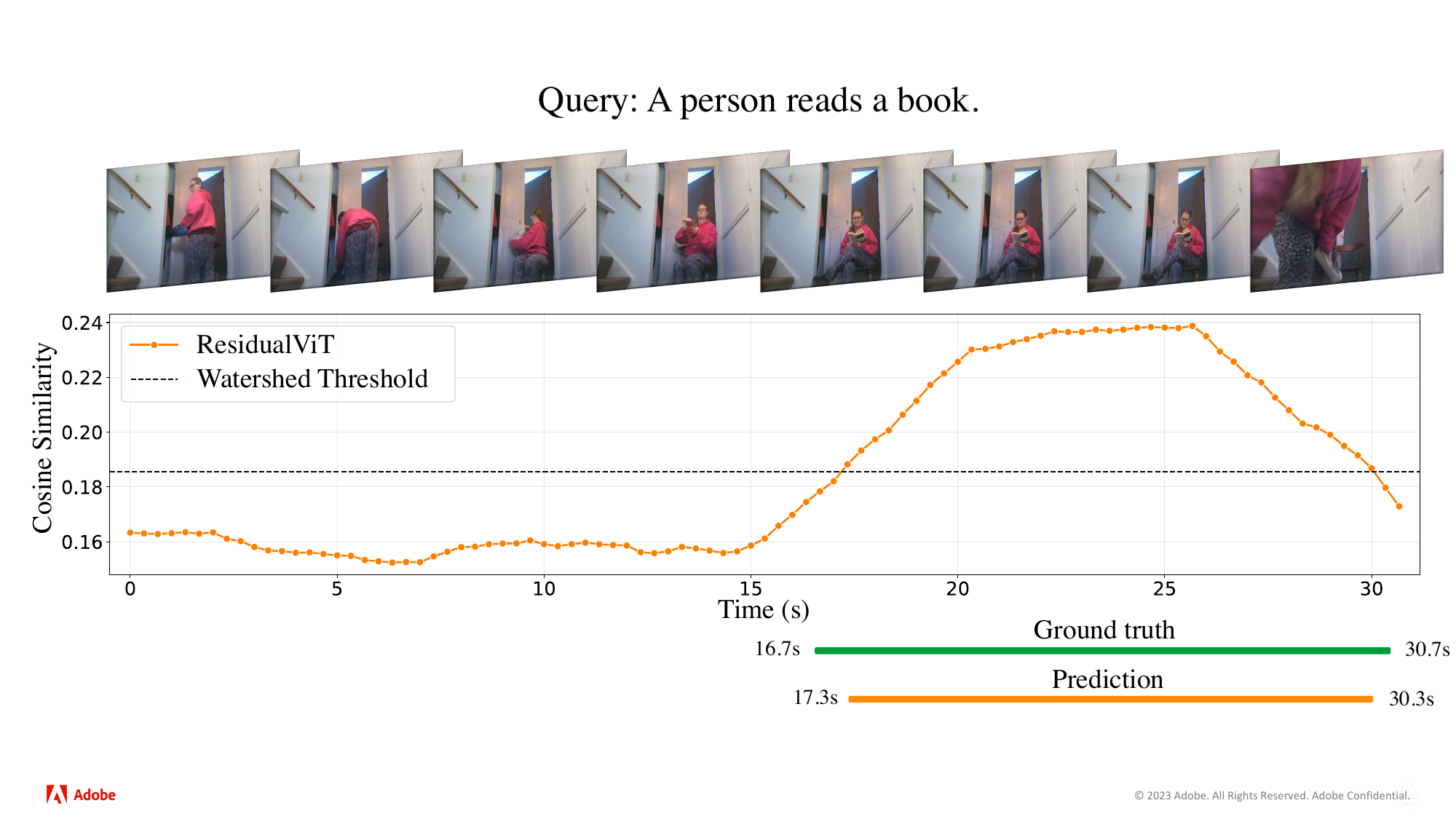}
        \caption{\textbf{Grounding example.} We observe an IoU $=0.93$ between the ground truth moment and the predicted one.}
        \label{fig:qualitative1b}
    \end{subfigure}
    \caption{\textbf{Qualitative results.} We present two different examples in which our zero-shot algorithm can effectively ground the sentence in the video. We showcase the comparison between the ground truth annotation (\textcolor{GTgreen}{\textbf{green}}) and our top-1 prediction (\textcolor{orange}{\textbf{orange}}). 
    \label{fig:qualitative1}}
\end{figure*}

   \begin{figure*}[!t]
    \centering
    \begin{subfigure}[b]{.95\textwidth}
        \centering
        \includegraphics[width=\textwidth, trim={.5cm 2cm .5cm 2cm}, clip]{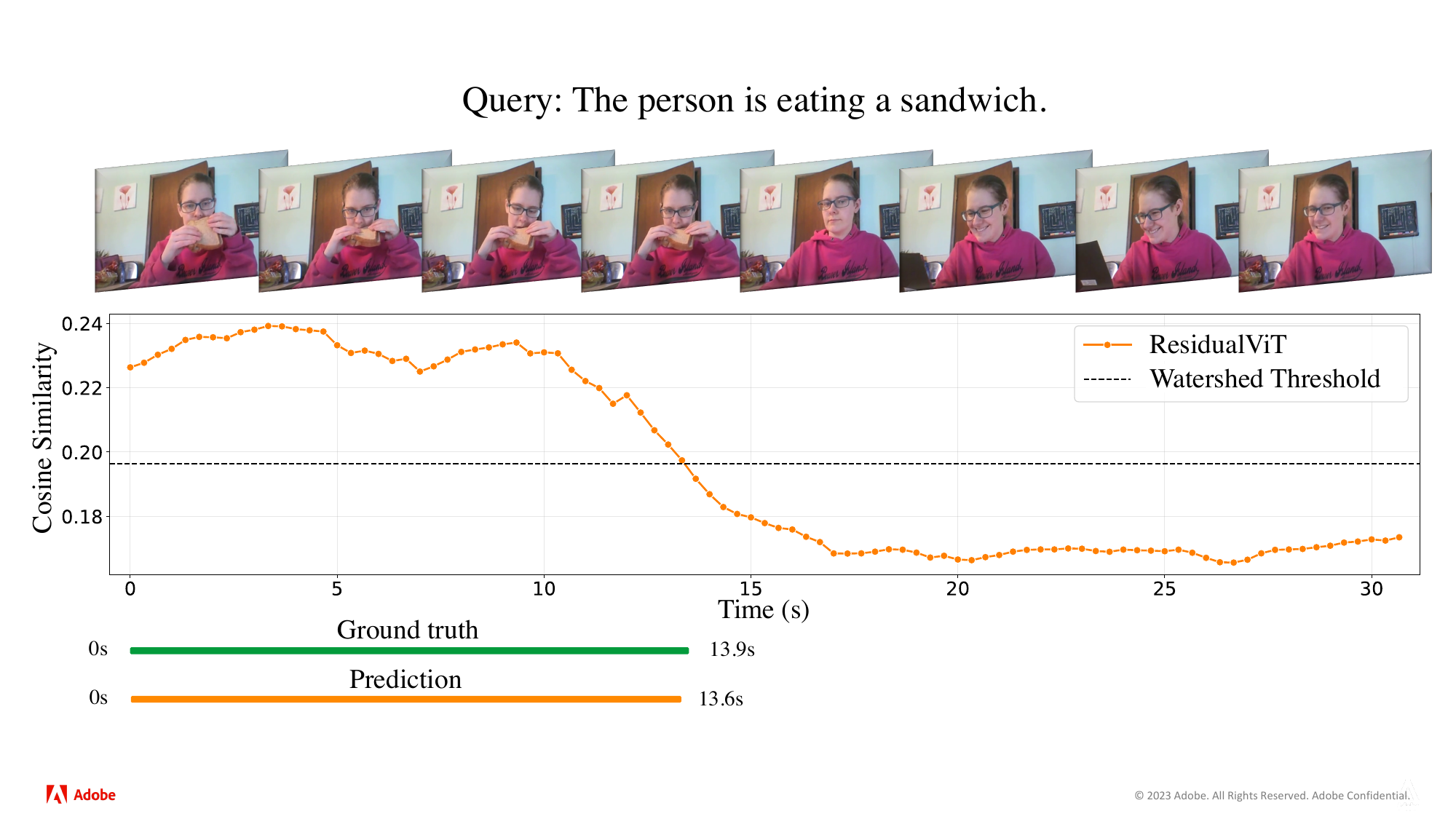}
        \caption{\textbf{Grounding example.} We observe an IoU $=0.98$ between the ground truth moment and the predicted one.}
        \label{fig:qualitative2a}
    \end{subfigure}
    \begin{subfigure}[b]{\textwidth}
        \centering
        \includegraphics[width=\textwidth, trim={.5cm 2cm .5cm 1cm}, clip]{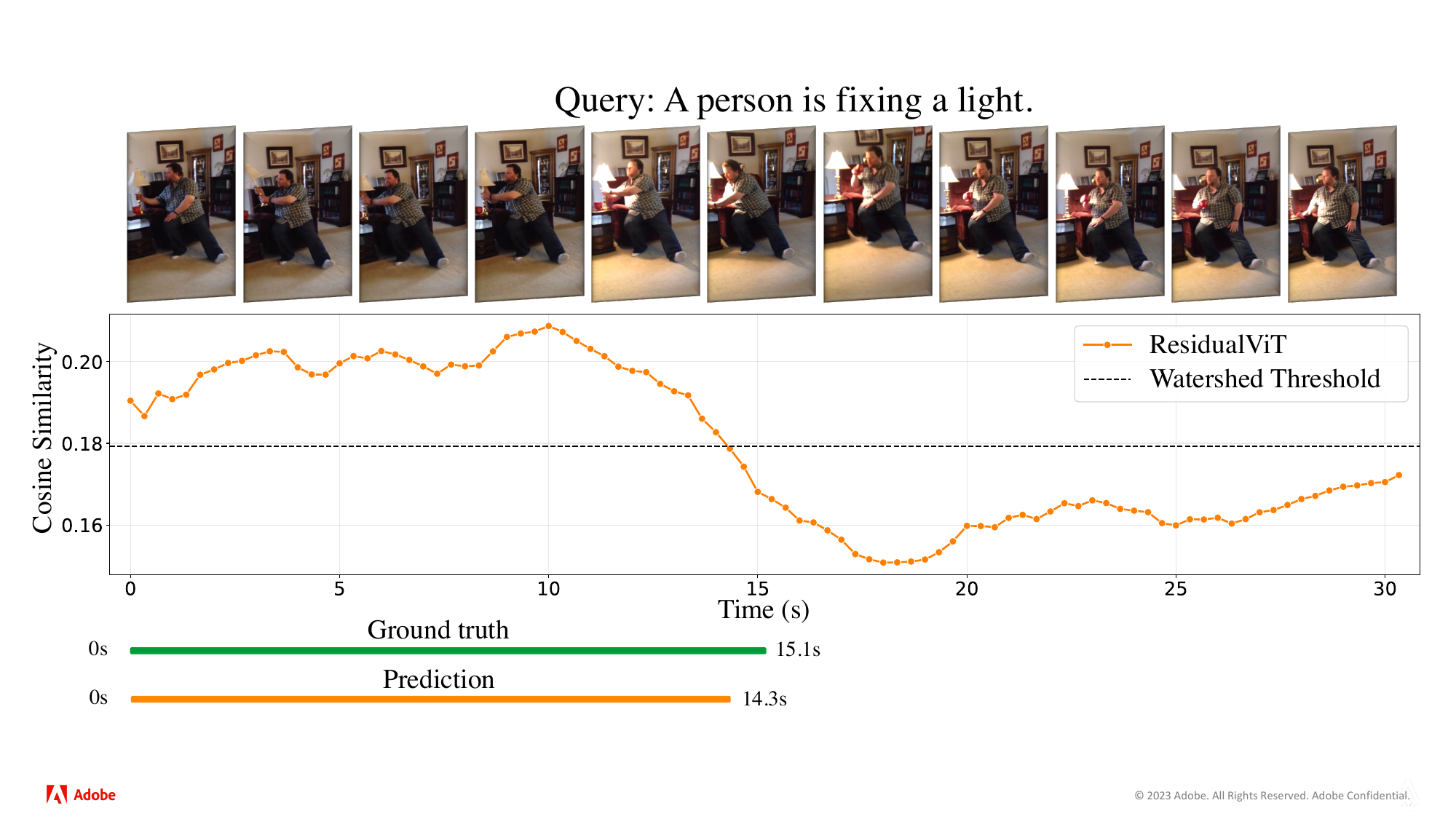}
        \caption{\textbf{Grounding example.} We observe an IoU $=0.95$ between the ground truth moment and the predicted one.}
        \label{fig:qualitative2b}
    \end{subfigure}
    \caption{\textbf{Qualitative results.} We present two different examples in which our zero-shot algorithm can effectively ground the sentence in the video. We showcase the comparison between the ground truth annotation (\textcolor{GTgreen}{\textbf{green}}) and our top-1 prediction (\textcolor{orange}{\textbf{orange}}). 
    \label{fig:qualitative2}}
\end{figure*}

\begin{figure*}[!t]
    \centering
    \begin{subfigure}[b]{.95\textwidth}
        \centering
        \includegraphics[width=\textwidth, trim={.5cm 2cm .5cm 1cm}, clip]{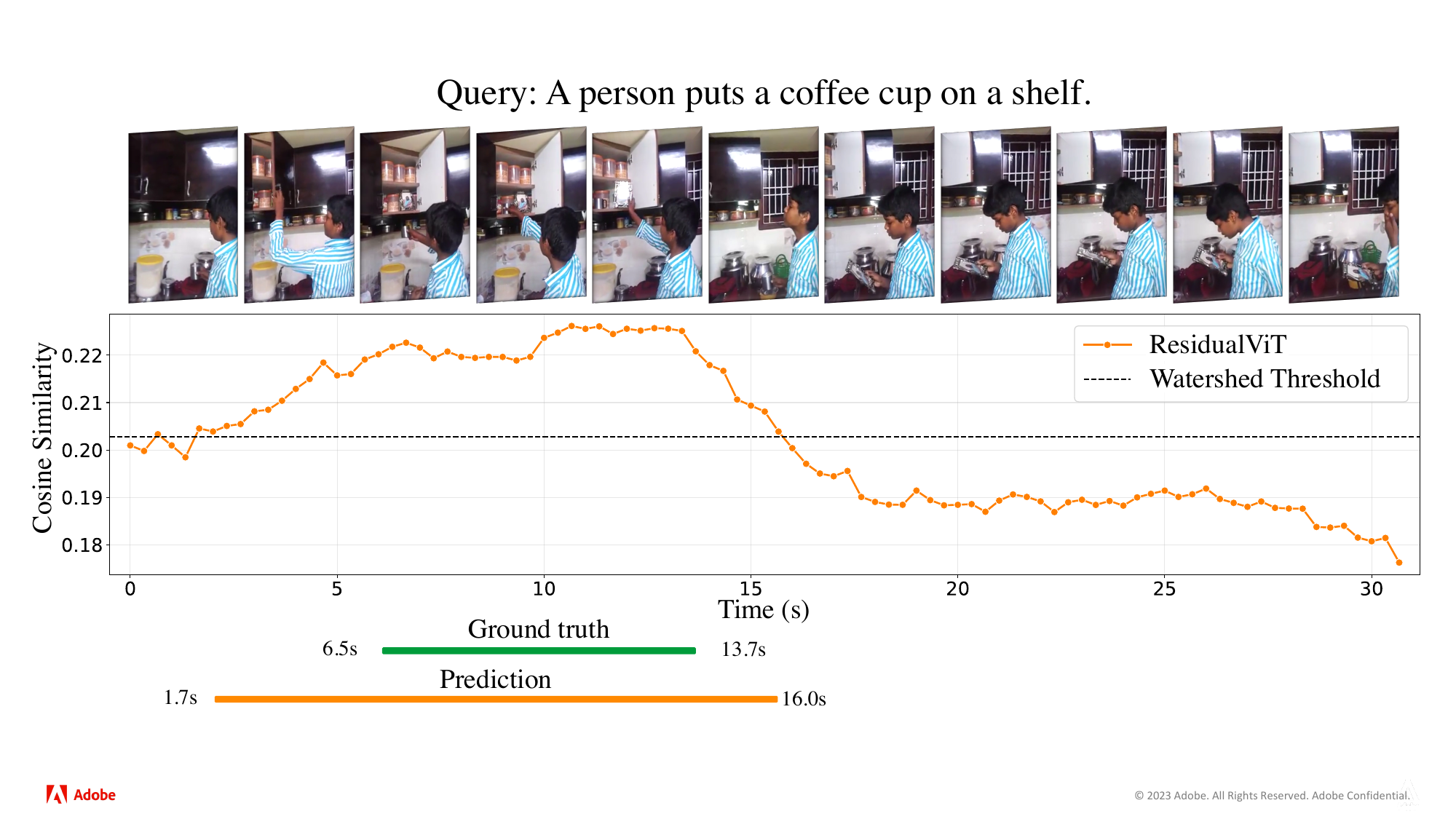}
        \caption{\textbf{Grounding example.} An IoU $=0.5$ is observed between the temporal annotation and our prediction.}
        \label{fig:qualitative3a}
    \end{subfigure}
    \begin{subfigure}[b]{\textwidth}
        \centering
        \includegraphics[width=\textwidth, trim={.5cm 2cm .5cm 1cm}, clip]{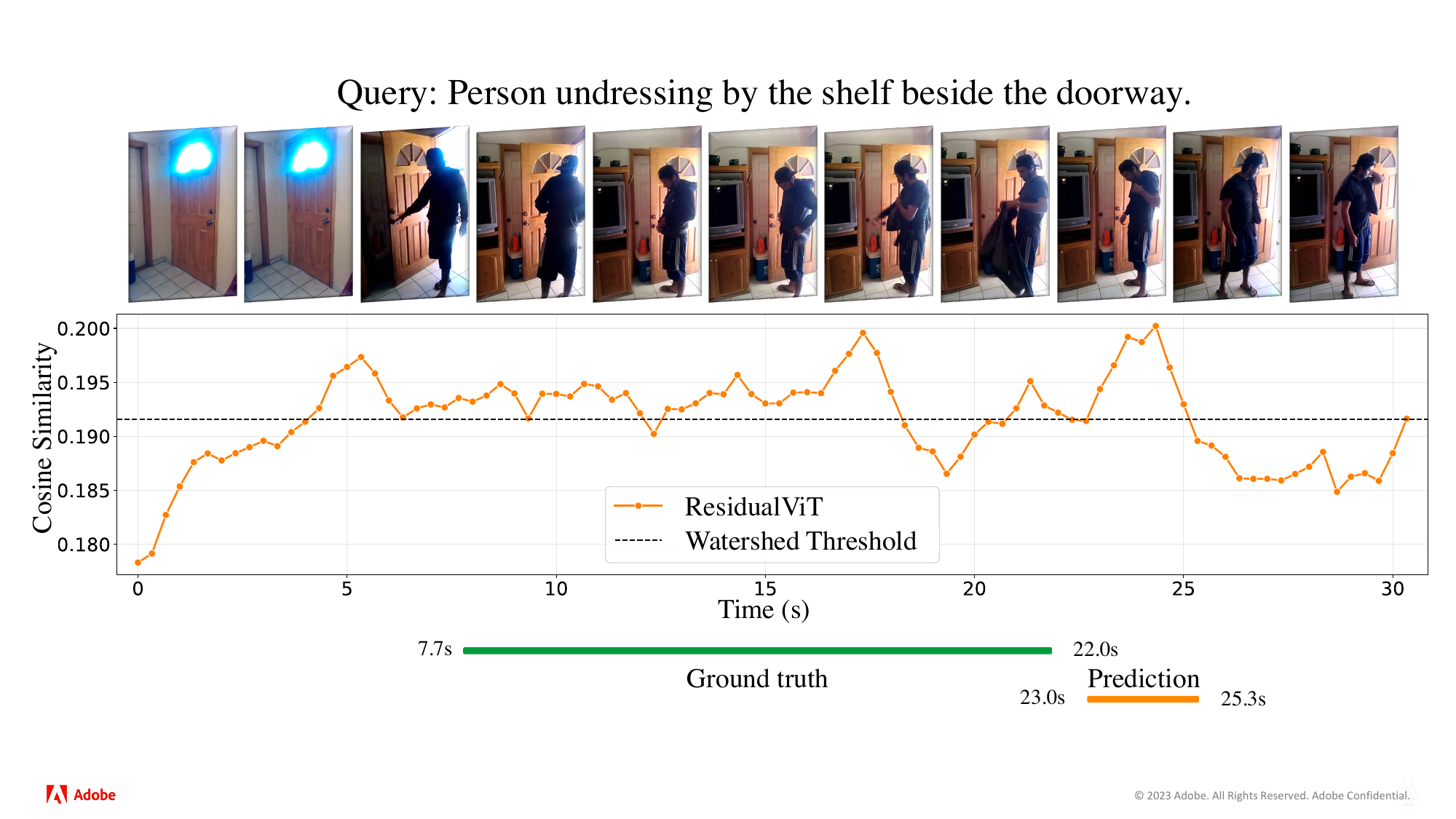}
        \caption{\textbf{Grounding example.} Our prediction does not overlap with the ground truth moment.}
        \label{fig:qualitative3b}
    \end{subfigure}
    \caption{\textbf{Qualitative results.} We present two different examples in which our zero-shot algorithm can effectively ground the sentence in the video. We showcase the comparison between the ground truth annotation (\textcolor{GTgreen}{\textbf{green}}) and our top-1 prediction (\textcolor{orange}{\textbf{orange}}). 
    \label{fig:qualitative3}}
\end{figure*}

\section{Qualitative Results of Natural Language Video Grounding}
\label{sec:supplementary-qualitative}
In Figure~\ref{fig:qualitative1}-\ref{fig:qualitative3}, we present a series of qualitative results from the Charades-STA dataset, demonstrating the efficacy of our zero-shot grounding baseline in identifying relevant event boundaries within video content.
In each example, we first show a subset of the video frames along with the textual query on top. Then, we illustrate the temporal sequence of similarity scores $\{S_{t}\}_{t=1}^{n_v}$ produced by computing the cosine similarity between each frame feature and the sentence feature. We also show the watershed threshold, which is used to determine the start and end moment predictions as detailed in Section~\ref{sec:grounding-algorithm}. 
For each example, the figure also illustrates the top-1 predicted temporal segment (\textcolor{orange}{\textbf{orange}}) and the ground truth annotation (\textcolor{GTgreen}{\textbf{green}}).

In the examples depicted in Figures~\ref{fig:qualitative1}\textcolor{navyblue}{(a-b)} and \ref{fig:qualitative2}\textcolor{navyblue}{(a-b)}, our algorithm is capable of discriminating subtle frame differences and produces very precise temporal boundaries that provide an IoU $>0.9$ with the ground truth. In example~\ref{fig:qualitative1a}, the feature representations of the frames and the sentence provide higher similarity when the television is present, in accordance with the query \textit{``The person is watching television''}. Similarly, in example~\ref{fig:qualitative1b}, the algorithm can distinguish whether the person is holding a book despite the high resemblance among all frames, correctly predicting the temporal span relative to the textual query \textit{``A person reads a book''} with IoU $=0.93$. The cosine similarity profile in example~\ref{fig:qualitative2a} clearly differentiates between the section of the video in which the person is eating a sandwich and when they are simply smiling at the camera, predicting the grounding of the action  \textit{``A person is eating a sandwich''}, achieving IoU $=0.98$. 
Example~\ref{fig:qualitative2b} presents a challenging scenario,\textit{``A person is fixing a light''}, where the model needs to recognize the light's transition from off to on. Despite these complexities, our method provides a correct prediction with an IoU $=0.95$.

Nonetheless, our approach can provide meaningful predictions that, however, do not align well with the ground truth moment. We detail one such example in Figure~\ref{fig:qualitative3a}. For the query  \textit{``A person puts a coffee cup on a shelf''}, we predict a temporal span that is correctly centered to the ground truth span but is twice as long as the ground truth moment, yielding an IoU of approximately $0.5$. However, if we pay attention to the video frames, one could argue the prediction is still correct, as it begins when the person opens the cabinet and finishes after the person has placed the coffee cup in it. 

Lastly, in Figure~\ref{fig:qualitative3b}, we depict an example in which our proposed solution fails. The action described by the query \textit{``Person undressing by the shelf beside the doorway''} shows a long duration, effectively producing a high similarity response for a good part of the video. This, in turn, affects the watershed threshold, which is proportional to the average similarity scores. Due to the high value of the threshold, our algorithm produces an incorrect prediction that does not overlap with the ground truth.

\newpage
\thispagestyle{empty}
\mbox{}                

\newpage
\thispagestyle{empty}
\mbox{}

\newpage
\thispagestyle{empty}
\mbox{}

\newpage
\thispagestyle{empty}
\mbox{}

\newpage
\thispagestyle{empty}
\mbox{}

\newpage
\thispagestyle{empty}
\mbox{}

\newpage
{
    \small
    \bibliographystyle{ieeenat_fullname}
    \bibliography{main}

\begin{thebibliography}{107}
\providecommand{\natexlab}[1]{#1}
\providecommand{\url}[1]{\texttt{#1}}
\expandafter\ifx\csname urlstyle\endcsname\relax
  \providecommand{\doi}[1]{doi: #1}\else
  \providecommand{\doi}{doi: \begingroup \urlstyle{rm}\Url}\fi

\bibitem[AcherStyx(2020)]{compressedvideoreader}
AcherStyx.
\newblock Compressed video reader, 2020.
\newblock Accessed: 2024-02.

\bibitem[Anderson et~al.(2016)Anderson, Fernando, Johnson, and Gould]{anderson2016spice}
Peter Anderson, Basura Fernando, Mark Johnson, and Stephen Gould.
\newblock Spice: Semantic propositional image caption evaluation.
\newblock In \emph{Computer Vision--ECCV 2016: 14th European Conference, Amsterdam, The Netherlands, October 11-14, 2016, Proceedings, Part V 14}, pages 382--398. Springer, 2016.

\bibitem[Bain et~al.(2021)Bain, Nagrani, Varol, and Zisserman]{bain2021frozen}
Max Bain, Arsha Nagrani, G{\"u}l Varol, and Andrew Zisserman.
\newblock Frozen in time: A joint video and image encoder for end-to-end retrieval.
\newblock In \emph{Proceedings of the IEEE/CVF International Conference on Computer Vision}, pages 1728--1738, 2021.

\bibitem[Banerjee and Lavie(2005)]{banerjee2005meteor}
Satanjeev Banerjee and Alon Lavie.
\newblock Meteor: An automatic metric for mt evaluation with improved correlation with human judgments.
\newblock In \emph{Proceedings of the acl workshop on intrinsic and extrinsic evaluation measures for machine translation and/or summarization}, pages 65--72, 2005.

\bibitem[Barrios et~al.(2023)Barrios, Soldan, Ceballos-Arroyo, Heilbron, and Ghanem]{barrios2023localizing}
Wayner Barrios, Mattia Soldan, Alberto~Mario Ceballos-Arroyo, Fabian~Caba Heilbron, and Bernard Ghanem.
\newblock Localizing moments in long video via multimodal guidance.
\newblock In \emph{Proceedings of the IEEE/CVF International Conference on Computer Vision}, pages 13667--13678, 2023.

\bibitem[Bolya et~al.(2022)Bolya, Fu, Dai, Zhang, Feichtenhofer, and Hoffman]{bolya2022token}
Daniel Bolya, Cheng-Yang Fu, Xiaoliang Dai, Peizhao Zhang, Christoph Feichtenhofer, and Judy Hoffman.
\newblock Token merging: Your vit but faster.
\newblock \emph{arXiv preprint arXiv:2210.09461}, 2022.

\bibitem[Buch et~al.(2022)Buch, Eyzaguirre, Gaidon, Wu, Fei-Fei, and Niebles]{buch2022revisiting}
Shyamal Buch, Crist{\'o}bal Eyzaguirre, Adrien Gaidon, Jiajun Wu, Li Fei-Fei, and Juan~Carlos Niebles.
\newblock Revisiting the" video" in video-language understanding.
\newblock In \emph{Proceedings of the IEEE/CVF conference on computer vision and pattern recognition}, pages 2917--2927, 2022.

\bibitem[Caba~Heilbron et~al.(2015)Caba~Heilbron, Escorcia, Ghanem, and Carlos~Niebles]{caba2015activitynet}
Fabian Caba~Heilbron, Victor Escorcia, Bernard Ghanem, and Juan Carlos~Niebles.
\newblock Activitynet: A large-scale video benchmark for human activity understanding.
\newblock In \emph{Proceedings of the ieee conference on computer vision and pattern recognition}, pages 961--970, 2015.

\bibitem[Carreira and Zisserman(2017)]{carreira2017quo}
Joao Carreira and Andrew Zisserman.
\newblock Quo vadis, action recognition? a new model and the kinetics dataset.
\newblock In \emph{proceedings of the IEEE Conference on Computer Vision and Pattern Recognition}, pages 6299--6308, 2017.

\bibitem[Castro and Heilbron(2022)]{castro2022fitclip}
Santiago Castro and Fabian~Caba Heilbron.
\newblock Fitclip: Refining large-scale pretrained image-text models for zero-shot video understanding tasks.
\newblock \emph{arXiv preprint arXiv:2203.13371}, 2022.

\bibitem[Chen et~al.(2020)Chen, Ma, Luo, Tang, and Wong]{chen2020look}
Zhenfang Chen, Lin Ma, Wenhan Luo, Peng Tang, and Kwan-Yee~K Wong.
\newblock Look closer to ground better: Weakly-supervised temporal grounding of sentence in video.
\newblock \emph{arXiv preprint arXiv:2001.09308}, 2020.

\bibitem[Dehghani et~al.(2023)Dehghani, Djolonga, Mustafa, Padlewski, Heek, Gilmer, Steiner, Caron, Geirhos, Alabdulmohsin, Jenatton, Beyer, Tschannen, Arnab, Wang, Riquelme, Minderer, Puigcerver, Evci, Kumar, van Steenkiste, Elsayed, Mahendran, Yu, Oliver, Huot, Bastings, Collier, Gritsenko, Birodkar, Vasconcelos, Tay, Mensink, Kolesnikov, Pavetić, Tran, Kipf, Lučić, Zhai, Keysers, Harmsen, and Houlsby]{dehghani2023scaling}
Mostafa Dehghani, Josip Djolonga, Basil Mustafa, Piotr Padlewski, Jonathan Heek, Justin Gilmer, Andreas Steiner, Mathilde Caron, Robert Geirhos, Ibrahim Alabdulmohsin, Rodolphe Jenatton, Lucas Beyer, Michael Tschannen, Anurag Arnab, Xiao Wang, Carlos Riquelme, Matthias Minderer, Joan Puigcerver, Utku Evci, Manoj Kumar, Sjoerd van Steenkiste, Gamaleldin~F. Elsayed, Aravindh Mahendran, Fisher Yu, Avital Oliver, Fantine Huot, Jasmijn Bastings, Mark~Patrick Collier, Alexey Gritsenko, Vighnesh Birodkar, Cristina Vasconcelos, Yi Tay, Thomas Mensink, Alexander Kolesnikov, Filip Pavetić, Dustin Tran, Thomas Kipf, Mario Lučić, Xiaohua Zhai, Daniel Keysers, Jeremiah Harmsen, and Neil Houlsby.
\newblock Scaling vision transformers to 22 billion parameters.
\newblock \emph{arXiv preprint arXiv:2302.05442}, 2023.

\bibitem[Ding et~al.(2023)Ding, Zhao, Zhang, Qian, Xiong, and Tian]{ding2023prune}
Shuangrui Ding, Peisen Zhao, Xiaopeng Zhang, Rui Qian, Hongkai Xiong, and Qi Tian.
\newblock Prune spatio-temporal tokens by semantic-aware temporal accumulation.
\newblock In \emph{Proceedings of the IEEE/CVF International Conference on Computer Vision}, pages 16945--16956, 2023.

\bibitem[Diwan et~al.(2023)Diwan, Peng, and Mooney]{diwan2023zero}
Anuj Diwan, Puyuan Peng, and Ray Mooney.
\newblock Zero-shot video moment retrieval with off-the-shelf models.
\newblock In \emph{Transfer Learning for Natural Language Processing Workshop}, pages 10--21. PMLR, 2023.

\bibitem[Dosovitskiy et~al.(2020)Dosovitskiy, Beyer, Kolesnikov, Weissenborn, Zhai, Unterthiner, Dehghani, Minderer, Heigold, Gelly, et~al.]{dosovitskiy2020image}
Alexey Dosovitskiy, Lucas Beyer, Alexander Kolesnikov, Dirk Weissenborn, Xiaohua Zhai, Thomas Unterthiner, Mostafa Dehghani, Matthias Minderer, Georg Heigold, Sylvain Gelly, et~al.
\newblock An image is worth 16x16 words: Transformers for image recognition at scale.
\newblock \emph{arXiv preprint arXiv:2010.11929}, 2020.

\bibitem[Escorcia et~al.(2019)Escorcia, Soldan, Sivic, Ghanem, and Russell]{escorcia2019finding}
Victor Escorcia, Mattia Soldan, Josef Sivic, Bernard Ghanem, and Bryan Russell.
\newblock Finding moments in video collections using natural language.
\newblock \emph{arXiv preprint arXiv:1907.12763}, 2019.

\bibitem[FAIR(2020)]{fvcore}
FAIR.
\newblock fvcore, 2020.
\newblock Accessed: 2024-02.

\bibitem[Fang et~al.(2023)Fang, Ma, and Wang]{fang2023structural}
Gongfan Fang, Xinyin Ma, and Xinchao Wang.
\newblock Structural pruning for diffusion models.
\newblock In \emph{Advances in Neural Information Processing Systems}, 2023.

\bibitem[Feichtenhofer(2020)]{Feichtenhofer_2020_CVPR}
Christoph Feichtenhofer.
\newblock X3d: Expanding architectures for efficient video recognition.
\newblock In \emph{Proceedings of the IEEE/CVF Conference on Computer Vision and Pattern Recognition (CVPR)}, 2020.

\bibitem[Feichtenhofer et~al.(2019)Feichtenhofer, Fan, Malik, and He]{feichtenhofer2019slowfast}
Christoph Feichtenhofer, Haoqi Fan, Jitendra Malik, and Kaiming He.
\newblock Slowfast networks for video recognition.
\newblock In \emph{Proceedings of the IEEE/CVF international conference on computer vision}, pages 6202--6211, 2019.

\bibitem[Gao and Xu(2021)]{gao2021learning}
Junyu Gao and Changsheng Xu.
\newblock Learning video moment retrieval without a single annotated video.
\newblock \emph{IEEE Transactions on Circuits and Systems for Video Technology}, 32\penalty0 (3):\penalty0 1646--1657, 2021.

\bibitem[Gao et~al.(2017)Gao, Sun, Yang, and Nevatia]{gao2017tall}
Jiyang Gao, Chen Sun, Zhenheng Yang, and Ram Nevatia.
\newblock Tall: Temporal activity localization via language query.
\newblock In \emph{Proceedings of the IEEE international conference on computer vision}, pages 5267--5275, 2017.

\bibitem[Gorti et~al.(2022)Gorti, Vouitsis, Ma, Golestan, Volkovs, Garg, and Yu]{gorti2022x}
Satya~Krishna Gorti, No{\"e}l Vouitsis, Junwei Ma, Keyvan Golestan, Maksims Volkovs, Animesh Garg, and Guangwei Yu.
\newblock X-pool: Cross-modal language-video attention for text-video retrieval.
\newblock In \emph{Proceedings of the IEEE/CVF conference on computer vision and pattern recognition}, pages 5006--5015, 2022.

\bibitem[Han et~al.(2023)Han, Bain, Nagrani, Varol, Xie, and Zisserman]{han2023autoad}
Tengda Han, Max Bain, Arsha Nagrani, G{\"u}l Varol, Weidi Xie, and Andrew Zisserman.
\newblock Autoad: Movie description in context.
\newblock In \emph{Proceedings of the IEEE/CVF Conference on Computer Vision and Pattern Recognition}, pages 18930--18940, 2023.

\bibitem[Hannan et~al.(2023)Hannan, Islam, Seidl, and Bertasius]{hannan2023rgnet}
Tanveer Hannan, Md~Mohaiminul Islam, Thomas Seidl, and Gedas Bertasius.
\newblock Rgnet: A unified retrieval and grounding network for long videos.
\newblock \emph{arXiv preprint arXiv:2312.06729}, 2023.

\bibitem[Hao et~al.(2022)Hao, Guo, Jia, Han, Tang, Zhang, Hu, and Wang]{hao2022learning}
Zhiwei Hao, Jianyuan Guo, Ding Jia, Kai Han, Yehui Tang, Chao Zhang, Han Hu, and Yunhe Wang.
\newblock Learning efficient vision transformers via fine-grained manifold distillation.
\newblock \emph{Advances in Neural Information Processing Systems}, 35:\penalty0 9164--9175, 2022.

\bibitem[Haurum et~al.(2023)Haurum, Escalera, Taylor, and Moeslund]{haurum2023tokens}
Joakim~Bruslund Haurum, Sergio Escalera, Graham~W Taylor, and Thomas~B Moeslund.
\newblock Which tokens to use? investigating token reduction in vision transformers.
\newblock In \emph{Proceedings of the IEEE/CVF International Conference on Computer Vision}, pages 773--783, 2023.

\bibitem[He et~al.(2016)He, Zhang, Ren, and Sun]{he2016deep}
Kaiming He, Xiangyu Zhang, Shaoqing Ren, and Jian Sun.
\newblock Deep residual learning for image recognition.
\newblock In \emph{Proceedings of the IEEE conference on computer vision and pattern recognition}, pages 770--778, 2016.

\bibitem[He et~al.(2017)He, Zhang, and Sun]{he2017channel}
Yihui He, Xiangyu Zhang, and Jian Sun.
\newblock Channel pruning for accelerating very deep neural networks.
\newblock In \emph{Proceedings of the IEEE international conference on computer vision}, pages 1389--1397, 2017.

\bibitem[Hendricks et~al.(2017)Hendricks, Wang, Shechtman, Sivic, Darrell, and Russell]{anne2017localizing}
Lisa Hendricks, Oliver Wang, Eli Shechtman, Josef Sivic, Trevor Darrell, and Bryan Russell.
\newblock Localizing moments in video with natural language.
\newblock In \emph{Proceedings of the IEEE international conference on computer vision}, pages 5803--5812, 2017.

\bibitem[Heo et~al.(2019)Heo, Kim, Yun, Park, Kwak, and Choi]{heo2019comprehensive}
Byeongho Heo, Jeesoo Kim, Sangdoo Yun, Hyojin Park, Nojun Kwak, and Jin~Young Choi.
\newblock A comprehensive overhaul of feature distillation.
\newblock In \emph{Proceedings of the IEEE/CVF International Conference on Computer Vision}, pages 1921--1930, 2019.

\bibitem[Hinton et~al.(2015)Hinton, Vinyals, and Dean]{hinton2015distilling}
Geoffrey Hinton, Oriol Vinyals, and Jeff Dean.
\newblock Distilling the knowledge in a neural network.
\newblock \emph{arXiv preprint arXiv:1503.02531}, 2015.

\bibitem[Holla and Lourentzou(2023)]{holla2023commonsense}
Meghana Holla and Ismini Lourentzou.
\newblock Commonsense for zero-shot natural language video localization.
\newblock \emph{arXiv preprint arXiv:2312.17429}, 2023.

\bibitem[Hou et~al.(2022{\natexlab{a}})Hou, Pang, Zhou, Wu, Song, Song, and Zhou]{hou2022token}
Le Hou, Richard~Yuanzhe Pang, Tianyi Zhou, Yuexin Wu, Xinying Song, Xiaodan Song, and Denny Zhou.
\newblock Token dropping for efficient bert pretraining.
\newblock \emph{arXiv preprint arXiv:2203.13240}, 2022{\natexlab{a}}.

\bibitem[Hou et~al.(2022{\natexlab{b}})Hou, Zhong, Ji, Gao, Yan, Chan, Ngo, Shou, and Duan]{hou2022cone}
Zhijian Hou, Wanjun Zhong, Lei Ji, Difei Gao, Kun Yan, Wing-Kwong Chan, Chong-Wah Ngo, Zheng Shou, and Nan Duan.
\newblock Cone: An efficient coarse-to-fine alignment framework for long video temporal grounding.
\newblock \emph{arXiv preprint arXiv:2209.10918}, 2022{\natexlab{b}}.

\bibitem[Howard et~al.(2017)Howard, Zhu, Chen, Kalenichenko, Wang, Weyand, Andreetto, and Adam]{howard2017mobilenets}
Andrew~G Howard, Menglong Zhu, Bo Chen, Dmitry Kalenichenko, Weijun Wang, Tobias Weyand, Marco Andreetto, and Hartwig Adam.
\newblock Mobilenets: Efficient convolutional neural networks for mobile vision applications.
\newblock \emph{arXiv preprint arXiv:1704.04861}, 2017.

\bibitem[Hu et~al.(2021)Hu, Shen, Wallis, Allen-Zhu, Li, Wang, Wang, and Chen]{hu2021lora}
Edward~J Hu, Yelong Shen, Phillip Wallis, Zeyuan Allen-Zhu, Yuanzhi Li, Shean Wang, Lu Wang, and Weizhu Chen.
\newblock Lora: Low-rank adaptation of large language models.
\newblock \emph{arXiv preprint arXiv:2106.09685}, 2021.

\bibitem[Huang et~al.(2021)Huang, Liu, Gong, and Jin]{huang2021cross}
Jiabo Huang, Yang Liu, Shaogang Gong, and Hailin Jin.
\newblock Cross-sentence temporal and semantic relations in video activity localisation.
\newblock In \emph{Proceedings of the IEEE/CVF international conference on computer vision}, pages 7199--7208, 2021.

\bibitem[Ilharco et~al.(2021)Ilharco, Wortsman, Wightman, Gordon, Carlini, Taori, Dave, Shankar, Namkoong, Miller, Hajishirzi, Farhadi, and Schmidt]{ilharco_gabriel_2021_5143773}
Gabriel Ilharco, Mitchell Wortsman, Ross Wightman, Cade Gordon, Nicholas Carlini, Rohan Taori, Achal Dave, Vaishaal Shankar, Hongseok Namkoong, John Miller, Hannaneh Hajishirzi, Ali Farhadi, and Ludwig Schmidt.
\newblock Openclip.
\newblock \url{https://doi.org/10.5281/zenodo.5143773}, 2021.
\newblock Version 0.1, Zenodo.

\bibitem[Ju et~al.(2022)Ju, Han, Zheng, Zhang, and Xie]{ju2022prompting}
Chen Ju, Tengda Han, Kunhao Zheng, Ya Zhang, and Weidi Xie.
\newblock Prompting visual-language models for efficient video understanding.
\newblock In \emph{European Conference on Computer Vision}, pages 105--124. Springer, 2022.

\bibitem[Kannojia and Jaiswal(2018)]{kannojia2018effects}
Suresh~Prasad Kannojia and Gaurav Jaiswal.
\newblock Effects of varying resolution on performance of cnn based image classification: An experimental study.
\newblock \emph{Int. J. Comput. Sci. Eng}, 6\penalty0 (9):\penalty0 451--456, 2018.

\bibitem[Kay et~al.(2017)Kay, Carreira, Simonyan, Zhang, Hillier, Vijayanarasimhan, Viola, Green, Back, Natsev, et~al.]{kay2017kinetics}
Will Kay, Joao Carreira, Karen Simonyan, Brian Zhang, Chloe Hillier, Sudheendra Vijayanarasimhan, Fabio Viola, Tim Green, Trevor Back, Paul Natsev, et~al.
\newblock The kinetics human action video dataset.
\newblock \emph{arXiv preprint arXiv:1705.06950}, 2017.

\bibitem[Kim et~al.(2023)Kim, Park, Lee, Park, and Sohn]{kim2023language}
Dahye Kim, Jungin Park, Jiyoung Lee, Seongheon Park, and Kwanghoon Sohn.
\newblock Language-free training for zero-shot video grounding.
\newblock In \emph{Proceedings of the IEEE/CVF Winter Conference on Applications of Computer Vision}, pages 2539--2548, 2023.

\bibitem[Koonce(2021)]{koonce2021mobilenetv3}
Brett Koonce.
\newblock Mobilenetv3.
\newblock In \emph{Convolutional neural networks with swift for tensorflow: image recognition and dataset categorization}, pages 125--144. Springer, 2021.

\bibitem[Krishna et~al.(2017)Krishna, Hata, Ren, Fei-Fei, and Carlos~Niebles]{Krishna_2017_ICCV}
Ranjay Krishna, Kenji Hata, Frederic Ren, Li Fei-Fei, and Juan Carlos~Niebles.
\newblock {Dense-Captioning Events in Videos}.
\newblock In \emph{Proceedings of the IEEE International Conference on Computer Vision (ICCV)}, 2017.

\bibitem[Lei et~al.(2021{\natexlab{a}})Lei, Berg, and Bansal]{lei2021detecting}
Jie Lei, Tamara~L Berg, and Mohit Bansal.
\newblock Detecting moments and highlights in videos via natural language queries.
\newblock \emph{Advances in Neural Information Processing Systems}, 34:\penalty0 11846--11858, 2021{\natexlab{a}}.

\bibitem[Lei et~al.(2021{\natexlab{b}})Lei, Berg, and Bansal]{moment-detr}
Jie Lei, Tamara~L Berg, and Mohit Bansal.
\newblock Detecting moments and highlights in videos via natural language queries.
\newblock In \emph{Advances in Neural Information Processing Systems}, pages 11846--11858. Curran Associates, Inc., 2021{\natexlab{b}}.

\bibitem[Li et~al.(2022)Li, Li, Xiong, and Hoi]{li2022blip}
Junnan Li, Dongxu Li, Caiming Xiong, and Steven Hoi.
\newblock Blip: Bootstrapping language-image pre-training for unified vision-language understanding and generation.
\newblock In \emph{International Conference on Machine Learning}, pages 12888--12900. PMLR, 2022.

\bibitem[Li et~al.(2021)Li, Guo, and Wang]{li2021proposal}
Kun Li, Dan Guo, and Meng Wang.
\newblock Proposal-free video grounding with contextual pyramid network.
\newblock In \emph{Proceedings of the AAAI Conference on Artificial Intelligence}, pages 1902--1910, 2021.

\bibitem[Lin(2004)]{lin2004rouge}
Chin-Yew Lin.
\newblock Rouge: A package for automatic evaluation of summaries.
\newblock In \emph{Text summarization branches out}, pages 74--81, 2004.

\bibitem[Lin et~al.(2019)Lin, Gan, and Han]{lin2019tsm}
Ji Lin, Chuang Gan, and Song Han.
\newblock Tsm: Temporal shift module for efficient video understanding.
\newblock In \emph{Proceedings of the IEEE/CVF international conference on computer vision}, pages 7083--7093, 2019.

\bibitem[Lin et~al.(2023{\natexlab{a}})Lin, Ahmed, Li, Lin, Azarnasab, Yang, Wang, Liang, Liu, Lu, et~al.]{lin2023mm}
Kevin Lin, Faisal Ahmed, Linjie Li, Chung-Ching Lin, Ehsan Azarnasab, Zhengyuan Yang, Jianfeng Wang, Lin Liang, Zicheng Liu, Yumao Lu, et~al.
\newblock Mm-vid: Advancing video understanding with gpt-4v (ision).
\newblock \emph{arXiv preprint arXiv:2310.19773}, 2023{\natexlab{a}}.

\bibitem[Lin et~al.(2023{\natexlab{b}})Lin, Zhang, Chen, Pramanick, Gao, Wang, Yan, and Shou]{lin2023univtg}
Kevin~Qinghong Lin, Pengchuan Zhang, Joya Chen, Shraman Pramanick, Difei Gao, Alex~Jinpeng Wang, Rui Yan, and Mike~Zheng Shou.
\newblock Univtg: Towards unified video-language temporal grounding.
\newblock In \emph{Proceedings of the IEEE/CVF International Conference on Computer Vision}, pages 2794--2804, 2023{\natexlab{b}}.

\bibitem[Lin et~al.(2022)Lin, Geng, Zhang, Gao, De~Melo, Wang, Dai, Qiao, and Li]{Lin2022frozen}
Ziyi Lin, Shijie Geng, Renrui Zhang, Peng Gao, Gerard De~Melo, Xiaogang Wang, Jifeng Dai, Yu Qiao, and Hongsheng Li.
\newblock Frozen clip models are efficient video learners.
\newblock In \emph{European Conference on Computer Vision}, pages 388--404. Springer, 2022.

\bibitem[Liu et~al.(2020{\natexlab{a}})Liu, Qu, Liu, Dong, Zhou, and Xu]{liu2020jointly}
Daizong Liu, Xiaoye Qu, Xiao-Yang Liu, Jianfeng Dong, Pan Zhou, and Zichuan Xu.
\newblock Jointly cross-and self-modal graph attention network for query-based moment localization.
\newblock In \emph{Proceedings of the 28th ACM International Conference on Multimedia}, pages 4070--4078, 2020{\natexlab{a}}.

\bibitem[Liu et~al.(2023{\natexlab{a}})Liu, Huang, Li, Feng, Wu, and Li]{liu2023revisiting}
Ruyang Liu, Jingjia Huang, Ge Li, Jiashi Feng, Xinglong Wu, and Thomas~H Li.
\newblock Revisiting temporal modeling for clip-based image-to-video knowledge transferring.
\newblock In \emph{Proceedings of the IEEE/CVF Conference on Computer Vision and Pattern Recognition}, pages 6555--6564, 2023{\natexlab{a}}.

\bibitem[Liu et~al.(2025)Liu, Zhao, Zohra, Soldan, Pardo, Xu, Alssum, Ramazanova, Alcázar, Cioppa, Giancola, Hinojosa, and Ghanem]{liu2025opentad}
Shuming Liu, Chen Zhao, Fatimah Zohra, Mattia Soldan, Alejandro Pardo, Mengmeng Xu, Lama Alssum, Merey Ramazanova, Juan~León Alcázar, Anthony Cioppa, Silvio Giancola, Carlos Hinojosa, and Bernard Ghanem.
\newblock Opentad: A unified framework and comprehensive study of temporal action detection.
\newblock \emph{arXiv preprint arXiv:2502.20361}, 2025.

\bibitem[Liu et~al.(2020{\natexlab{b}})Liu, Shen, Yu, and Wang]{liu2020efficient}
Yifan Liu, Chunhua Shen, Changqian Yu, and Jingdong Wang.
\newblock Efficient semantic video segmentation with per-frame inference.
\newblock In \emph{Computer Vision--ECCV 2020: 16th European Conference, Glasgow, UK, August 23--28, 2020, Proceedings, Part X 16}, pages 352--368. Springer, 2020{\natexlab{b}}.

\bibitem[Liu et~al.(2023{\natexlab{b}})Liu, Matsoukas, Strand, Azizpour, and Smith]{liu2023patchdropout}
Yue Liu, Christos Matsoukas, Fredrik Strand, Hossein Azizpour, and Kevin Smith.
\newblock Patchdropout: Economizing vision transformers using patch dropout.
\newblock In \emph{Proceedings of the IEEE/CVF Winter Conference on Applications of Computer Vision}, pages 3953--3962, 2023{\natexlab{b}}.

\bibitem[Liu et~al.(2022)Liu, Ning, Cao, Wei, Zhang, Lin, and Hu]{vswin}
Ze Liu, Jia Ning, Yue Cao, Yixuan Wei, Zheng Zhang, Stephen Lin, and Han Hu.
\newblock Video swin transformer.
\newblock In \emph{Proceedings of the IEEE/CVF Conference on Computer Vision and Pattern Recognition (CVPR)}, 2022.

\bibitem[Luo et~al.(2024)Luo, Huang, Gong, Jin, and Liu]{luo2024zero}
Dezhao Luo, Jiabo Huang, Shaogang Gong, Hailin Jin, and Yang Liu.
\newblock Zero-shot video moment retrieval from frozen vision-language models.
\newblock In \emph{Proceedings of the IEEE/CVF Winter Conference on Applications of Computer Vision}, pages 5464--5473, 2024.

\bibitem[Luo et~al.(2022)Luo, Ji, Zhong, Chen, Lei, Duan, and Li]{luo2022clip4clip}
Huaishao Luo, Lei Ji, Ming Zhong, Yang Chen, Wen Lei, Nan Duan, and Tianrui Li.
\newblock Clip4clip: An empirical study of clip for end to end video clip retrieval and captioning.
\newblock \emph{Neurocomputing}, 508:\penalty0 293--304, 2022.

\bibitem[Ma et~al.(2022)Ma, Xu, Sun, Yan, Zhang, and Ji]{ma2022x}
Yiwei Ma, Guohai Xu, Xiaoshuai Sun, Ming Yan, Ji Zhang, and Rongrong Ji.
\newblock X-clip: End-to-end multi-grained contrastive learning for video-text retrieval.
\newblock In \emph{Proceedings of the 30th ACM International Conference on Multimedia}, pages 638--647, 2022.

\bibitem[Molchanov et~al.(2016)Molchanov, Tyree, Karras, Aila, and Kautz]{molchanov2016pruning}
Pavlo Molchanov, Stephen Tyree, Tero Karras, Timo Aila, and Jan Kautz.
\newblock Pruning convolutional neural networks for resource efficient inference.
\newblock \emph{arXiv preprint arXiv:1611.06440}, 2016.

\bibitem[Moon et~al.(2023)Moon, Hyun, Lee, and Heo]{moon2023correlation}
WonJun Moon, Sangeek Hyun, SuBeen Lee, and Jae-Pil Heo.
\newblock Correlation-guided query-dependency calibration in video representation learning for temporal grounding.
\newblock \emph{arXiv preprint arXiv:2311.08835}, 2023.

\bibitem[Mu et~al.(2024)Mu, Mo, and Li]{mu2024snag}
Fangzhou Mu, Sicheng Mo, and Yin Li.
\newblock Snag: Scalable and accurate video grounding.
\newblock In \emph{Proceedings of the IEEE/CVF Conference on Computer Vision and Pattern Recognition}, pages 18930--18940, 2024.

\bibitem[Mun et~al.(2020)Mun, Cho, and Han]{Mun_2020_CVPR}
Jonghwan Mun, Minsu Cho, and Bohyung Han.
\newblock {Local-Global Video-Text Interactions for Temporal Grounding}.
\newblock In \emph{Proceedings of the IEEE/CVF Conference on Computer Vision and Pattern Recognition (CVPR)}, 2020.

\bibitem[Nam et~al.(2021)Nam, Ahn, Kang, Ha, and Choi]{nam2021zero}
Jinwoo Nam, Daechul Ahn, Dongyeop Kang, Seong~Jong Ha, and Jonghyun Choi.
\newblock Zero-shot natural language video localization.
\newblock In \emph{Proceedings of the IEEE/CVF International Conference on Computer Vision}, pages 1470--1479, 2021.

\bibitem[Ni et~al.(2022)Ni, Peng, Chen, Zhang, Meng, Fu, Xiang, and Ling]{ni2022expanding}
Bolin Ni, Houwen Peng, Minghao Chen, Songyang Zhang, Gaofeng Meng, Jianlong Fu, Shiming Xiang, and Haibin Ling.
\newblock Expanding language-image pretrained models for general video recognition.
\newblock In \emph{European Conference on Computer Vision}, pages 1--18. Springer, 2022.

\bibitem[Otani et~al.(2020)Otani, Nakashima, Rahtu, and Heikkil{\"a}]{otani2020uncovering}
Mayu Otani, Yuta Nakashima, Esa Rahtu, and Janne Heikkil{\"a}.
\newblock Uncovering hidden challenges in query-based video moment retrieval.
\newblock \emph{arXiv preprint arXiv:2009.00325}, 2020.

\bibitem[Pan et~al.(2022)Pan, Lin, Zhu, Shao, and Li]{pan2022st}
Junting Pan, Ziyi Lin, Xiatian Zhu, Jing Shao, and Hongsheng Li.
\newblock St-adapter: Parameter-efficient image-to-video transfer learning.
\newblock \emph{Advances in Neural Information Processing Systems}, 35:\penalty0 26462--26477, 2022.

\bibitem[Pan et~al.(2023)Pan, He, Gong, Lv, Shen, Peng, and Zhao]{Pan_2023_ICCV}
Yulin Pan, Xiangteng He, Biao Gong, Yiliang Lv, Yujun Shen, Yuxin Peng, and Deli Zhao.
\newblock Scanning only once: An end-to-end framework for fast temporal grounding in long videos.
\newblock In \emph{Proceedings of the IEEE/CVF International Conference on Computer Vision (ICCV)}, pages 13767--13777, 2023.

\bibitem[Park et~al.(2023)Park, Lee, and Sohn]{park2023dual}
Jungin Park, Jiyoung Lee, and Kwanghoon Sohn.
\newblock Dual-path adaptation from image to video transformers.
\newblock In \emph{Proceedings of the IEEE/CVF Conference on Computer Vision and Pattern Recognition}, pages 2203--2213, 2023.

\bibitem[Radford et~al.(2019)Radford, Wu, Child, Luan, Amodei, Sutskever, et~al.]{radford2019language}
Alec Radford, Jeffrey Wu, Rewon Child, David Luan, Dario Amodei, Ilya Sutskever, et~al.
\newblock Language models are unsupervised multitask learners.
\newblock \emph{OpenAI blog}, 1\penalty0 (8):\penalty0 9, 2019.

\bibitem[Radford et~al.(2021)Radford, Kim, Hallacy, Ramesh, Goh, Agarwal, Sastry, Askell, Mishkin, Clark, et~al.]{radford2021learning}
Alec Radford, Jong~Wook Kim, Chris Hallacy, Aditya Ramesh, Gabriel Goh, Sandhini Agarwal, Girish Sastry, Amanda Askell, Pamela Mishkin, Jack Clark, et~al.
\newblock Learning transferable visual models from natural language supervision.
\newblock In \emph{International conference on machine learning}, pages 8748--8763. PMLR, 2021.

\bibitem[Richter et~al.(2021)Richter, Byttner, Krumnack, Wiedenroth, Schallner, and Shenk]{richter2021input}
Mats~L Richter, Wolf Byttner, Ulf Krumnack, Anna Wiedenroth, Ludwig Schallner, and Justin Shenk.
\newblock (input) size matters for cnn classifiers.
\newblock In \emph{Artificial Neural Networks and Machine Learning--ICANN 2021: 30th International Conference on Artificial Neural Networks, Bratislava, Slovakia, September 14--17, 2021, Proceedings, Part II 30}, pages 133--144. Springer, 2021.

\bibitem[Roerdink and Meijster(2000)]{roerdink2000watershed}
Jos~BTM Roerdink and Arnold Meijster.
\newblock The watershed transform: Definitions, algorithms and parallelization strategies.
\newblock \emph{Fundamenta informaticae}, 41\penalty0 (1-2):\penalty0 187--228, 2000.

\bibitem[Sandler et~al.(2018)Sandler, Howard, Zhu, Zhmoginov, and Chen]{sandler2018mobilenetv2}
Mark Sandler, Andrew Howard, Menglong Zhu, Andrey Zhmoginov, and Liang-Chieh Chen.
\newblock Mobilenetv2: Inverted residuals and linear bottlenecks.
\newblock In \emph{Proceedings of the IEEE conference on computer vision and pattern recognition}, pages 4510--4520, 2018.

\bibitem[Shen et~al.(2021)Shen, Li, Tan, Bansal, Rohrbach, Chang, Yao, and Keutzer]{shen2021much}
Sheng Shen, Liunian~Harold Li, Hao Tan, Mohit Bansal, Anna Rohrbach, Kai-Wei Chang, Zhewei Yao, and Kurt Keutzer.
\newblock How much can clip benefit vision-and-language tasks?
\newblock \emph{arXiv preprint arXiv:2107.06383}, 2021.

\bibitem[Simonyan and Zisserman(2014)]{simonyan2014very}
Karen Simonyan and Andrew Zisserman.
\newblock Very deep convolutional networks for large-scale image recognition.
\newblock \emph{arXiv preprint arXiv:1409.1556}, 2014.

\bibitem[Soldan et~al.(2021)Soldan, Xu, Qu, Tegner, and Ghanem]{soldan2021vlg}
Mattia Soldan, Mengmeng Xu, Sisi Qu, Jesper Tegner, and Bernard Ghanem.
\newblock Vlg-net: Video-language graph matching network for video grounding.
\newblock In \emph{Proceedings of the IEEE/CVF International Conference on Computer Vision}, pages 3224--3234, 2021.

\bibitem[Soldan et~al.(2022)Soldan, Pardo, Alc{\'a}zar, Caba, Zhao, Giancola, and Ghanem]{soldan2022mad}
Mattia Soldan, Alejandro Pardo, Juan~Le{\'o}n Alc{\'a}zar, Fabian Caba, Chen Zhao, Silvio Giancola, and Bernard Ghanem.
\newblock Mad: A scalable dataset for language grounding in videos from movie audio descriptions.
\newblock In \emph{Proceedings of the IEEE/CVF Conference on Computer Vision and Pattern Recognition}, pages 5026--5035, 2022.

\bibitem[Speer et~al.(2017)Speer, Chin, and Havasi]{speer2017conceptnet}
Robyn Speer, Joshua Chin, and Catherine Havasi.
\newblock Conceptnet 5.5: An open multilingual graph of general knowledge.
\newblock In \emph{Proceedings of the AAAI conference on artificial intelligence}, 2017.

\bibitem[Sun et~al.(2023)Sun, Gao, Zhu, Wang, and Zhou]{sun2023video}
Xin Sun, Jialin Gao, Yizhe Zhu, Xuan Wang, and Xi Zhou.
\newblock Video moment retrieval via comprehensive relation-aware network.
\newblock \emph{IEEE Transactions on Circuits and Systems for Video Technology}, 2023.

\bibitem[Tong et~al.(2022)Tong, Song, Wang, and Wang]{tong2022videomae}
Zhan Tong, Yibing Song, Jue Wang, and Limin Wang.
\newblock Videomae: Masked autoencoders are data-efficient learners for self-supervised video pre-training.
\newblock \emph{Advances in neural information processing systems}, 35:\penalty0 10078--10093, 2022.

\bibitem[Tran et~al.(2015)Tran, Bourdev, Fergus, Torresani, and Paluri]{tran2015learning}
Du Tran, Lubomir Bourdev, Rob Fergus, Lorenzo Torresani, and Manohar Paluri.
\newblock Learning spatiotemporal features with 3d convolutional networks.
\newblock In \emph{Proceedings of the IEEE international conference on computer vision}, pages 4489--4497, 2015.

\bibitem[Tu et~al.(2023)Tu, Dai, Wu, Cheng, Hu, and Jiang]{tu2023implicit}
Shuyuan Tu, Qi Dai, Zuxuan Wu, Zhi-Qi Cheng, Han Hu, and Yu-Gang Jiang.
\newblock Implicit temporal modeling with learnable alignment for video recognition.
\newblock \emph{arXiv preprint arXiv:2304.10465}, 2023.

\bibitem[Vedantam et~al.(2015)Vedantam, Lawrence~Zitnick, and Parikh]{vedantam2015cider}
Ramakrishna Vedantam, C Lawrence~Zitnick, and Devi Parikh.
\newblock Cider: Consensus-based image description evaluation.
\newblock In \emph{Proceedings of the IEEE conference on computer vision and pattern recognition}, pages 4566--4575, 2015.

\bibitem[Wang et~al.(2022{\natexlab{a}})Wang, Wu, Liu, and Yan]{wang2022prompt}
Guolong Wang, Xun Wu, Zhaoyuan Liu, and Junchi Yan.
\newblock Prompt-based zero-shot video moment retrieval.
\newblock In \emph{Proceedings of the 30th ACM International Conference on Multimedia}, pages 413--421, 2022{\natexlab{a}}.

\bibitem[Wang et~al.(2016)Wang, Xiong, Wang, Qiao, Lin, Tang, and Van~Gool]{TSN2016ECCV}
Limin Wang, Yuanjun Xiong, Zhe Wang, Yu Qiao, Dahua Lin, Xiaoou Tang, and Luc Van~Gool.
\newblock Temporal segment networks: Towards good practices for deep action recognition.
\newblock In \emph{Proceedings of the European Conference on Computer Vision (ECCV)}, 2016.

\bibitem[Wang et~al.(2022{\natexlab{b}})Wang, Li, Li, He, Huang, Zhao, Zhang, Xu, Liu, Wang, et~al.]{wang2022internvideo}
Yi Wang, Kunchang Li, Yizhuo Li, Yinan He, Bingkun Huang, Zhiyu Zhao, Hongjie Zhang, Jilan Xu, Yi Liu, Zun Wang, et~al.
\newblock Internvideo: General video foundation models via generative and discriminative learning.
\newblock \emph{arXiv preprint arXiv:2212.03191}, 2022{\natexlab{b}}.

\bibitem[Wang et~al.(2024)Wang, Li, Li, Yu, He, Chen, Pei, Zheng, Xu, Wang, et~al.]{wang2024internvideo2}
Yi Wang, Kunchang Li, Xinhao Li, Jiashuo Yu, Yinan He, Guo Chen, Baoqi Pei, Rongkun Zheng, Jilan Xu, Zun Wang, et~al.
\newblock {InternVideo2}: Scaling video foundation models for multimodal video understanding.
\newblock \emph{arXiv preprint arXiv:2403.15377}, 2024.

\bibitem[Weng et~al.(2023)Weng, Yang, Li, Wu, and Jiang]{weng2023open}
Zejia Weng, Xitong Yang, Ang Li, Zuxuan Wu, and Yu-Gang Jiang.
\newblock Transforming clip to an open-vocabulary video model via interpolated weight optimization.
\newblock \emph{arXiv preprint arXiv:2302.00624}, 2023.

\bibitem[Wu et~al.(2018)Wu, Zaheer, Hu, Manmatha, Smola, and Kr{\"a}henb{\"u}hl]{wu2018compressed}
Chao-Yuan Wu, Manzil Zaheer, Hexiang Hu, R Manmatha, Alexander~J Smola, and Philipp Kr{\"a}henb{\"u}hl.
\newblock Compressed video action recognition.
\newblock In \emph{Proceedings of the IEEE conference on computer vision and pattern recognition}, pages 6026--6035, 2018.

\bibitem[Wu et~al.(2022{\natexlab{a}})Wu, Li, Mangalam, Fan, Xiong, Malik, and Feichtenhofer]{wu2022memvit}
Chao-Yuan Wu, Yanghao Li, Karttikeya Mangalam, Haoqi Fan, Bo Xiong, Jitendra Malik, and Christoph Feichtenhofer.
\newblock Memvit: Memory-augmented multiscale vision transformer for efficient long-term video recognition.
\newblock In \emph{Proceedings of the IEEE/CVF Conference on Computer Vision and Pattern Recognition}, pages 13587--13597, 2022{\natexlab{a}}.

\bibitem[Wu et~al.(2022{\natexlab{b}})Wu, Zhang, Peng, Liu, Xiao, Fu, and Yuan]{tiny_vit}
Kan Wu, Jinnian Zhang, Houwen Peng, Mengchen Liu, Bin Xiao, Jianlong Fu, and Lu Yuan.
\newblock Tinyvit: Fast pretraining distillation for small vision transformers.
\newblock In \emph{European conference on computer vision (ECCV)}, 2022{\natexlab{b}}.

\bibitem[Xu et~al.(2020)Xu, Zhao, Rojas, Thabet, and Ghanem]{Xu_2020_CVPR}
Mengmeng Xu, Chen Zhao, David~S. Rojas, Ali Thabet, and Bernard Ghanem.
\newblock G-tad: Sub-graph localization for temporal action detection.
\newblock In \emph{Proceedings of the IEEE/CVF Conference on Computer Vision and Pattern Recognition (CVPR)}, 2020.

\bibitem[Xu et~al.(2023)Xu, Soldan, Gao, Liu, P{\'e}rez-R{\'u}a, and Ghanem]{xu2023boundary}
Mengmeng Xu, Mattia Soldan, Jialin Gao, Shuming Liu, Juan-Manuel P{\'e}rez-R{\'u}a, and Bernard Ghanem.
\newblock Boundary-denoising for video activity localization.
\newblock \emph{arXiv preprint arXiv:2304.02934}, 2023.

\bibitem[Xue et~al.(2022)Xue, Sun, Liu, Fu, Song, Li, and Luo]{xue2022clip}
Hongwei Xue, Yuchong Sun, Bei Liu, Jianlong Fu, Ruihua Song, Houqiang Li, and Jiebo Luo.
\newblock Clip-vip: Adapting pre-trained image-text model to video-language representation alignment.
\newblock \emph{arXiv preprint arXiv:2209.06430}, 2022.

\bibitem[Yang et~al.(2023)Yang, Zhu, Xie, Zhang, Chen, and Li]{yang2023aim}
Taojiannan Yang, Yi Zhu, Yusheng Xie, Aston Zhang, Chen Chen, and Mu Li.
\newblock Aim: Adapting image models for efficient video action recognition.
\newblock \emph{arXiv preprint arXiv:2302.03024}, 2023.

\bibitem[Zeng et~al.(2020)Zeng, Xu, Huang, Chen, Tan, and Gan]{Zeng_2020_CVPR}
Runhao Zeng, Haoming Xu, Wenbing Huang, Peihao Chen, Mingkui Tan, and Chuang Gan.
\newblock {Dense Regression Network for Video Grounding}.
\newblock In \emph{Proceedings of the IEEE/CVF Conference on Computer Vision and Pattern Recognition (CVPR)}, 2020.

\bibitem[Zhang et~al.(2022)Zhang, Wu, and Li]{zhang2022actionformer}
Chen-Lin Zhang, Jianxin Wu, and Yin Li.
\newblock Actionformer: Localizing moments of actions with transformers.
\newblock In \emph{European Conference on Computer Vision}, pages 492--510. Springer, 2022.

\bibitem[Zhang et~al.(2021)Zhang, Sun, Jing, and Zhou]{zhang2021towards}
Hao Zhang, Aixin Sun, Wei Jing, and Joey~Tianyi Zhou.
\newblock Towards debiasing temporal sentence grounding in video.
\newblock \emph{arXiv preprint arXiv:2111.04321}, 2021.

\bibitem[Zhang et~al.(2020)Zhang, Peng, Fu, and Luo]{zhang2020learning}
Songyang Zhang, Houwen Peng, Jianlong Fu, and Jiebo Luo.
\newblock Learning 2d temporal adjacent networks for moment localization with natural language.
\newblock In \emph{Proceedings of the AAAI Conference on Artificial Intelligence}, pages 12870--12877, 2020.

\bibitem[Zhang et~al.(2019)Zhang, Kishore, Wu, Weinberger, and Artzi]{zhang2019bertscore}
Tianyi Zhang, Varsha Kishore, Felix Wu, Kilian~Q Weinberger, and Yoav Artzi.
\newblock Bertscore: Evaluating text generation with bert.
\newblock \emph{arXiv preprint arXiv:1904.09675}, 2019.

\bibitem[Zheng et~al.(2022)Zheng, Huang, Chen, Peng, and Liu]{zheng2022weakly2}
Minghang Zheng, Yanjie Huang, Qingchao Chen, Yuxin Peng, and Yang Liu.
\newblock Weakly supervised temporal sentence grounding with gaussian-based contrastive proposal learning.
\newblock In \emph{Proceedings of the IEEE/CVF Conference on Computer Vision and Pattern Recognition}, pages 15555--15564, 2022.

\bibitem[Zheng et~al.(2023)Zheng, Gong, Jin, Peng, and Liu]{zheng2023generating}
Minghang Zheng, Shaogang Gong, Hailin Jin, Yuxin Peng, and Yang Liu.
\newblock Generating structured pseudo labels for noise-resistant zero-shot video sentence localization.
\newblock In \emph{Proceedings of the 61st Annual Meeting of the Association for Computational Linguistics (Volume 1: Long Papers)}, pages 14197--14209, 2023.

\end{thebibliography}
}

\end{document}